\documentclass{article}

\usepackage{natbib}
\usepackage[preprint,nonatbib]{neurips_2021}
\usepackage{nicefrac}

\usepackage[utf8]{inputenc} %
\usepackage[T1]{fontenc}    %
\usepackage{hyperref}       %
\usepackage{url}            %
\usepackage{booktabs}       %
\usepackage{amsfonts}       %
\usepackage{nicefrac}       %
\usepackage{microtype}      %
\usepackage{xcolor}         %
\usepackage{tcolorbox}
\usepackage{bbm}

\usepackage{microtype}
\usepackage{graphicx,subfigure}
\usepackage{booktabs} %
\usepackage{amsmath}
\usepackage{amssymb}
\usepackage{amsthm}
\usepackage{mathtools}
\usepackage{caption}
\usepackage{framed}
\usepackage{enumitem}
\usepackage{thmtools} 
\usepackage{thm-restate}
\title{Vanishing Curvature and the Power of Adaptive Methods in Randomly Initialized Deep Networks}

\author{%
  Antonio Orvieto\thanks{Equal contribution}
   \qquad
 Jonas Kohler\footnotemark[1]
  \qquad
 Dario Pavllo\AND
 Thomas Hofmann
  \qquad
 Aurelien Lucchi\vspace{4.5mm}\\
 Department of Computer Science\\
 ETH Zurich
}

\renewcommand{\a}{{\bf a}}

\newcommand{\eb}{{\bf e}}

\newcommand{\h}{{\bf h}}

\newcommand{\w}{{\bf w}}

\newcommand{\x}{{\bf x}}
\newcommand{\y}{{\bf y}}
\newcommand{\z}{{\bf z}}

\newcommand{\vxi}{{\boldsymbol \xi}}

\newcommand{\va}{{\boldsymbol \alpha}}

\def\Am{{\bf A}}
\def\Bm{{\bf B}}

\def\Dm{{\bf D}}

\def\Km{{\bf K}}

\def\Im{{\bf I}}

\def\Qm{{\bf Q}}

\def\Wm{{\bf W}}
\def\M{{\bf W}}
\def\Dx{{\bf D}_{\x}}

\newcommand{\wM}[1]{\mb W^{#1}}
\newcommand{\wMp}[1]{\mb W^{#1}_\phi}
\newcommand{\wMt}[1]{\mb W^{#1\top}}
\newcommand{\mb}{\mathbf}
\newcommand{\jacobi}[2]{\frac{\partial #1}{\partial #2}}
\def\W{{\bf W}}

\def\Ym{{\bf Y}}
\def\Xm{{\bf X}}

\newcommand{\Ls}{{\mathcal L}}
\newcommand{\E}{{\mathbb E}}

\newcommand{\R}{{\mathbb{R}}}

\DeclareMathOperator{\prob}{Pr}

\newtheorem{theorem}{Theorem}
\newtheorem{lemma}[theorem]{Lemma}
\newtheorem{proposition}[theorem]{Proposition}
\newtheorem{corollary}[theorem]{Corollary}
\newtheorem{assumption}[theorem]{Assumption}

\usepackage{tcolorbox}
\usepackage{hyperref}
\tcbset{boxsep=0mm,boxrule=0pt,colframe=white,arc=0mm,left=0.5mm,right=0.5mm}

\usepackage{wrapfig}

\newcommand{\figsize}{0.38}

\begin{document}

\maketitle

\begin{abstract}
This paper revisits the so-called \textit{vanishing gradient phenomenon}, which commonly occurs in deep randomly initialized neural networks. Leveraging an in-depth analysis of neural chains, we first show that vanishing gradients cannot be circumvented when the network width scales with less than $O(\text{depth})$, even when initialized with the popular Xavier and He initializations. Second, we extend the analysis to second-order derivatives and show that random i.i.d. initialization also gives rise to Hessian matrices with eigenspectra that vanish as networks grow in depth. Whenever this happens, optimizers are initialized in a very flat, saddle point-like plateau, which is particularly hard to escape with stochastic gradient descent (SGD) as its escaping time is inversely related to curvature. We believe that this observation is crucial for fully understanding (a)  historical difficulties of training deep nets with vanilla SGD, (b) the success of adaptive gradient methods (which naturally adapt to curvature and thus quickly escape flat plateaus) and (c) the effectiveness of modern architectural components like residual connections and normalization layers.
\end{abstract}

\section{Introduction and related work}\label{sec:intro}
In the last decade, network depth has emerged as a key component for the success of modern deep learning \citep{he2016identity}, providing significant improvements in terms of generalization, particularly in the field of computer vision \citep{he2016deep,he2016identity} and natural language processing \citep{brown2020language}. These benefits are mostly attributed to gains in representational power. In fact, it was shown in \citep{telgarsky2016benefits} that there exist deep neural networks of bounded width that cannot be approximated by shallow networks, unless their layers grow exponentially wide. From an optimization perspective, however, depth introduces several problems when training after random initialization. First, in the infinite depth limit, a collapse in the rank of the network mapping prevents information propagation, which renders learning impossible \citep{schoenholz2016deep,pennington2018emergence,daneshmand2020batch}. Secondly, even in finite but large depth the so-called \textit{vanishing gradient} problem commonly makes it difficult to train deep (un-normalized) networks with stochastic gradient descent \citep{hochreiter1991untersuchungen,bengio1994learning,pascanu2013difficulty}.\looseness=-1

 \begin{figure}
 \begin{center}

  \begin{minipage}[b]{\figsize\linewidth}
\centering
\includegraphics[width=1.05\textwidth]{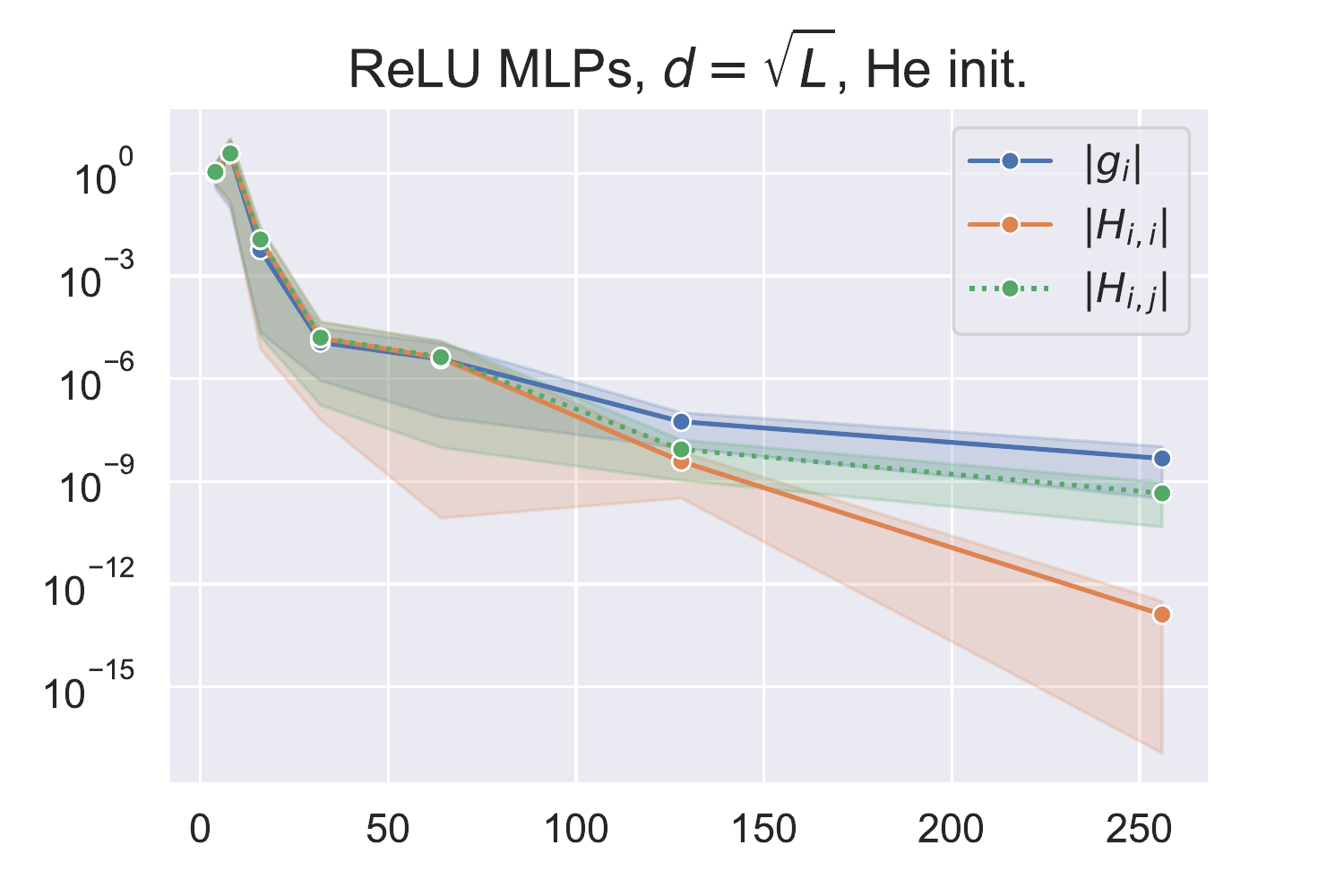}
\end{minipage}
\hspace{0.005cm}
\begin{minipage}[b]{\figsize\linewidth}
\centering
\includegraphics[width=1.05\textwidth]{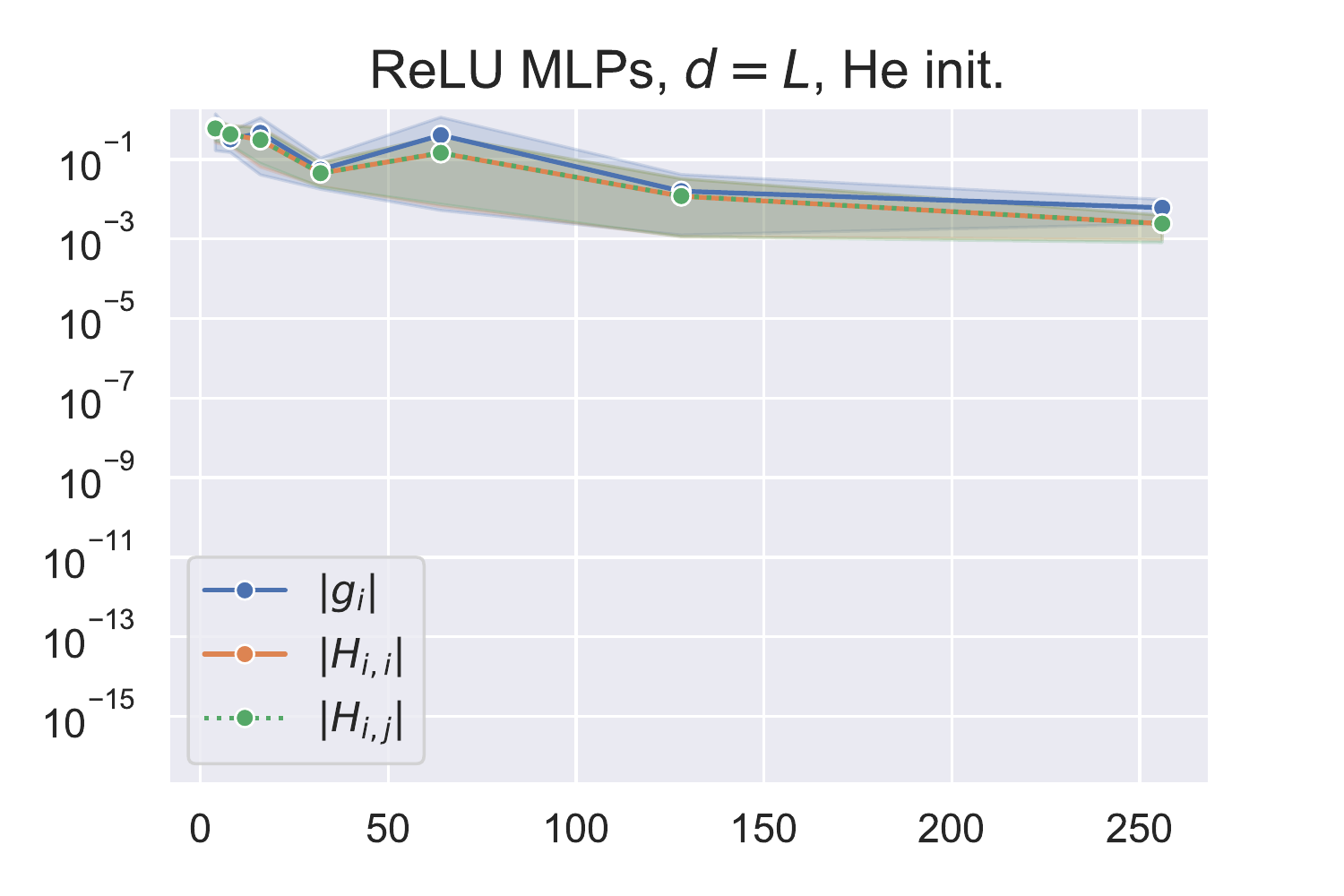}
\end{minipage}

\end{center}
\vspace{-3mm}
\caption{ \small \textbf{ \small Effect of width in ReLU MLPs:} Gradient and curvature scaling on Fashion-MNIST over depth. While quantities vanish on the left ($d=\sqrt{L}$), the right shows stable magnitudes. Mean and 95\% CI of 15 runs.}
\label{fig:width_effects_mlp}
\vspace{-3mm}
\end{figure}

In this work, we focus on the latter setting. First, we show that networks that are deeper than wide suffer from \textit{vanishing gradients at initialization even when following the initialization schemes proposed in} \citep{glorot2010understanding,he2015delving}. This is somewhat surprising at first, since the analysis undertaken in these seminal works suggests stable gradient norms for any depth, in expectation. However, as we show in Section \ref{sec:width}, the expectation analysis has an important shortcoming: When networks are deeper than wide, the initialization variance suggested to be optimal in order to stabilize the expected forward- and backpropagation norms yields exploding higher moments of these quantities. As a result, their distribution becomes fat tailed in depth and hence the expected value itself is increasingly unlikely to be observed. Thus, \textit{we instead study the median of the propagation norms, which we find to indeed vanish in depth}.  

Second, we reveal that, in the above settings, not only the gradient vanishes as depth increases, but also the entries of the Hessian become smaller and smaller. Hence, by Gershgorin's theorem \citep{gershgorin1931uber}, the \textit{eigenvalues shrink in depth}. As a result, random i.i.d. initialization ends up positioning optimization methods in a \textit{very flat, saddle point like region} around the origin, where gradients are small and both negative as well as positive curvature exist. %
This is particularly\textit{ unfortunate for stochastic gradient descent} (SGD) as its saddle escaping time is inversely related to the magnitude of the most negative eigenvalue (see Prop.~\ref{prop:GD_slow} as well as \citep{daneshmand2018escaping,fang2019sharp}). This observation complements the vanishing gradient argument, whose implications for optimization were so far rather vague since small gradients on their own are not a-priori problematic for gradient-based learning: In fact, it is often the Lipschitz constant (i.e. curvature) and not the norm of the gradients that determines the speed of convergence of gradient descent~\citep{nesterov2018lectures}. Even around saddle points, sufficient negative curvature allows stochastic gradient descent to make fast progress despite small gradient norms.\footnote{To be more precise, the anisotropic noise of SGD has been shown to be aligned with negative curvature directions \citep{zhu2019anisotropic,daneshmand2018escaping,fang2019sharp}, which allows for a per-step progress of $\mathcal{O}(\lambda^3)$ where $\lambda$ is the curvature along the current search direction~\citep{curtis2019exploiting}.} Hence, vanishing gradients on their own fall short of a satisfactory description for the difficulty of training very deep networks with plain SGD.

We believe that it is precisely the vanishing curvature phenomenon, combined with the deficiencies of SGD in escaping flat saddles, that lead to developments on three fronts:  (i) \textit{modern architectural components}: batch normalization (BN)~\citep{ioffe2015batch} and residual connections~\citep{he2016deep}, which collectively circumvent vanishing gradients/curvature even in non-linear networks (Fig.~\ref{fig:cnn_vanishing}). (ii) \textit{robust initialization schemes}: orthogonal-~\citep{saxe2013exact}, which is restricted to \textit{linear} networks (Fig.~\ref{fig:orthogonal_init}) and Fixup/Skip initialization~\citep{zhang2019fixup,de2020batch}, which downscale the parametric branch of residual architectures proportionally to their depth, thus trivially yielding identity mappings in the limit. %
(iii) \textit{Adaptive training algorithms}: notably several years before the emergence of BN and residual connections, so-called adaptive gradient methods~\citep{duchi2011adaptive,tieleman2012lecture} have shown remarkable success in training deep networks. We believe that their increased popularity is (at least partially) attributable to the fact that adaptive gradients methods are robust to this failure mode of random i.i.d. initialization because they naturally adapt to curvature, which allows them to escape saddles in time independent of their flatness \citep{dauphin2015equilibrated,staib2019escaping}.

In summary, our contribution is threefold:
\vspace{-2mm}
\begin{itemize}[leftmargin=*]
  \setlength\itemsep{0.1em}
    \item We discuss shortcomings of the analysis in \cite{glorot2010understanding,he2015delving} and clarify when gradients vanish even with the initialization proposed in these seminal works.
    \item We enhance the understanding of the vanishing gradient phenomenon by showing that they co-occur with vanishing curvature, giving rise to flat plateaus at initialization. This effect gives a comprehensive understanding of the difficulty of training (classical) deep nets with (vanilla) SGD.
    \item Finally, we link the remarkable curvature adaptation capability of adaptive gradient methods to this phenomenon and show that it allows them to optimize networks of any depth.
\end{itemize}

\section{Notation and setting}
\vspace{-2mm}
In our theoretical analysis, we consider the L2 loss associated with a multilayer perceptron~(MLP)
\begin{align}\label{eq:setting}
\Ls(\W) = \frac{1}{2n} \sum_{i=1}^n \|\y_i - \Bm \Dm^L \wMp{L:1}\Am\x_i\|^2_2, \quad \M_\phi^{L:1} :=\M^L\Dm^{L-1}\M^{L-1} \cdots \M^2\Dm^1\M^1\Dm^0
\end{align}
where $\x_i\in\mathbb{R}^{d_{in}}, \y_i\in\mathbb{R}^{d_{out}}$, $\Am \in \mathbb{R}^{d\times d_{in}}$,$\Bm \in \mathbb{R}^{d_{out}\times d}$, and $\M^\ell \in \R^{d \times d}, \forall \ell=1,\ldots,L$ . $\Dm^\ell$ is the diagonal matrix of activation gates w.r.t the non-linearity $\phi$ at layer $\ell$, which we consider to be either $\phi(x)=x$ (linear networks) or $\phi(x)=\max\{x,0\}$ (ReLU networks). 

\begin{assumption}[Random initialization]\label{ass:init}
Each entry of $\W^{\ell}$~($\ell=1,\ldots,L$) is initialized i.i.d. with some distribution $\mathcal{P}$ symmetric around zero with variance $\sigma^2<\infty$ and fourth moment $\mu_4<\infty$.
\end{assumption}
\vspace{-2mm}
For more than a decade, the standard choice of initialization variance was $\sigma^2=\frac{1}{3d}$~\citep{lecun2012efficient}. Motivated by repeated observations of the \textit{vanishing gradient problem}, an improved initialization was suggested by ~\citep{glorot2010understanding} and~\citep{he2015delving}. In the following, we define the parameter $p=1$ for the linear case and $p=1/2$ for the ReLU case.

\begin{tcolorbox}
\begin{proposition}[\citep{glorot2010understanding}, \citep{he2015delving}]\label{prop:glorot_he}
Under Assumption \ref{ass:init}, the variance of the weight gradients $\text{Var}(\partial \Ls(\W)/\partial \Wm^\ell)$ scales as $(pd\sigma^2)^L$ across all layers $\ell=1,\ldots,L$. When initializing with $\sigma^2=\tfrac{1}{3d}$ (LeCun init.), this quantity vanishes in depth. Instead, choosing $\sigma^2 = 1/d$ in the linear case (Xavier init.), and $\sigma^2 = 2/d$ in the ReLU case (He init), stabilizes the variance.
\end{proposition}
\end{tcolorbox}
\vspace{-3mm}
\begin{proof}
Let $\a^{\ell+1}=\M^\ell \h^{\ell}$ be the preactivation of layer $\ell+1$, computed using $\h^{\ell}=\Dm^\ell \a^{\ell}$, the activation at layer $\ell$ . Let $a^{\ell+1}$, $w^\ell$ and $h^{\ell}$ represent the random variables corresponding to each element in $\a^{\ell+1}$, $\M^\ell$ and $\h^{\ell}$ respectively. Since $w^\ell$ is zero mean, we have that $\text{Var}[a^{\ell+1}] = d \cdot \text{Var}[w^{\ell}] \cdot \E[(h^{\ell})^2]$. Finally, since $\E[(h^{\ell})^2]= p \text{Var}[(a^{\ell})^2]$ ($p=1$ for linear nets and $1/2$ for ReLU nets), we end up with $\text{Var}[a^{\ell+1}] = d \sigma^2 p \cdot \text{Var}[a^{\ell}]$, which yields  $\text{Var}[a^{\ell+1}] = \text{Var}[a^{\ell}]$ for $\sigma^2 =\frac{1}{dp}$.
\end{proof}
\vspace{-3mm}
For example, for the uniform initialization $\mathcal{P} =\mathcal{U}[-\tau,\tau]$,  we have $\sigma^2 = \tau^2/3$ and hence the "optimal" initialization range amounts to $\tau =\sqrt{3/d}$ in the linear - and $\tau =\sqrt{6/d}$ in the ReLU case.

\section{Vanishing in neural chains and implications for optimization}\label{sec:width}
\vspace{-2mm}
To illustrate an important shortcoming in the analyses of \citep{glorot2010understanding} and \citep{he2015delving}, we consider a deep linear network of width one (henceforth called \textit{neural chain}). While these networks are utterly useless for practical applications, they are sufficient to exhibit some critical properties of the loss landscape, that generalize to wider nets~(see next section). 

\begin{figure}
\begin{center}
    \begin{minipage}[b]{0.32\linewidth}
\centering
\includegraphics[width=1.15\textwidth]{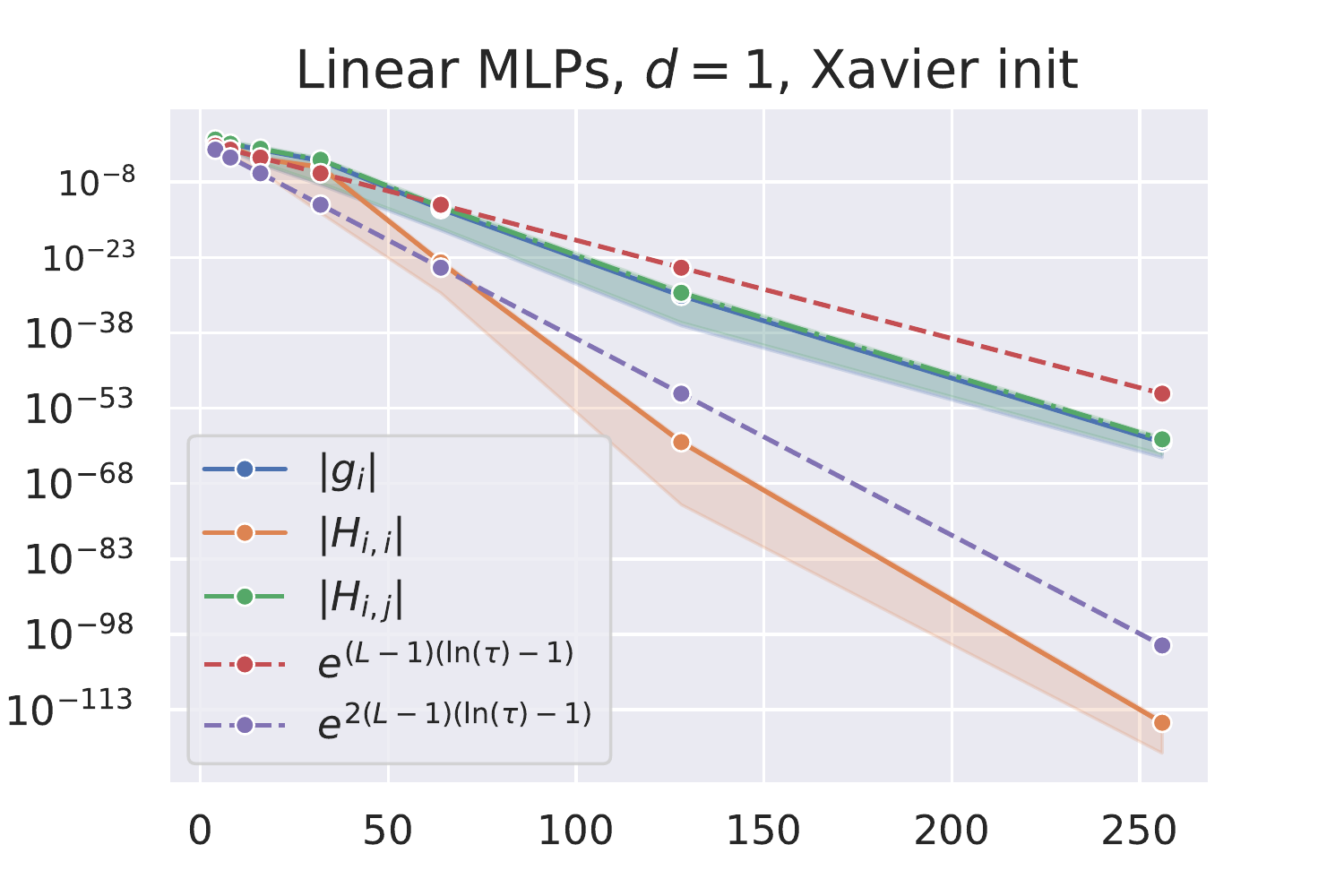}
\hspace*{0.35cm}\tiny{depth}
\end{minipage}
\begin{minipage}[b]{0.32\linewidth}
\centering
\includegraphics[width=1.15\textwidth]{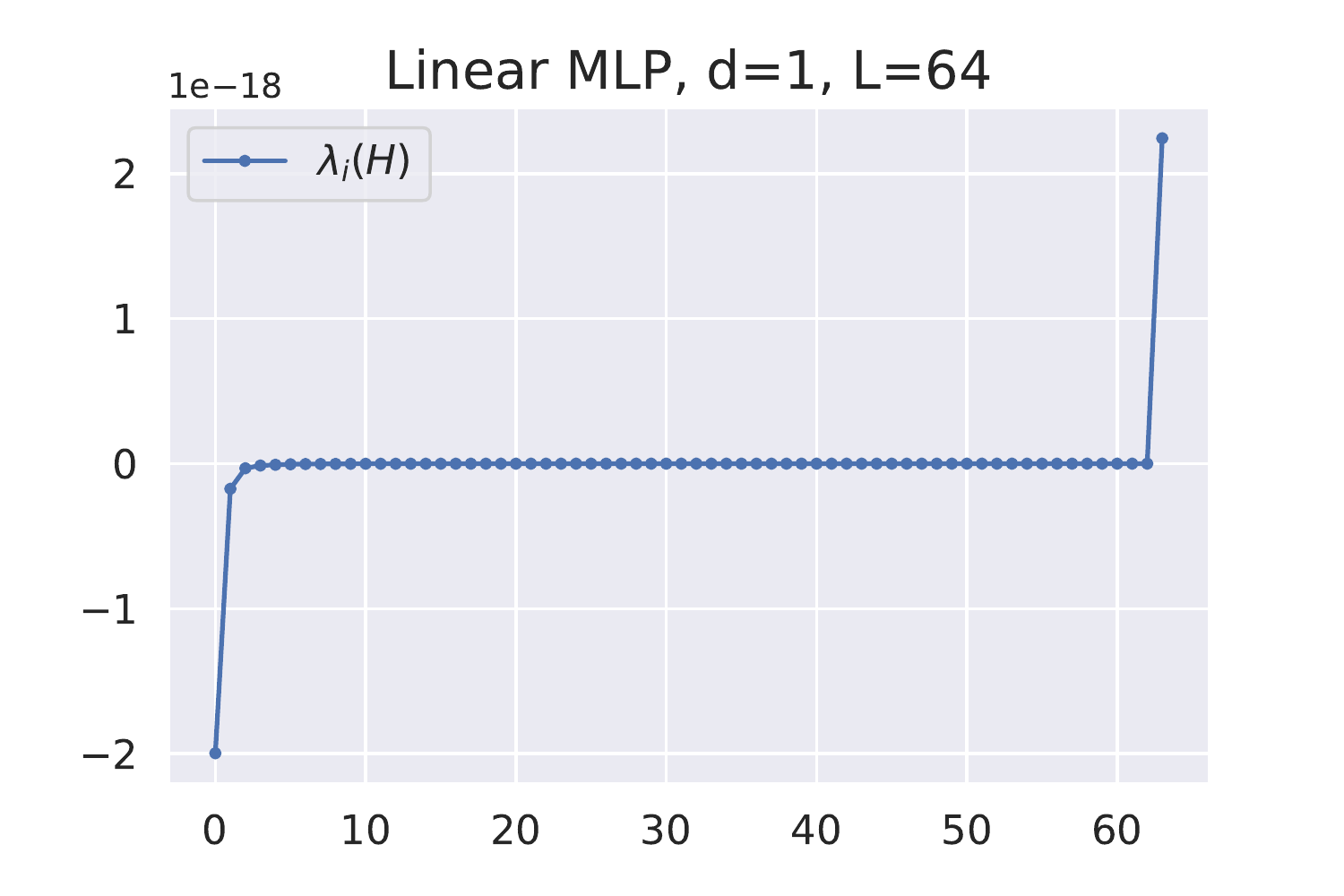}
\hspace*{0.4cm}\tiny{eigenvalue index}
\end{minipage}
\begin{minipage}[b]{0.32\linewidth}
\centering
\includegraphics[width=1.15\textwidth]{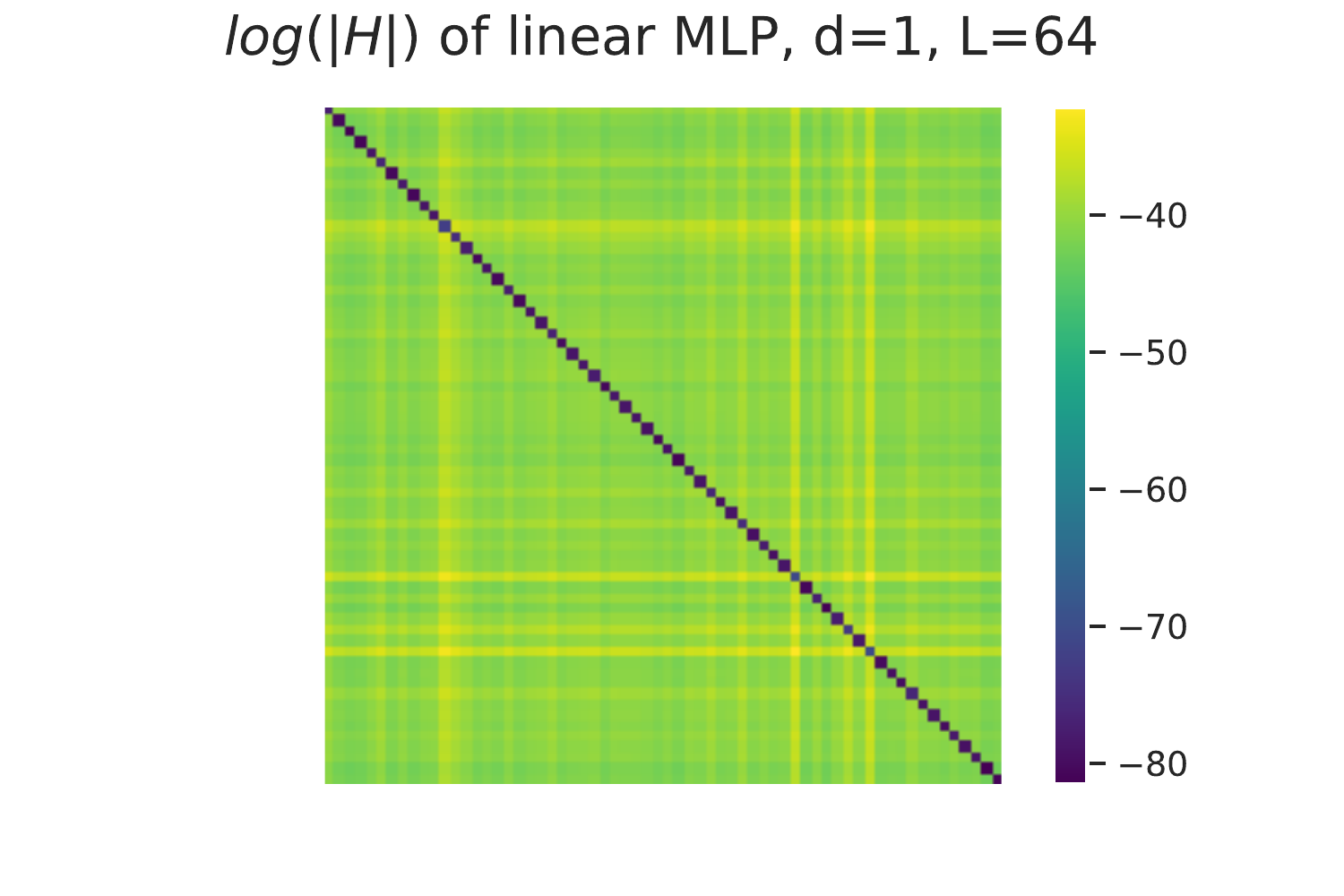}
\hspace*{0.5cm}\tiny{  }

\end{minipage}

\end{center}
\vspace{-3mm}
\caption{\small\textbf{Left:} Gradient and Hessian entry magnitudes for deep neural chains (Xavier init, Mean and 95$\%$ CI of 10 random seeds) \textbf{Middle, Right:} Eigenvalues and log Hessian entry maginute at init. for $L=64$.}\label{fig:chain}
\vspace{-4mm}
\end{figure}

In the neural chain case, if $\Am=\Bm=1$, Eq.\eqref{eq:setting} simplifies to $\Ls(\w)=\sum_{i=1}^{n}(y_i-w_L...w_1x_i)^2/(2n)$ and we consider each $w_i\in\mathbb{R}$ to be drawn uniformly at random in $[-\tau,\tau]$. Prop.~\ref{prop:glorot_he} suggests that both forward pass and gradient remain stable in magnitude when choosing $\tau=\sqrt{3}$. While this is true \textit{in expectation}, it is not the case when initializing individual models, where the expected value becomes an increasingly atypical event~(see Thm.~\ref{thm:as}) as the chain grows in depth ($L$). Indeed, in Fig.~\ref{fig:chain} we see that all quantities vanish under the ``optimal'' initialization. Perhaps the most intuitive indication for this pathological behavior comes from writing down the following population quantities for the absolute value of the input-output map.
\begin{tcolorbox}
\begin{proposition}[Forward pass statistics chain]
\label{prop:exploding_vanish}Consider the absolute value of a forward pass on the chain, i.e. the random variable $ v_{\tau,L} 
:= \prod_{k=1}^L (\tau w_k),$ with $w_k \stackrel{\text{iid}}\sim \mathcal{U}(0,1]$. Then, 
\vspace{-2mm}
\begin{equation}\label{eq:chain_moments}
\E\left[v_{\tau,L} \right] = \left(\cfrac{\tau}{2}\right)^L, \quad  \E[v_{\tau,L}^2]=\left(\cfrac{\tau^2}{3}\right)^L, \quad  \E[v_{\tau,L}^3]=\left(\cfrac{\tau^3}{4}\right)^L.
\end{equation}
\vspace{-2mm}
Clearly, $\tau=\sqrt{3}$ (Xavier init.) leads to $\E\left[v_{\tau,L} \right]\to 0$, $\E\left[v_{\tau,L}^2 \right]=1 $ and $\E[v_{\tau,L}^3]\to\infty$, as $L\to \infty$.
\end{proposition}
\end{tcolorbox}
It might be tempting to conclude from Eq.~\ref{eq:chain_moments} that picking $\tau=2$ instead of $\sqrt{3}$ solves the problem. Yet, this is not the case since then $\E[v_{\tau,L}^2]\to\infty$ and by Mallows inequality~\citep{mallows1969inequalities} the mean becomes an unreliable predictor for the median~(see Fig.~\ref{fig:dist}), as their difference is bounded by one standard deviation~(exploding). In fact, the above proposition reveals that one cannot stabilize any pair of moments of $v_{\tau,L}$ simultaneously and hence in both cases $\tau=\sqrt{3}$ and $\tau=2$, the distribution of $v_{\tau,L}$ becomes fat-tailed as $L\to\infty$, which leads to slow convergence of the central limit theorem\footnote{The speed of convergence in the CLT, as bounded by the Berry-Esseen inequality, is proportional to $\E[|v|^3]$.}. As we show in Sec.~\ref{sec:vanishing_curvature}, this basic moment trade off prevails in wider nets (see Eq.~\ref{eq:moments_wide}).

Therefore, one has to go beyond the population analysis in order to better understand this phenomenon. In a first step, we characterize the distribution of the magnitude of the input-output map in log scale.

\vspace{-2mm}
\noindent
\begin{minipage}[t]{0.7\textwidth}
\begin{tcolorbox}
\begin{lemma}[Distribution of input-output]
\label{lemma:dist}
In the setting of Prop.~\ref{prop:exploding_vanish},
\begin{align*}
-\ln(v_{\tau,L}) = z-L \ln(\tau), \quad 
z\sim \mathrm{Erlang}(L,1).
 \end{align*}
Hence $\prob(- \ln (v_{\tau,L})  \le \zeta)
=  1- e^{-\xi} \sum_{k=0}^{L-1} \frac{\xi^k}{k!}
$, $\xi := \zeta + L \ln \tau$.
\end{lemma}
\end{tcolorbox}
\end{minipage}\hfill
\begin{minipage}[t]{0.29\textwidth}
\includegraphics[width=\textwidth]{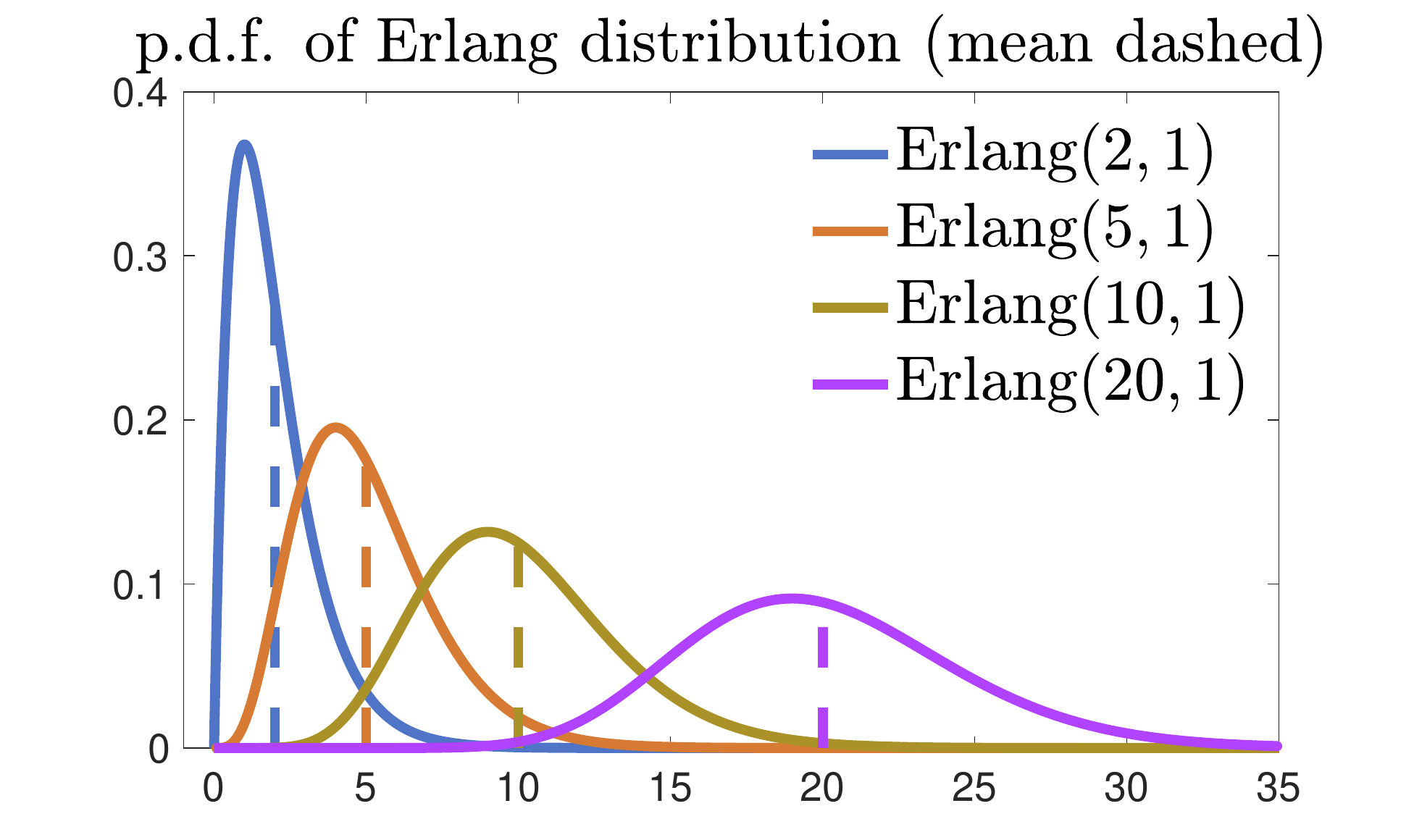}
\end{minipage}
\vspace{-2mm}
\begin{proof}
Clearly, $-\ln(w_k))$ is exponentially distribution with parameter $1$. Furthermore, if random variables $V_k\sim \mathrm{Exp}(\lambda)$ are independent, then  $\sum_{k=1}^{L}V_k\sim \mathrm{Erlang}(L,\lambda)$~\citep{temme1996special,devroye2006nonuniform}. The CDF follows from the properties of the Erlang distribution, which concludes the proof.
\end{proof}
\vspace{-2mm}
This allows to characterize the median and provide an asymptotic\footnote{In the context of asymptotic expansions, we write $f \sim g$ if $\lim_{L\to\infty} f(L)/g(L)=1$.} bound on the forward pass norm.

\begin{tcolorbox}
\begin{proposition}[Expectation is not predictive for input-output map magnitude] \label{lemma:slln}We have
\vspace{-1mm}
\begin{equation*}
\mathrm{median}\left[v_{\tau,L}\right] = e^{L\ln(\tau)-\tilde L}, \quad\quad \text{with } L-\nicefrac{1}{3}\le \tilde L\le L-1+\ln(2).
\end{equation*}
Therefore, if $\tau=2$, $\mathrm{median}\left[v_{\tau,L}\right]\to 0$ while $\E\left[v_{\tau,L}\right]=1$ and $\E\left[v_{\tau,L}^2\right]\to\infty$. For $\tau=\sqrt{3}$, also $\mathrm{median}\left[v_{\tau,L}\right]\to 0$.
Yet, the median is stabilized for $\tau = e$, since
\begin{align*} 
\lim_{L \to \infty} (v_{\tau,L})^{\frac 1L} \stackrel{a.s.}= \tau/e, \ \ \text{ which implies  }\quad v_{\tau,L} \sim \exp\left(-L(1-\ln \tau)\right).
\end{align*}
\end{proposition}
\end{tcolorbox}
\vspace{-2mm}
\begin{proof}
The moments follow from Prop.~\ref{prop:exploding_vanish}. For the median, we solve $\prob(- \ln v_{\tau,L}  \le \zeta) = 1/2$ for $\zeta$, which by  Lemma~\ref{lemma:dist} is equivalent to solving $1- e^{-\xi} \sum_{k=0}^{L-1} \frac{\xi^k}{k!} = 1/2$ w.r.t. $\xi := \zeta + L \ln \tau$. The solution, termed $\tilde L $, is approximated with a Ramanujan formula~(1913), as in~\citep{choi1994medians}. Since $v_{\tau,L} = e^{L \ln \tau - z}$, then $(v_{\tau,L})^{\frac 1L} = \tau e^{-z/L}.$ We conclude using the strong law of large numbers.
\end{proof}

\vspace{-2mm}
We now apply the idea behind the last result to analyze the first and second order partial derivatives.
\begin{tcolorbox}
\begin{restatable}[Almost sure vanishing]{theorem}{propas}
\label{thm:as}
Assume bounded data and $w_i\sim\mathcal{U}[-\tau,\tau]$, with fixed $\tau$. For each $k,\ell\le L$ we asymptotically~(as $L\to\infty$) have almost surely that
\vspace{-1mm}
\begin{align*}
   \left|\frac{\partial\Ls_{\text{chain}}(\w)}{\partial w_k}\right|,\left|\frac{\partial^2\Ls_{\text{chain}}(\w)}{\partial w_k\partial w_{\ell\ne k}}\right| &= \mathcal{O} \left(e^{-(L-1)(1-\ln \tau)}\right),\\
   \left|\frac{\partial^2\Ls_{\text{chain}}(\w)}{\partial w_k\partial w_{k}}\right| &= \mathcal{O} \left( e^{-2(L-1)(1-\ln \tau)}\right).
\end{align*}
In particular, as for $v_{\tau,L}$, all these quantities asymptotically vanish if $\tau<e$ and explode if $\tau>e$.\\
In the case of Xavier init. $\tau =\sqrt{3}$, the Hessian vanishes in norm~(hence eigenvalues vanish) and becomes hollow, i.e. diagonal elements become exponentially smaller than off-diagonal elements. 
\end{restatable}
\end{tcolorbox}
\vspace{-2mm}

The proof is presented in App. \ref{sec:proof}. Fig.~\ref{fig:chain} and Fig.~\ref{fig:many_widths_effect} (top row) show that the result is very precise.

\begin{figure}[ht]
    \centering
    \includegraphics[width=\textwidth]{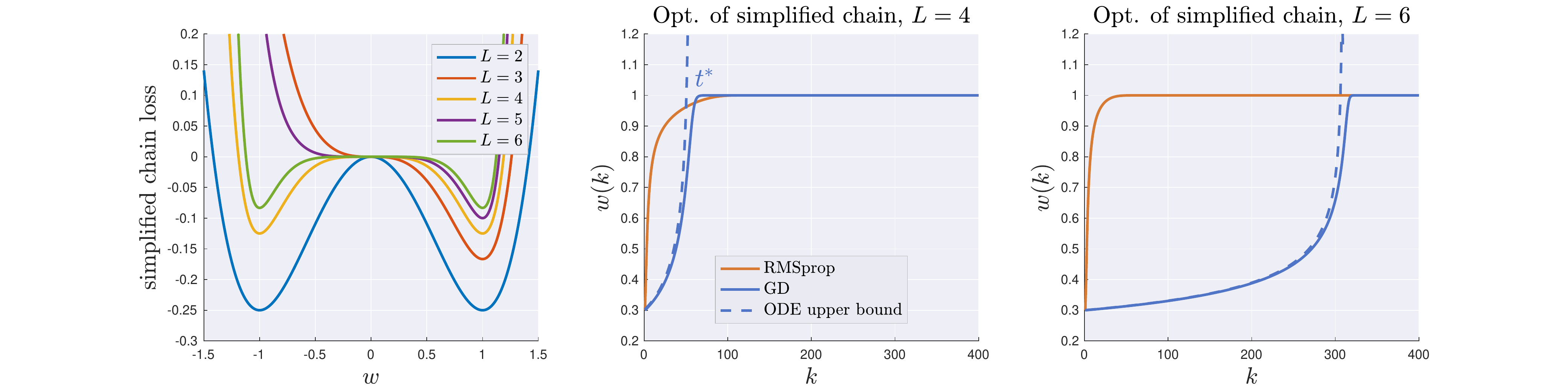}
    \vspace{-3mm}
    \caption{\small  Chain setting of Prop.~\ref{prop:GD_slow}. \textit{Fast convergence of RMSprop} with $\beta_2=0.9$ and stepsize decay. For GD, $\eta=0.1$ is used since \textit{bigger $\eta$ leads to instability}). Plotted is also the loss corresponding to the integrated gradient $w^{2L-1} + w^{L-1}$. For the discretizing the bound, we use the equivalence $\eta k \equiv t$.}
    \label{fig:simplified_chain}
    \vspace{-4mm}
\end{figure}

\vspace{-3mm}
\paragraph{Implications on landscape and optimization.} In narrow networks, our results show vanishing gradients and hollow Hessians with positive and negative eigenvalues of decreasing magnitude (also see Fig. \ref{fig:eigenvalues_over_depth}). Hence, the initialization landscape constitutes a plateau that resembles a barely curved saddle~(see Fig.~\ref{fig:simplified_chain}). As discussed next, this is particularly bad for optimization with plain SGD but adaptive methods escape the plateau quickly due to a notable curvature adaptation capability.

To illustrate this point, we consider a single datapoint pair $(x,y)$ with $x,y>0$ and study the gradient flow on a neural chain of depth $L$ with initialization $0<w_1(0)=w_2(0)=\cdots=w_L(0):= w_0\in\R$. The gradient $\nabla_{w_i} \Ls(\w) = \prod_{j\ne i}w_j\left(\prod_{r}w_r x-y\right)$, is invariant w.r.t. any permutation of the $w_i$'s. Hence, each coordinate of the gradient flow solution will satisfy $w_1(t)=w_2(t)=\cdots=w_L(t)=: w(t)$ and $w(t)\to w^*=(y/x)^{1/L}$ (as $L\to\infty$). The gradient flow is $\dot w(t)=-w(t)^{2L-1} x + w(t)^{L-1}y$. To simplify this we drop the first term (negative) and hence get an upper bounding solution (since $w(t)$ is increasing) which explodes in finite time $t_e$~(see Fig.~\ref{fig:simplified_chain}):
\begin{equation}
w(t) \le \left[(L-2)(t_e-yt)\right]^{-\frac{1}{L-2}},\quad t_e = w_0^{2-L}/(L-2).
\end{equation}
To investigate the consequences of this upper bound, we can take the special case $x=y$, which leads to $w^*=1$. In this case, the upper bound for $w(t)$ reaches $w^*=1$ at time $t^* = t_e-\frac{1}{L-2}$, which is exponential in the network depth $L$. This provides a proof for the following proposition.

\begin{tcolorbox}
\begin{proposition}[Slow convergence of Gradient Flow on the chain]
\label{prop:GD_slow}
On neural chains, in the worst case, gradient flow takes exponential (in depth) time to reach an $\epsilon$-neighbour of the solution.
\end{proposition}
\end{tcolorbox}

\paragraph{Curvature Adaptation of RMSprop.}
As can be seen in Fig.~\ref{fig:simplified_chain}, RMSprop~\citep{tieleman2012lecture} is able to optimize the neural chain, in a number of iterations independent of the network depth. Importantly, this finding is also observable in wider MLPs (Fig.~\ref{fig:train_mlp}) and deep convnets (Fig.\ref{fig:train_resnet}).

To provide some intuition around this phenomenon, we apply RMSprop to the neural chain gradient flow approximation $\dot w(t)=w(t)^{L-1}$. This gradient flow approximation is tight during the first steps of the optimizer if $L$ is big. The RMSprop flow then solves $\dot w(t)=w(t)^{L-1}/\sqrt{v(t)}$, where $v(t)$ is a low-pass filter on the approximate square gradient $w(t)^{2L-2}$. Since $w(t)^{L-1}$ is increasing, $v(t)$ is also increasing and the filter delay guarantees $\sqrt{v(t)}\le w(t)^{L-1}$. It directly follows that $\dot w(t)>1$ for $t$ small, regardless of the network depth, which allows RMSprop to quickly escape the flat plateau.

Given the vanishing curvature setting predicted by Thm.~\ref{thm:as} and confirmed in Fig.\ref{fig:chain} \&~\ref{fig:simplified_chain}~(leftmost plot), this speedup is not extremely surprising. Indeed, as we discuss thoroughly in App.~\ref{app:RMS}, many recent works report an improved curvature adaptation of adaptive methods compared to SGD~\citep{dauphin2015equilibrated,kunstner2019limitations}. For instance,~\cite{staib2019escaping} recently showed that RMSprop is provably faster than SGD around flat saddle points~(see Section 5.2 of their paper). This result, \textit{in combination with our findings on the flatness of the initialization landscape}, gives an explanation for the historical difficulties of training deep nets with SGD and for the success of adaptive methods.
\vspace{-3mm}
\paragraph{Effect of noise.} While it is known that the inherent sampling noise of SGD is anisotropic and in many settings aligned with negative curvature \citep{daneshmand2018escaping,zhu2018anisotropic,li2020hessian}, the saddle escape time still depends inversely on the magnitude of the smallest eigenvalue \citep{daneshmand2018escaping,curtis2019exploiting}. As a result, similar to gradient flow on the chain, SGD is unable to train networks with vanishing gradients/curvature despite the presence of inherent noise (see Fig.~\ref{fig:train_mlp} \& \ref{fig:train_resnet}). Another possibility is to directly add noise to the updates \citep{du2017gradient,du2019gradient}. In Fig.~\ref{fig:RMSprop_app_d10_true_1}-\ref{fig:RMSprop_app_d10_true_3}, we provide evidence that noise can indeed accelerate GD on the chain, but it is still orders of magnitude slower than RMSprop, for any noise level and stable learning rate.

\section{Vanishing in MLPs of arbitrary width}\label{sec:vanishing_curvature}
\vspace{-2mm}
In analogy with Prop.~\ref{prop:exploding_vanish}, we first note~(proof in App.~\ref{sec:proof}) the important fact that also in the general MLP case different population quantities cannot be jointly stabilized using standard i.i.d. initialization. 

\begin{tcolorbox}
\begin{proposition}[Forward pass statistics MLP] \label{prop:moments_wide} Let $\kappa = \mu_4/\sigma_4$ be the kurtosis~(fourth standardized moment).
Let $p=1$ in the linear case and $p=1/2$ in the ReLU case. Then we have
\begin{align}
\E\|\wMp{k:1}\Am\x\|_2^2 &= (d\sigma^2 p)^k \E\|\Am\x\|^2_2.\\
\begin{pmatrix}\E\|\wMp{k:1}\z\|_2^4\\\E\|\wMp{k:1}\z\|_4^4\end{pmatrix} &= 
    \left(p^2 d\sigma^4\right)^k 
    \Qm^k \begin{pmatrix}\E\|\Am\x\|_2^4\\\E\|\Am \x\|_4^4\end{pmatrix},\quad 
    \Qm:=\begin{pmatrix}
    d+2&\frac{\kappa-3+(1-p)(d+2)}{p}\\ 3 & \frac{\kappa-3p}{p}
    \end{pmatrix}.
\end{align}
\end{proposition} 
\end{tcolorbox}

For deep linear nets of arbitrary width $d$ and Gaussian initialization~($\kappa =3$) the above simplifies to
\begin{equation}\label{eq:moments_wide}
\E\|\wM{L:1}\Am\x\|_2^2 = (d\sigma^2)^L \E\|\Am\x\|^2_2 \quad    \text{and} \quad \E\|\wM{L:1}\Am\x\|_2^4= \left(d\sigma^4\right)^L \left(d+2\right)^L\E\|\Am\x\|^4_2.
\end{equation}
Hence, as for the neural chain~(see Prop.~\ref{eq:chain_moments}), picking the Xavier initialization $\sigma^2=\tfrac{1}{d}$ stabilizes $\E\|\wM{L:1}\Am\x\|_2^2$, but $\E\|\wM{L:1}\Am\x\|_2^4=\left(\frac{d+2}{d}\right)^L\E\|\Am\x\|_2^4$ explodes \textit{unless} $d$ grows faster than $L$. This points to an important shortcoming of the initialization proposed in \cite{glorot2010understanding} \& \cite{he2015delving}, which --- as we note next --- is only guaranteed to prevent vanishing gradients and curvature in networks that are wider than deep. The next result is verified in Fig.~\ref{fig:many_widths_effect} in the appendix.
\begin{tcolorbox}
\begin{theorem}\label{thm:d_inf}
The initialization in  \cite{glorot2010understanding} \& \cite{he2015delving} is guaranteed to stabilize both the mean and the median of the squared forward pass norm for $d=\mathcal{O}(L)$.
\end{theorem}
\end{tcolorbox}
\vspace{-3mm}
\begin{proof}We carry out the proof for the Gaussian linear case, but the other settings are conceptually equivalent. First, in the case $\sigma^2= 1/d$, we have
\vspace{-3mm}
\begin{equation}
    \textrm{Var}\|\wM{L:1}\Am\x\|_2^2 = \E\|\wM{L:1}\Am\x\|_2^4 -\left( \E\|\wM{L:1}\Am\x\|_2^2\right)^2 \overset{Eq.~\eqref{eq:moments_wide}}{=} \left(\frac{d+2}{d}\right)^L- 1.
\end{equation}
By Mallows inequality~\citep{mallows1969inequalities}, the square root of this quantity bounds
$\left|\text{median}\left(\|\wM{L:1}\Am\x\|_2^2\right)-\E\|\wM{L:1}\Am\x\|_2^2\right|$. If we want to guarantee a \textit{non-vanishing median}, say in $[1-\alpha,1+\alpha]$, for $1>\alpha>0$, we need to have $d$ s.t. $\left(\frac{d+2}{d}\right)^L -1 \le \alpha^2$ which implies $d\ge\frac{2}{(\alpha^2+1)^{\frac{1}{L}}-1}$. For $L\to\infty$ we have $(\alpha^2+1)^{\frac{1}{L}} = 1+\ln(\alpha^2+1)/L +\mathcal{O}(1/L^2)$, thus we conclude $d\ge \mathcal{O}(L)$.
\end{proof}
\vspace{-2mm}
Before discussing the regime $d\ll L$, we first consolidate our core claim about vanishing curvature by generalizing the results of \citep{glorot2010understanding,he2015delving} to second-order derivatives.
\vspace{-1mm}

\begin{tcolorbox}
\begin{theorem}[Gradient and Hessian in expectation]\label{thm:H_thm}
Under Assumption \ref{ass:init}, the \textbf{expected} norm of any Hessian diagonal block $
\E [\| \tfrac{\partial^2 \Ls(\Wm)}{\partial \Wm_k \partial \Wm_k} \|_F]$ in linear networks scales as $\mathcal{O}\left((d\sigma^2)^{L}\right)$, while off-diagonal blocks $
\E [\| \tfrac{\partial^2 \Ls(\Wm)}{\partial \Wm_k \partial \Wm_\ell} \|_F]$ as well as the gradient $
\E [\| \tfrac{\partial \Ls(\Wm)}{\partial \Wm_k} \|_F]$ scale as $\mathcal{O}\left((d\sigma^2)^{\frac{L}{2}}\right)$. In ReLU networks the scaling amounts to $\mathcal{O}\left(\left(\frac{d}{2}\sigma^2\right)^{L}\right)$ and $\mathcal{O}\left(\left(\frac{d}{2}\sigma^2\right)^{\frac{L}{2}}\right)$ respectively. 
\end{theorem}
\end{tcolorbox}
This result is (to the best of our knowledge) the first to study the effects of depth on second-order derivatives at random initialization. Its proof, which mainly builds upon Prop.~\ref{prop:moments_wide}, can be found in App.~\ref{sec:proof}. A simple application of Gershgorin's theorem yields the following bound on the eigenvalues.
\begin{tcolorbox}
\begin{corollary}\label{corr:ev}
Under Assumption \ref{ass:init}, the \textbf{expected} magnitude of the largest eigenvalue $\lambda_{\max}$ is upper bound as  
$\E [|\lambda^{\text{linear}}_{\max}|]\leq Ld\cdot \mathcal{O}\left((d\sigma^2)^{\frac{L}{2}}\right)$
and
$
\E [|\lambda^{\text{ReLU}}_{\max}|]\leq Ld\cdot\mathcal{O}\left((\frac{d}{2}\sigma^2)^{\frac{L}{2}}\right)
$
respectively.
\end{corollary}
\end{tcolorbox}
\vspace{-3mm}
\begin{figure}[ht]
\centering
\includegraphics[width=0.33\textwidth]{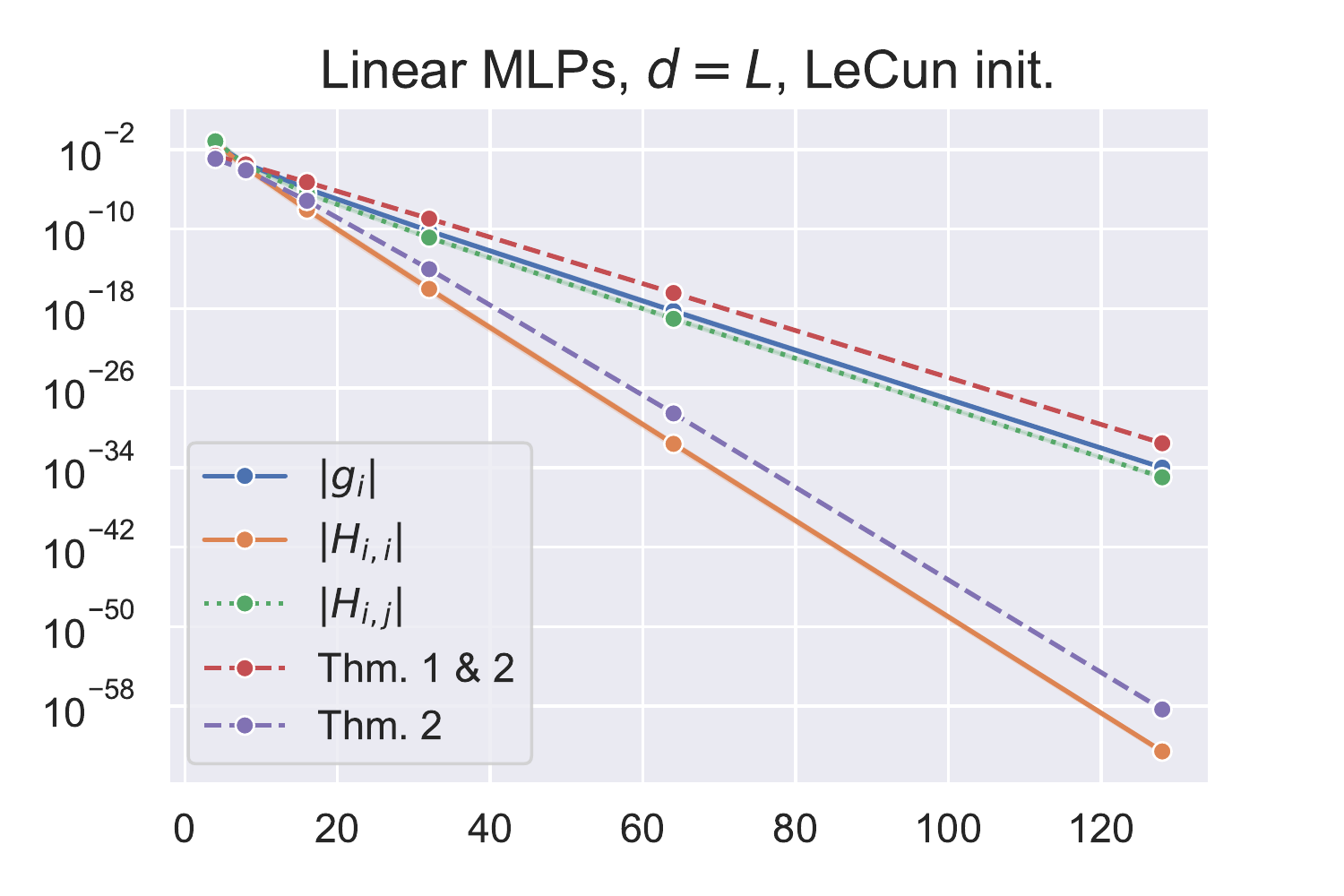}\hfill
\includegraphics[width=0.33\textwidth]{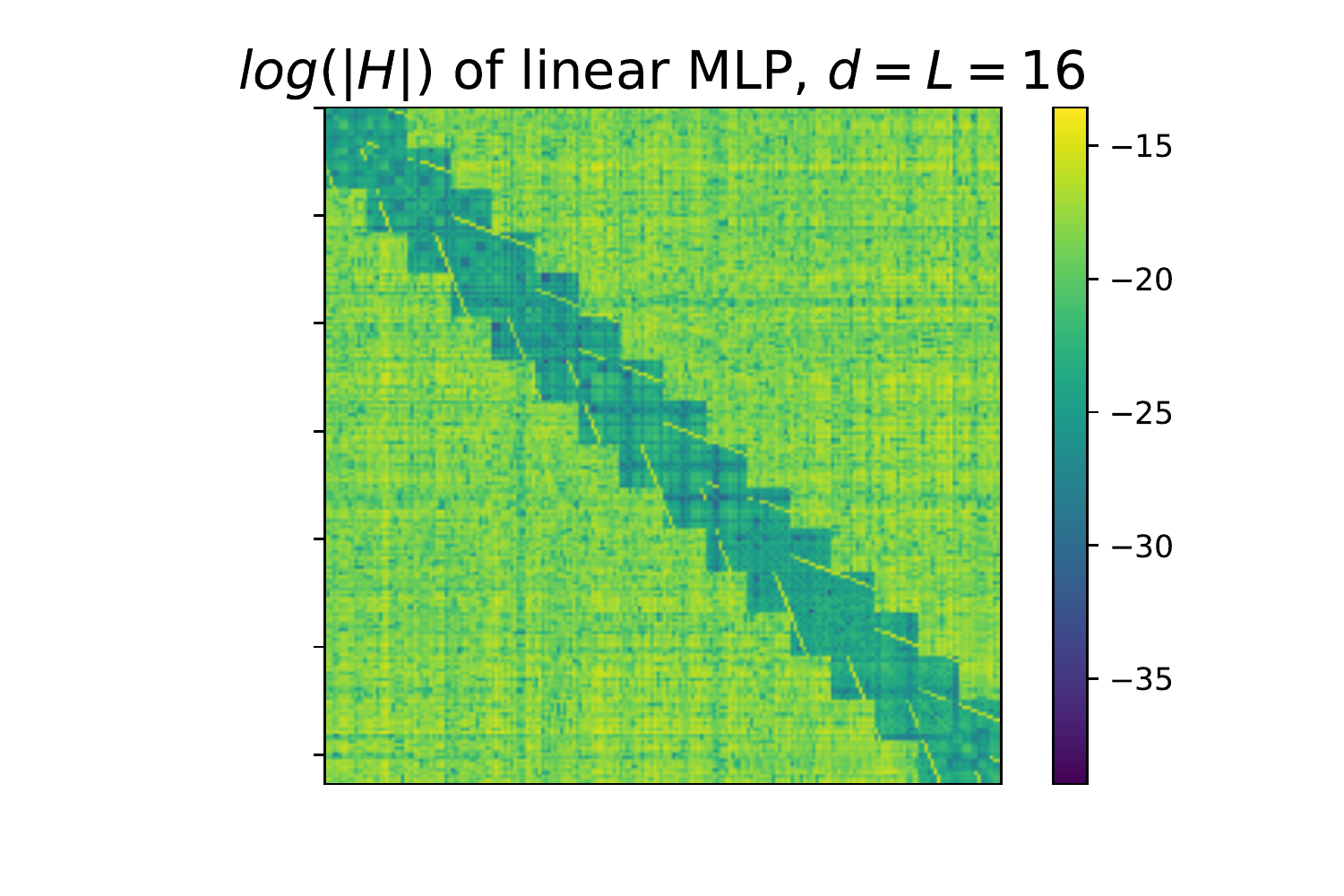}\hfill
\includegraphics[width=0.33\textwidth]{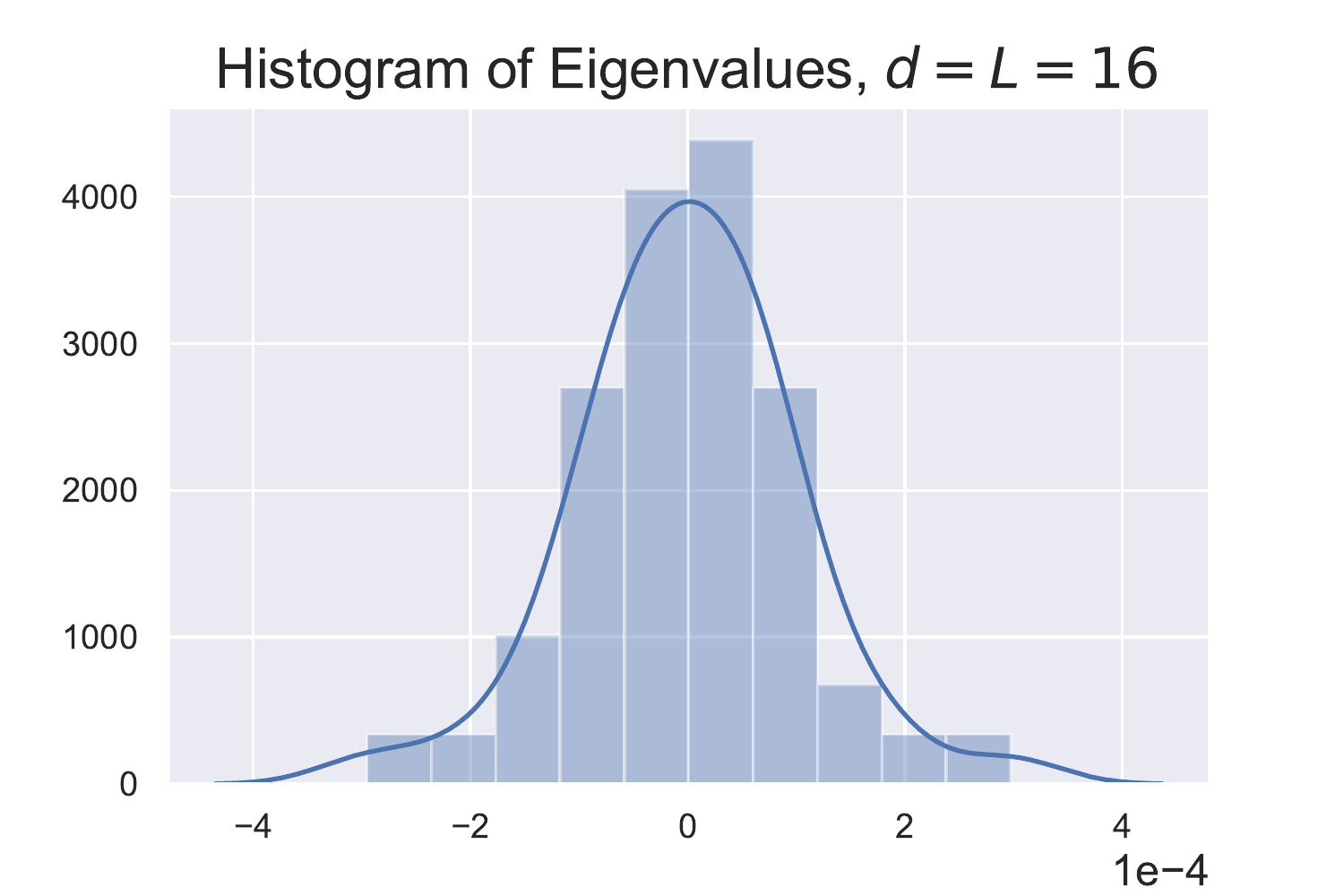}
\vspace{-3mm}
\caption{\small \textbf{Left:} Vanishing gradient and Hessian for deep linear MLPs with LeCun init. x-axis depicts depth \textit{and} width. \textbf{Middle/Right}: log abs. Hessian entries and eigenvalues at random init. (see ReLU MLP in Fig. \ref{fig:width_effects_mlp})}\label{fig:figure1_lin_1}
\vspace{-3mm}
\end{figure}

\paragraph{Consequences for traditional LeCun init.} The above results point to an important consequence of the standard way of initialization prior to \citep{glorot2010understanding}. When choosing $\sigma^2=\tfrac{1}{3d}$ as in LeCun init. \citep{lecun2012efficient}, gradient- and Hessian off-diagonal norms in (e.g.) linear nets vanish as
$\mathcal{O}( (\tfrac{1}{3})^{L/2})$ and Hessian diagonal blocks vanish even faster, namely at $ \mathcal{O}((\tfrac{1}{3})^{L})$ (Fig. \ref{fig:figure1_lin_1} \& \ref{fig:figure1_relu}). This points to an important fact about the eigenspectrum of the Hessian. Since,  $\E\|\nabla^2\Ls(\Wm)\|_F$ scales as $(d\sigma^2)^{\frac{L}{2}}$ by Thm.~\ref{thm:H_thm}, one of the $Ld^2$ eigenvales must have a magnitude $(d\sigma^2)^{\frac{L}{2}}$ in expectation. Yet, the fast diagonal vanishing lets the trace scale as $(d\sigma^2)^{L}$. As a result, the sum of the eigenvalues is exponentially smaller than the maximum eigenvalue and hence there must be eigenvalues of opposite sign (see e.g. Fig.~\ref{fig:figure1_lin_1}). In summary, as in neural chains, optimizers are initialized in a flat plateau with almost no gradient signal and both positive as well as negative, but very small eigenvalues.

\begin{figure}[ht]
\includegraphics[width=.33\textwidth]{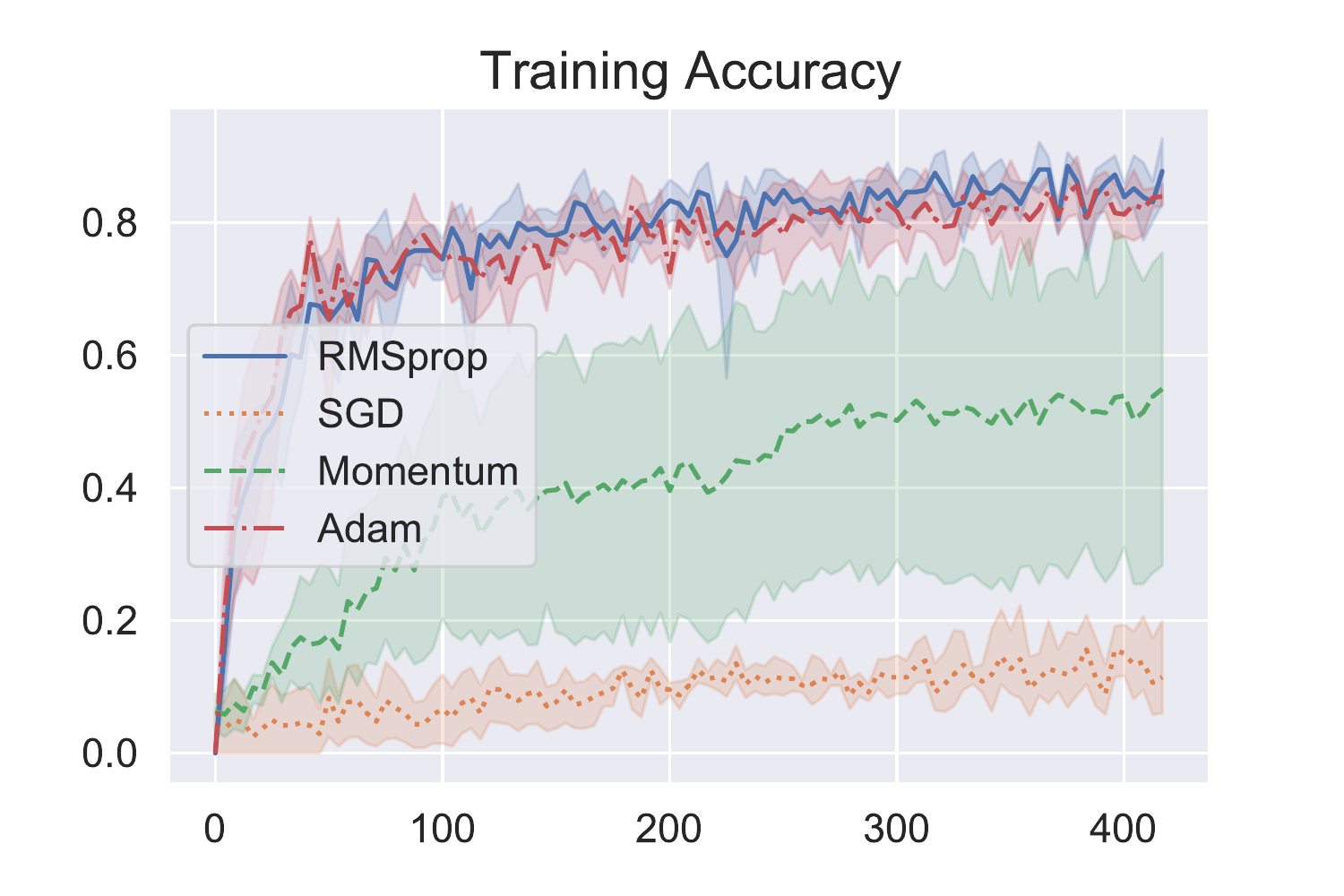}\hfill
\includegraphics[width=.33\textwidth]{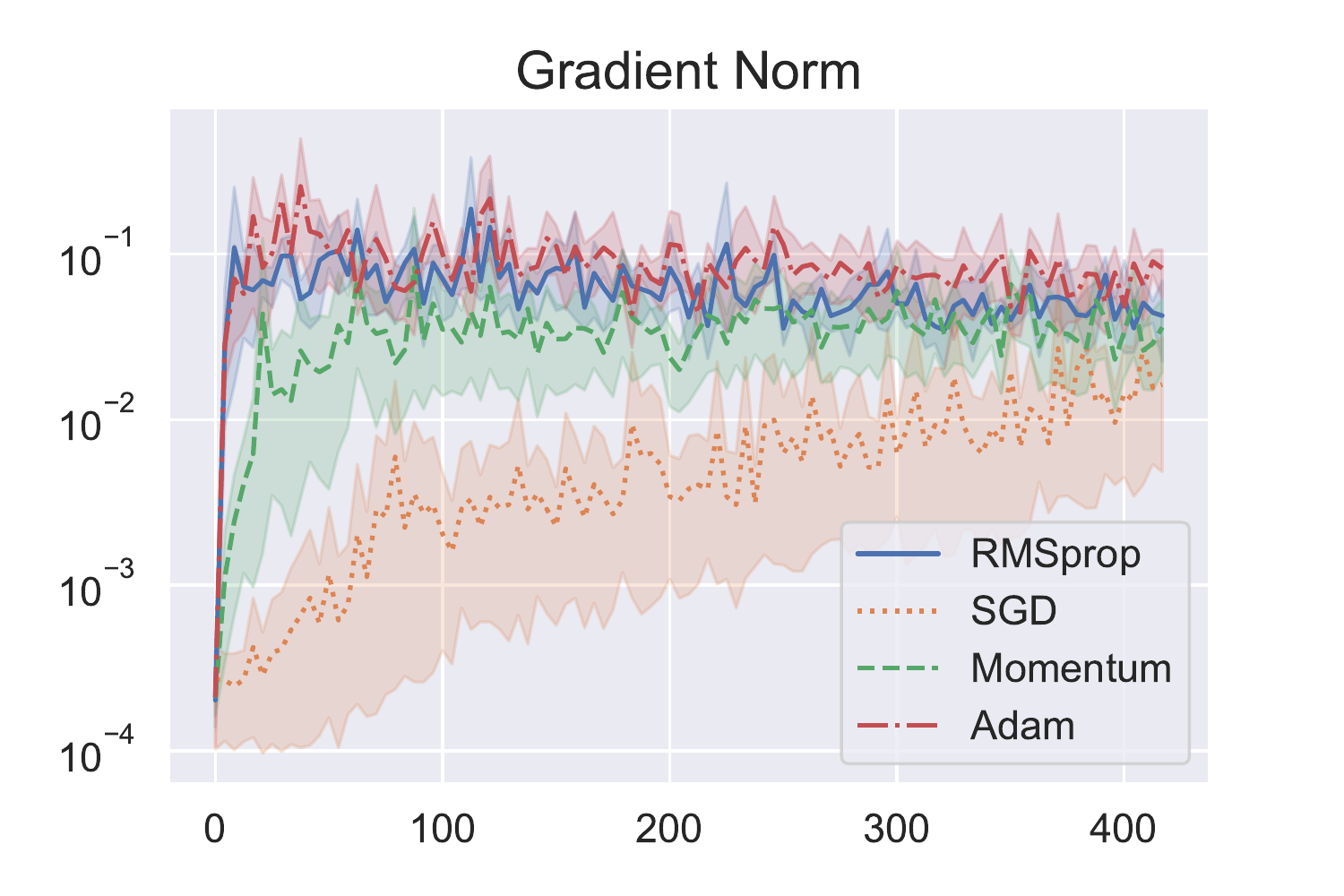}\hfill
\includegraphics[width=.33\textwidth]{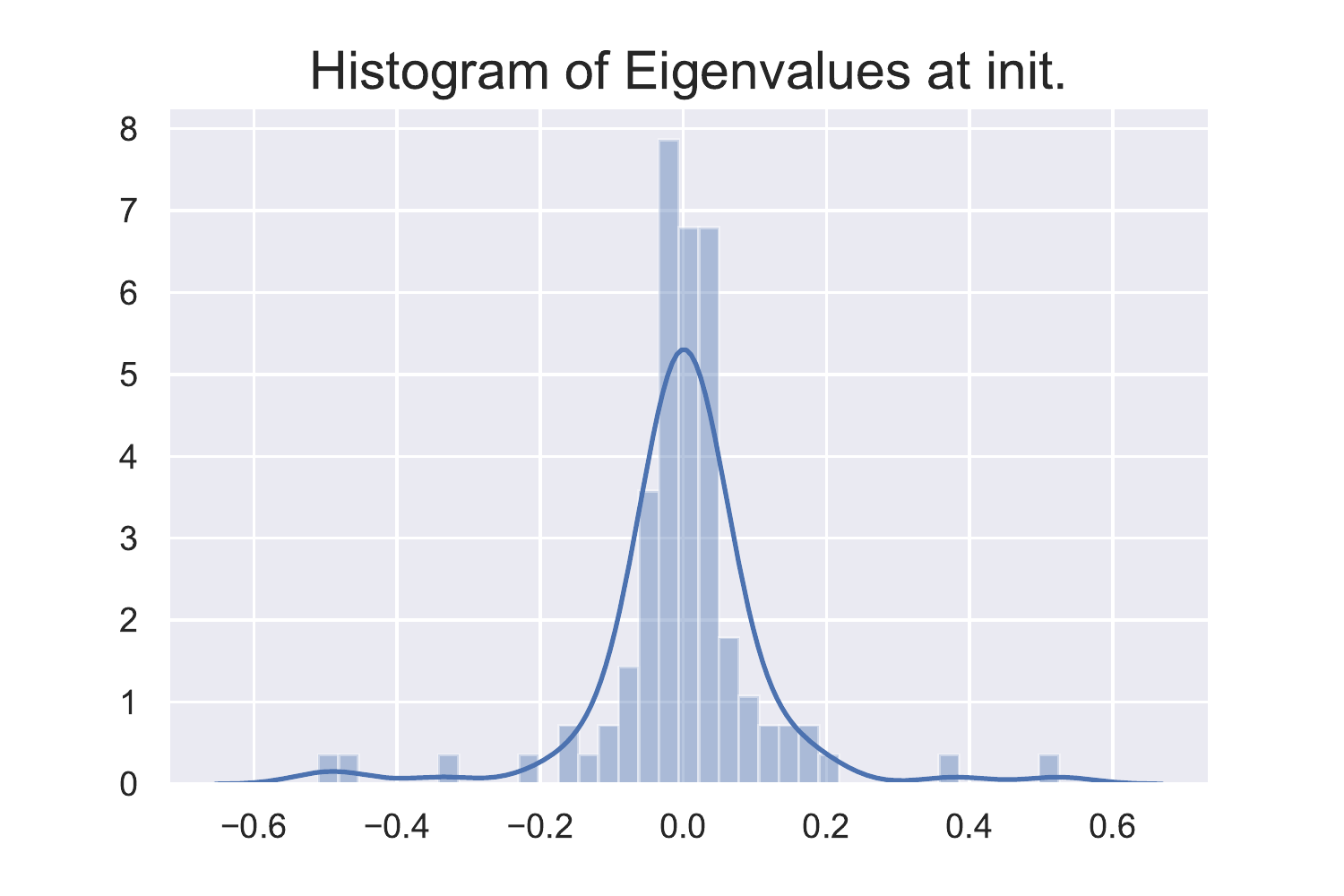}
\vspace{-2mm}
\caption{\small Fashion-MNIST on a narrow (32 hidden units) 128 layer ReLU MLP with \textit{He init}. Mean and 95$\%$ CI of 10 random seeds. See App. \ref{app:hyperparams} for hyperparameters and Fig.~\ref{fig:test_accs} for test accuracy. Note that RMSProp successfully trains despite the fact that both gradients and curvature vanished (also compare Fig.~\ref{fig:width_effects_mlp}). Yet, as can be deduced from how gradient norms evolve over time, SGD struggles to escape the flat plateau which is not surprising given that the negative eigenvalues are very small (compare Fig.~\ref{fig:fm_apx_wide} for a wide version of this network.)}
\vspace{-3mm}
\label{fig:train_mlp}
\end{figure}

\paragraph{Consequences for Xavier and He init.} 
Thm.~\ref{thm:d_inf} \& \ref{thm:H_thm} provide simple theoretical grounding for the benefits of increasing width in random neural networks.\footnote{\citep{hanin2019products,allen2019convergence} come to similar conclusions albeit with more complex analysis.} However, they also point out an important limitation in the analysis and applicability of \citep{glorot2010understanding,he2015delving}. What happens in cases where width scales sub-linearly with depth remains open for theoretical analysis. 

Our study on the neural chain~(Thm.~\ref{thm:as}) along with Thm.~\ref{thm:d_inf} suggests that no initialization variance $\sigma^2$ is capable of stabilizing the full \textit{distribution} of all of activation-, gradient- and the Hessian norm when the width is small. Yet, characterizing the almost sure behaviour of the gradient and Hessian for networks with $d>1$ is intricate since paths in fully connected networks overlap, such that one cannot treat them as a set of independent products of random variables, which would allow a straight forward generalization of Thm.~\ref{thm:as} using Berry-Esseen inequality~\citep{berry1941accuracy}. Similarly, deriving the distribution of the forward pass is very challenging. In fact, already in the scalar case (neural chain), product distributions for both uniform \citep{dettmann2009product} and Gaussian initialization become very complex \citep{springer1970distribution}. We thus retreat to empirical simulations, but stress the fact that these are informative because they are undertaken in a controlled setting, where the only source of randomness (weight initialization) is well controlled by running multiple seeds.

Our empirical results highlight that $\sigma^2=\frac{1}{d}$ is indeed not an optimal choice for narrow networks. Indeed, as can be seen in Fig.\ref{fig:width_effects_mlp}, gradients and curvature vanish in narrow but stay stable in wide ReLU MLPs with He init. The same happens in linear networks with Xavier init (Fig. \ref{fig:width_effects_mlp_lin}). Again, we find both negative and positive eigenvalues at initialization (Fig. \ref{fig:eigenvalues_over_depth}). Interestingly, Fig.~\ref{fig:many_widths_effect} depicts that the optimal initialization variance $\sigma^2$ is a function $g(d,L)$ of the width to depth ratio. This function is highly non-trivial as it depends not only on the number of paths but also on their mutual overlap. In the following section, we show that these considerations extend to convolutional neural networks. %

\section{Vanishing in convolutional networks}\label{subsec:conv}
\begin{figure}[h]
 \begin{center}

\begin{minipage}[b]{\figsize\linewidth}
\centering
\includegraphics[width=1.05\textwidth]{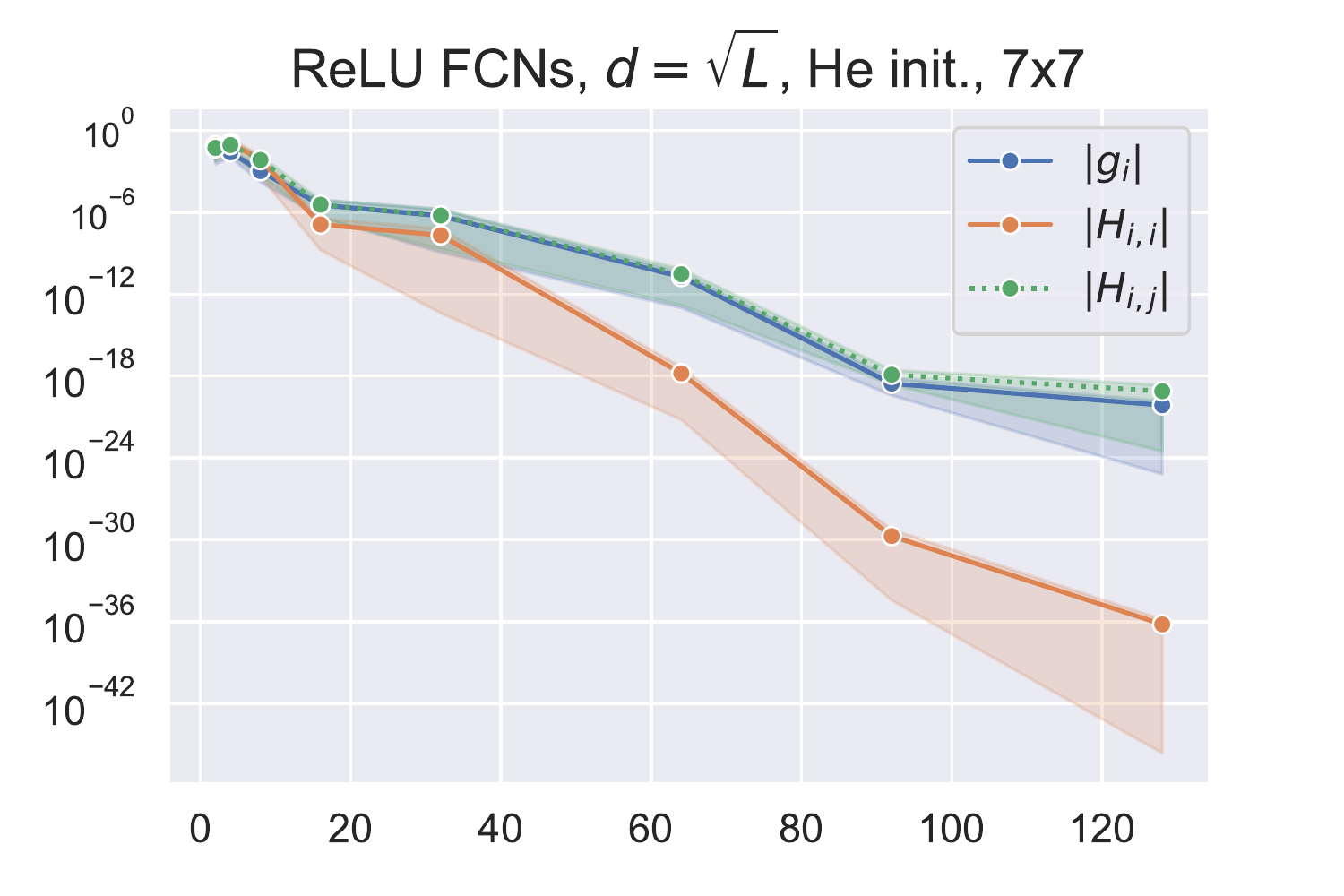}
\hspace*{0.75cm}\tiny{depth}

\end{minipage}
\hspace{0.005cm}
\begin{minipage}[b]{\figsize\linewidth}
\centering
\includegraphics[width=1.05\textwidth]{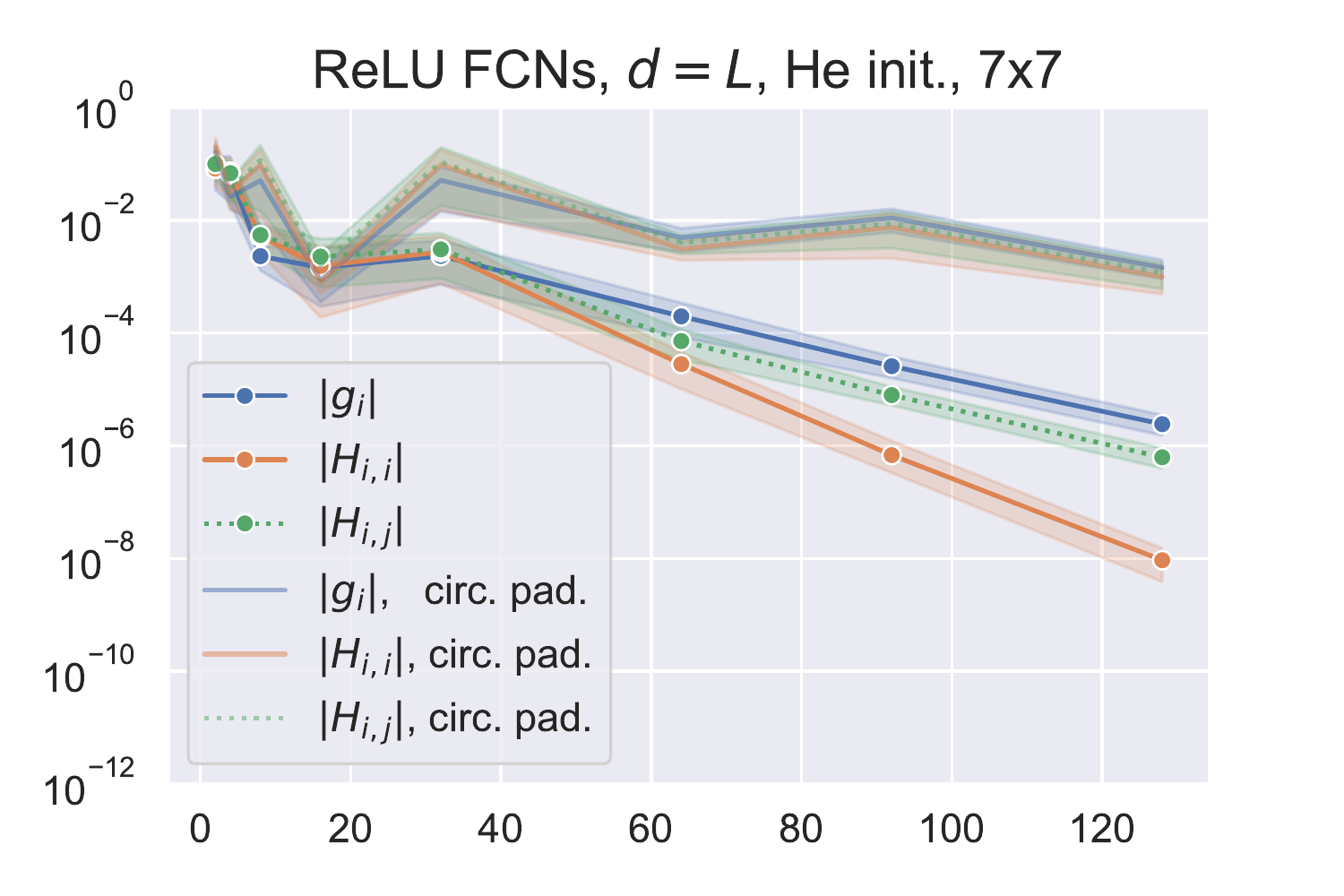}
\hspace*{0.75cm}\tiny{depth}
\end{minipage}
\end{center}
\vspace{-3mm}
\caption{\small \textbf{Effect of width in CNNs:} Gradient/curvature on FCNs over depth. Plots has different scales. Mean and 95\% CI of 15 runs. While increasing width helps for large resolution (see Fig.~\ref{fig:width_effects_cnn_apx}), for small resolutions such as 7x7 (above) vanishing occurs even when width scales as $\mathcal{O}(L)$ (albeit at slower rate (see y-axis)). Interestingly, replacing zero- with circular padding mitigates this effect.}
\label{fig:width_effects_cnn}
\vspace{-3mm}
\end{figure}

\vspace{-3mm}
Next, we discuss the effects of increasing width in convolutional architectures. For simplicity, we consider an image to image learning setting with images of resolution $r\times r$, using fully convolutional networks (FCNs) with $k\times k$ filters (where $k$ is odd), $c$ channels in each layer and a padding of $(k-1)/2$ (such that the resolution does not change over depth). We first note that width is not as straight-forward to define in CNNs as it is in MLPs. In Sec. \ref{sec:width}, we argue that it is the number of paths leading to an output neuron (as well as their level of overlap) that determines its activation magnitude. While there are $|\Gamma|=d^L$ paths in MLPs, the FCN architecture yields $|\Gamma|=(k^2c)^L$ (assuming non-zero padding). Consequently, we define the effective network width as $d:=k^2c$. Obviously, there are then two ways to increase width, namely via the filter size $k$ and the number of channels $c$. As in prior theoretical \citep{arora2019exact} and practical works \citep{zagoruyko2016wide}, we focus on the latter option. %
\vspace{-3mm}
\paragraph{Empirical findings.} We first consider the above introduced image to image learning setting with CIFAR-10 images as inputs and targets, once at the original $32\times32$ resolution and once downsized to $7\times7$. We fix the kernel size $k:=3$ and increase both the number of layers and channels. As can be seen in Fig.\ref{fig:width_effects_cnn}, gradients/curvature vanish when depth grows faster than width and again the Hessian diagonal decreases fastest (top). Due to weight sharing, the effect is more pronounced as in MLPs (see App. \ref{sec:width_cnns} for discussion). When width grows linearly in depth (bottom), however, the beneficial effect is reduced when operating at small resolutions, where the pixel to padding ratio goes down such that zero-padding has a detrimental effect on the gradient/Hessian magnitude over depth. As indicated on the bottom right, this effect can be circumvented by opting for non-zero padding (circular in this case).\looseness=-1

\begin{figure}[ht]
\begin{center}
\begin{minipage}[b]{\figsize\linewidth}
\centering
\includegraphics[width=1.05\textwidth]{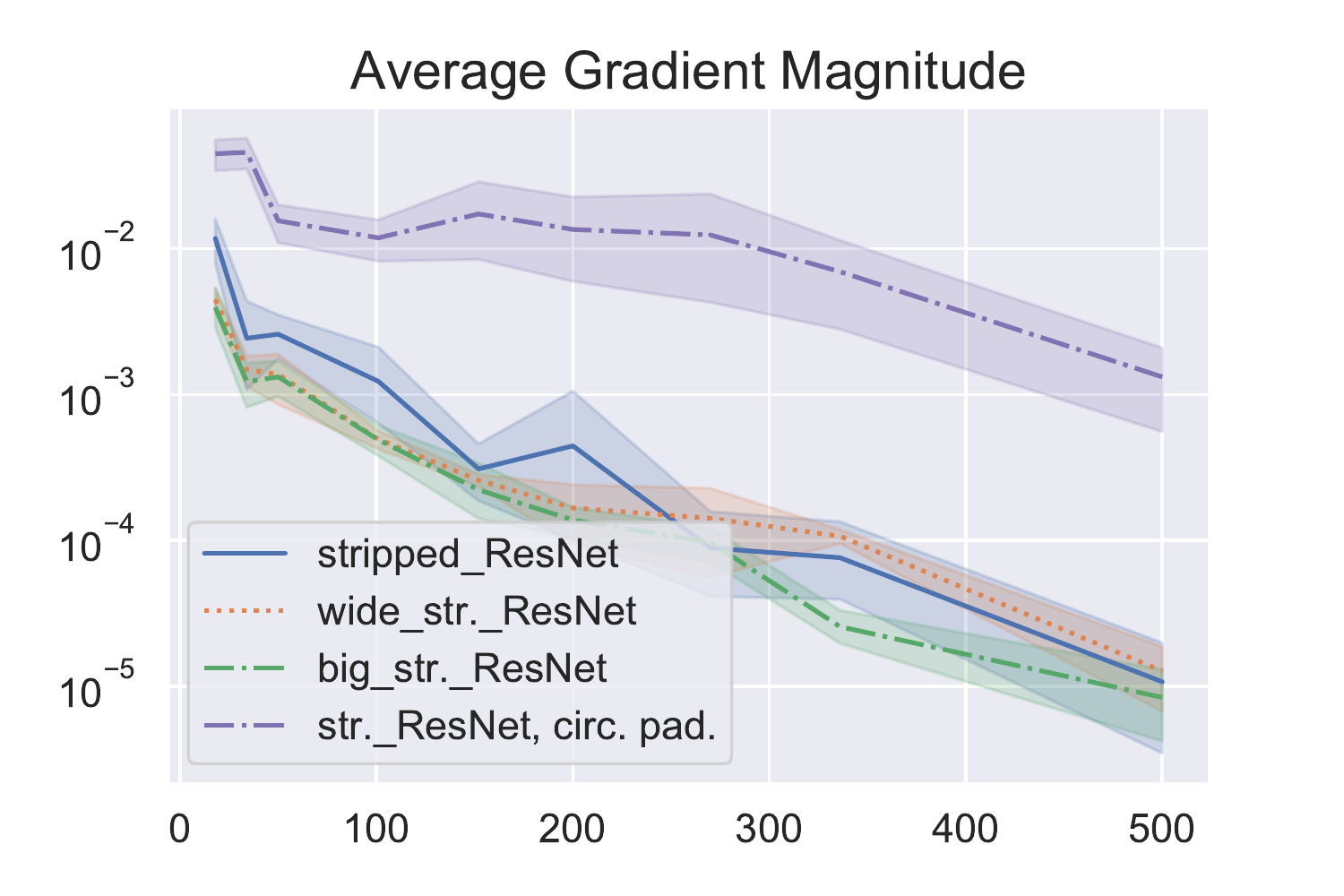}
\hspace*{0.5cm}\tiny{depth}
\end{minipage}
\hspace{0.005cm}
\begin{minipage}[b]{\figsize\linewidth}
\centering
\includegraphics[width=1.05\textwidth]{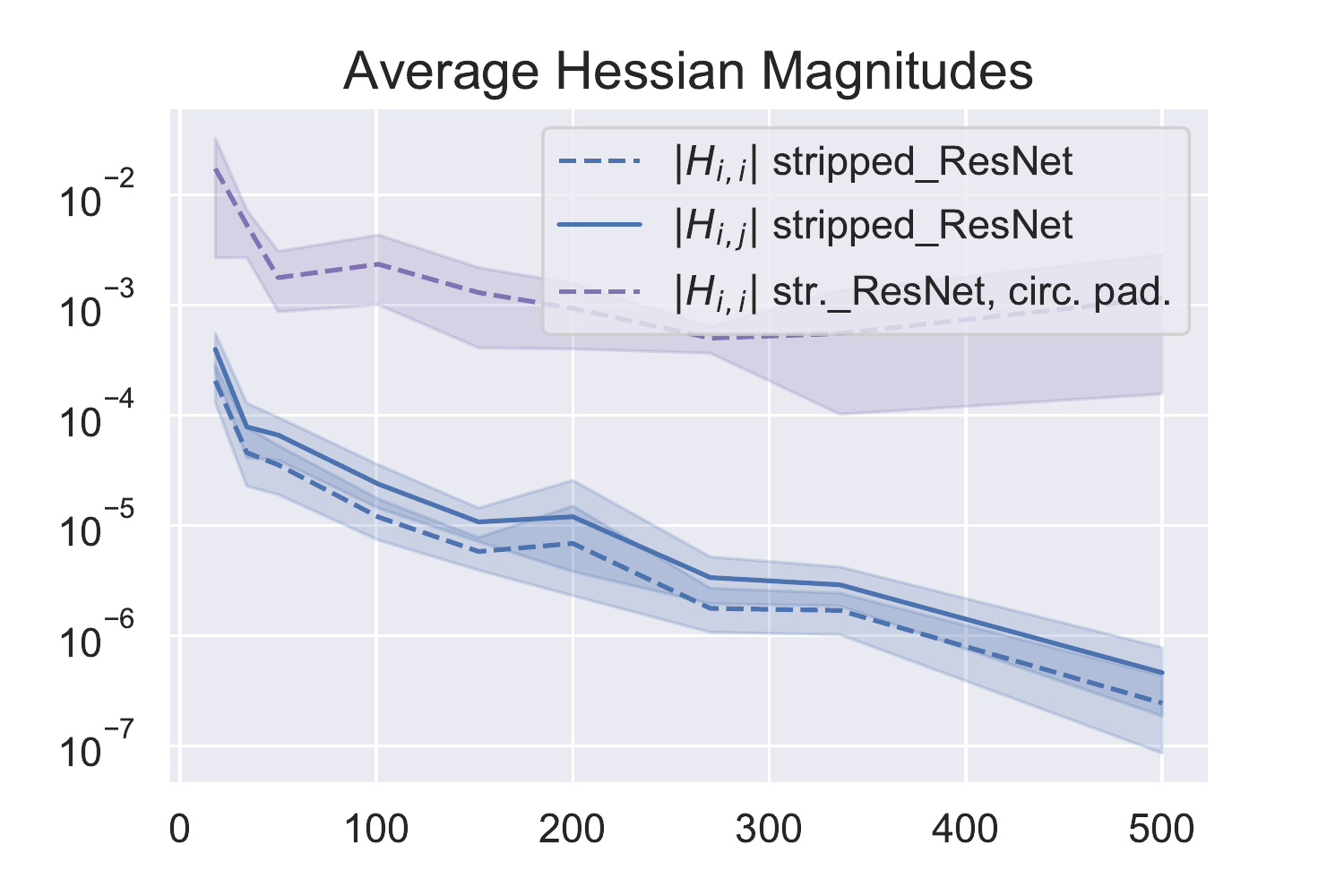}
\hspace*{0.5cm}\tiny{depth}
\end{minipage}
\end{center}
\vspace{-3mm}
\caption{ \small \textbf{Left:} Gradient in \textit{stripped\_ResNets} with He init. on CIFAR-10. Increasing number of channels (\textit{wide\_stripped\_ResNet}) and kernel size (\textit{big\_stripped\_ResNet}) does not help. Yet, as in Fig. \ref{fig:width_effects_cnn}, circular padding slows the effect down. \textbf{Right:} Average Hessian entry magnitudes in stripped ResNet (wide and big not computed due to memory limitations). Mean and 95\% CI of 10 runs.\looseness=-1 }
\vspace{-3mm}
\label{fig:cnn_vanishing}

\end{figure}

At first, $7\times7$ may seem unrealistically small but we remark that even when training on $224\times 224$ ImageNet inputs, the last block of layers in ResNets indeed operates exactly at this scale \citep{he2016deep}. Consequently, vanishing also occurs in ResNet-type architectures, despite their large number of channels (Fig. \ref{fig:cnn_vanishing}). Here, similarly to \cite{yao2019pyhessian}, we feed images through ResNet architectures stripped of both batch normalization and residual connections, which we term \textit{stripped\_ResNets}. Interestingly, neither increasing the number of channels by two as in \citep{zagoruyko2016wide} (\textit{wide\_stripped\_ResNets}) nor increasing the filter sizes from $1\times1$ to $3\times3$ and from $3\times3$ to $5\times5$ (\textit{big\_stripped\_ResNets}) prevents gradient/curvature vanishing in deep \textit{stripped ResNets}.

\vspace{-2mm}
\paragraph{Architectural improvements}
As mentioned in the introduction, convolutional networks of depth 500 and more are not unheard-of. In fact, \cite{he2016identity} find that even going up to ResNet-1001 can improve test performance. Compared to the \textit{stripped\_ResNets} considered above, their architecture includes batch normalization and residual connections. Understanding the inner working of these components is an active area of research (e.g. \citep{hardt2016identity,bjorck2018understanding,kohler2019exponential,yao2019pyhessian,zhang2019convergence}). Most related to our cases, \cite{labatie2019characterizing} suggests that only the combination of the two is effective in stabilizing information flow in the large depth limit, which is line with our finding that ResNets suffer from \textit{exploding} gradients when taking either one of BN or residual connections out (Fig.~\ref{fig:cnn_vanishing_apx}).\footnote{Figure \ref{fig:mlp_vanishing_apx_BN} \& \ref{fig:cnn_vanishing_apx_BN} confirm this for MLPs and FCNs. See also \citep{yang2019mean} and \citep{zhang2019fixup}} %

\begin{figure}[ht]
\includegraphics[width=.33\textwidth]{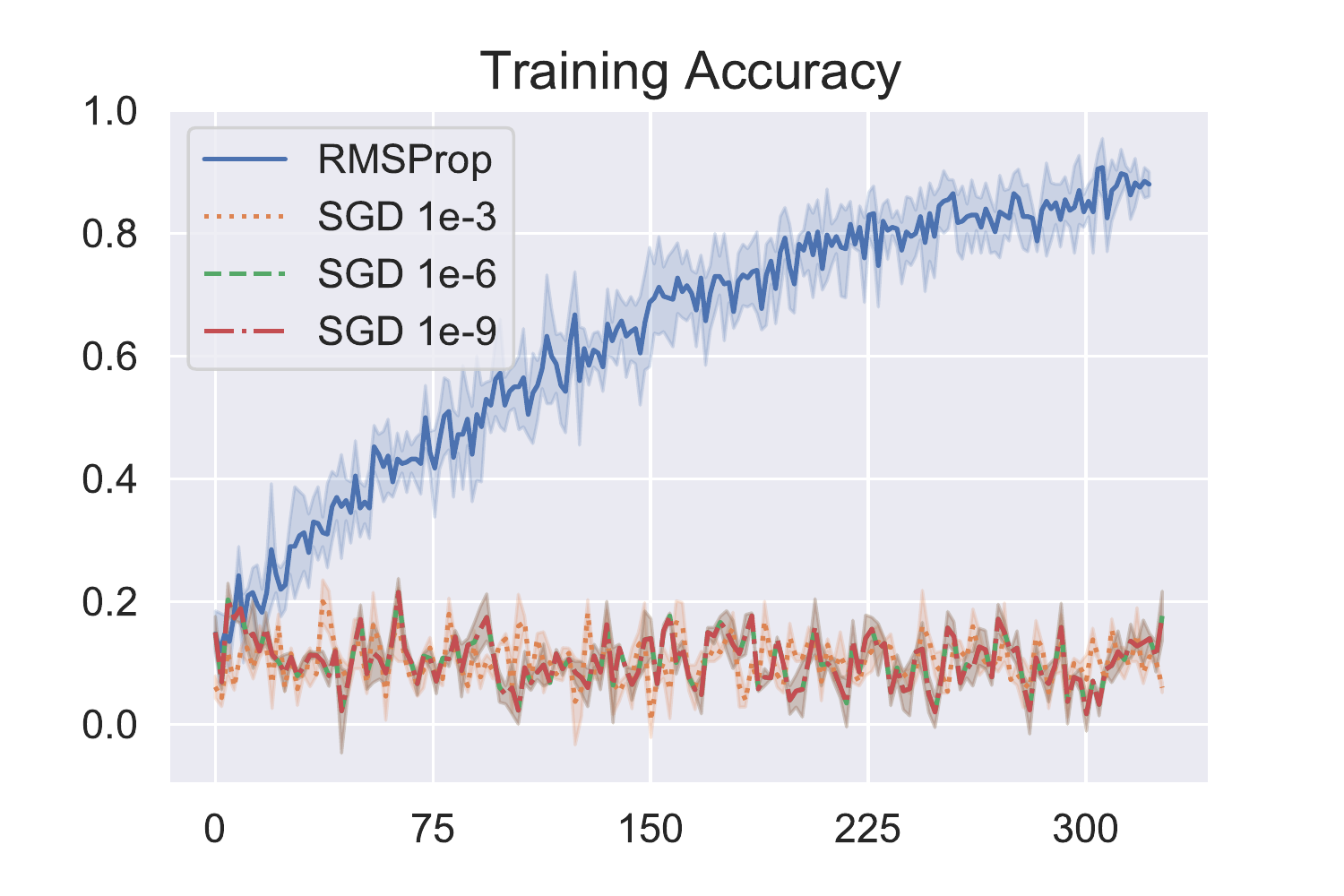}\hfill
\includegraphics[width=.33\textwidth]{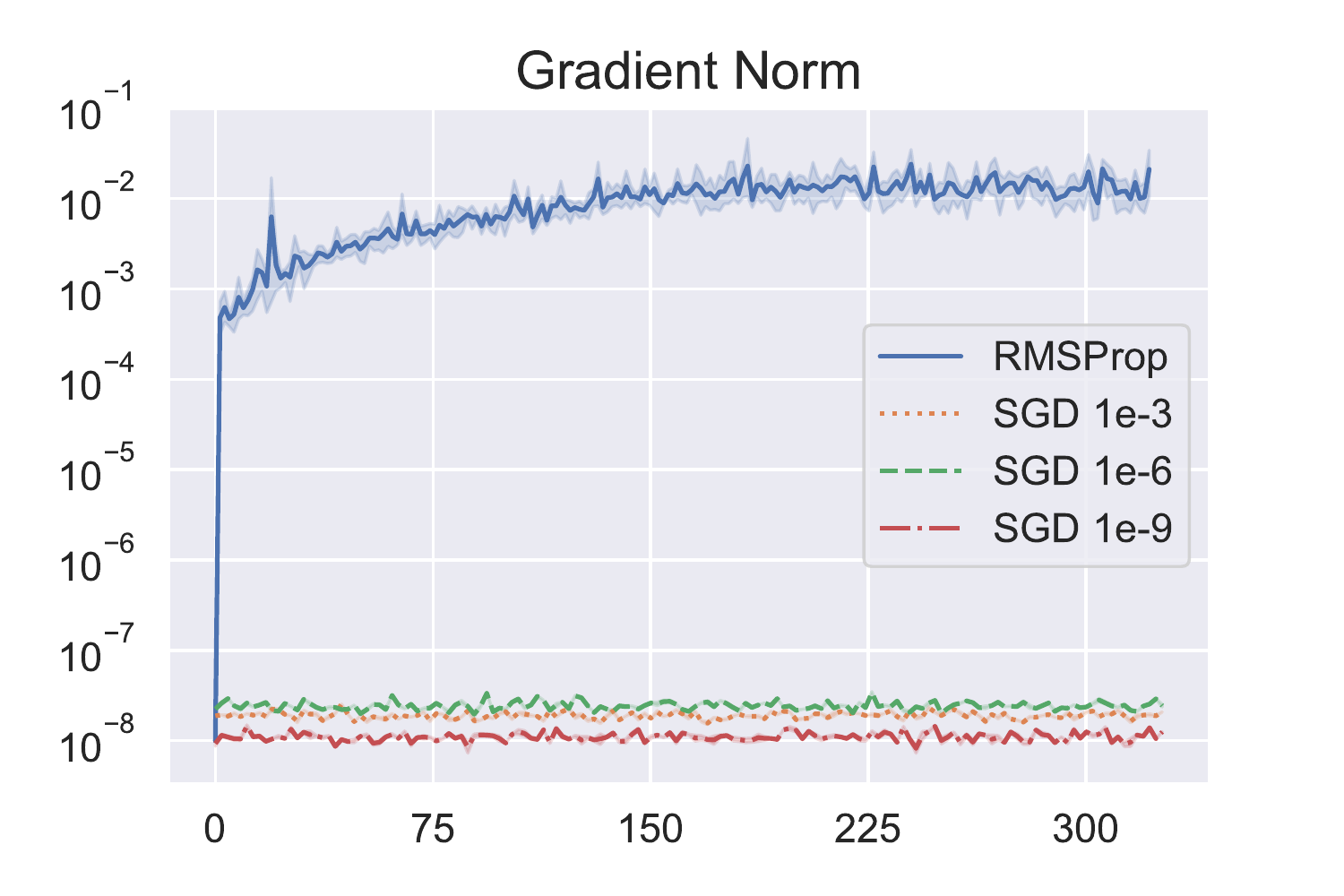}\hfill
\includegraphics[width=.33\textwidth]{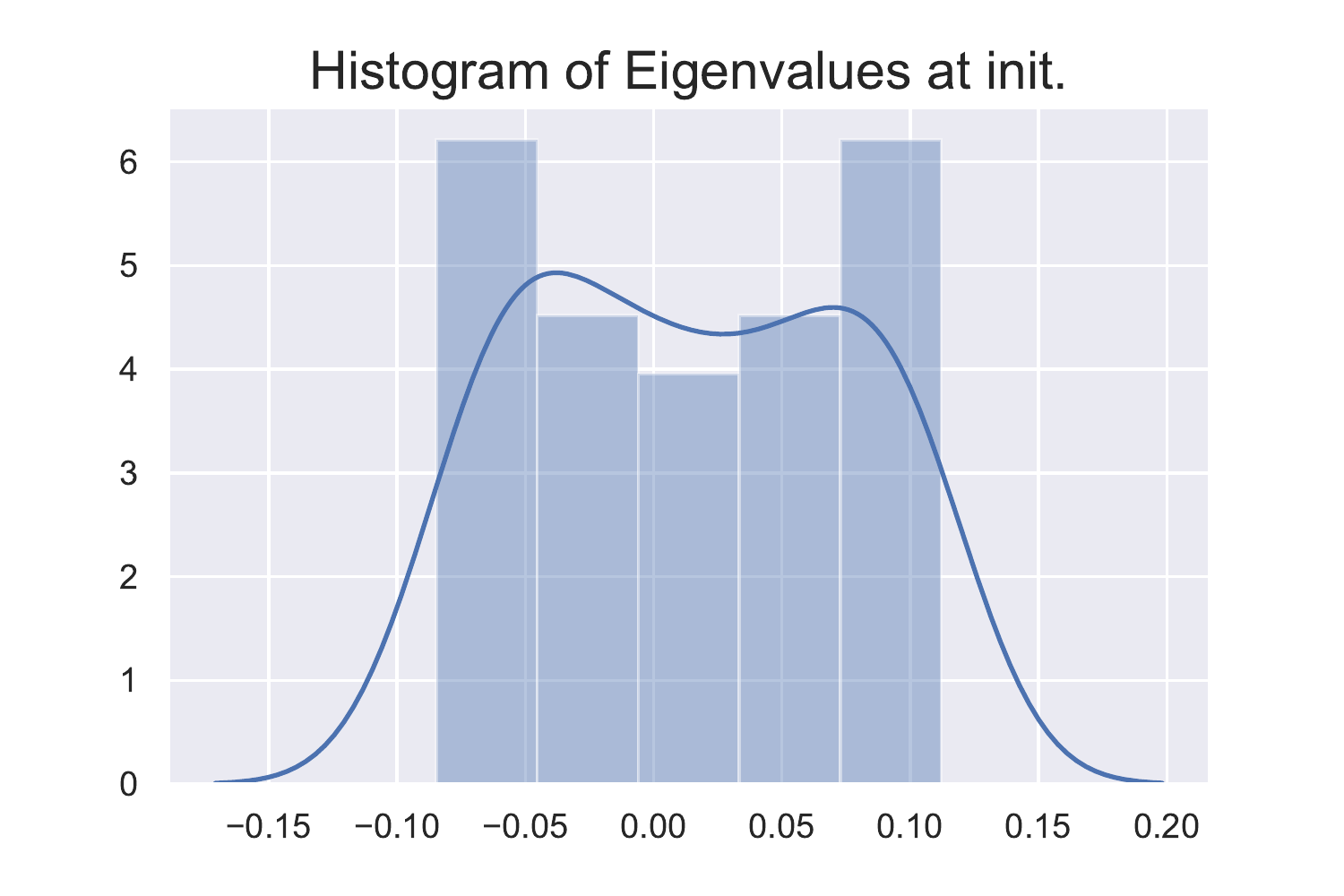}
\vspace{-2mm}
\caption{\small CIFAR-10 on a 500 layer \textit{stripped\_ResNet} with He init. Accuracy and gradient norm over epochs as well as eigenvalue histogram at initialization. Mean and 95$\%$ CI of 10 random seeds. See Appendix \ref{app:hyperparams} for hyperparameters and Figure \ref{fig:test_accs} for test accuracy}
\vspace{-3mm}
\label{fig:train_resnet}
\end{figure}

\vspace{-3mm}
\paragraph{Algorithmic improvements.} An obvious alternative is to generate robustness towards gradient/curvature vanishing directly in the training algorithm. As discussed in Section \ref{sec:width}, adaptive gradient methods (that were introduced about five years prior to batch normalization) are indeed able to escape flat initialization plateaus quickly. As a result, RMSprop can train a 500 layer \textit{stripped\_ResNet} without any normalization, skip-connections or learning-rate scheduling (Fig. \ref{fig:train_resnet}).

\section{Conclusion and outlook}
\vspace{-2mm}
Despite its long standing history, the phenomenon of vanishing and exploding gradients still lacks a comprehensive explanation. In this retrospective work we extended the current state of knowledge by: (i) showing that vanishing gradient co-occur with vanishing curvature, which clarifies the detrimental effect of depth on standard gradient descent at random initialization; (ii) examining when and why vanishing gradients occur in the first place. In this regard, we highlighted the role of effective width as well as the importance of combining Batch Normalization with residual connections; (iii) pointing to a remarkable curvature adaption property of adaptive gradients methods which allows them to train very deep networks despite the above mentioned issues. 

Looking ahead, we regard investigations into quantifying the optimal variance for narrow MLPs as well as for un-normalized convnets an interesting follow-up. Furhtermore, clarifying the interplay of initialization and activation functions (Fig.~\ref{fig:cnn_vanishing_apx_BN}) poses yet another interesting question. On a different route, we consider moving away from i.i.d. initialization and instead developing balanced initialization schemes that couple layers but preserve randomness a promising direction for designing simple deep neural network architectures that train and generalize well with gradient based methods.

\if@neuripsfinal
\paragraph{Broader impact statement.} We consider our work fundamental research on the principles of learning in high dimensional systems. Hence a broader impact discussion is not applicable.
\fi

\newpage
\bibliography{vanishing}
\bibliographystyle{neurips_2021}

\appendix
\onecolumn
\section{Additional experimental details and results}

\subsection{Experimental setup}\label{app:hyperparams}
All experiments are conducted in PyTorch 1.7 \citep{paszke2019pytorch}. We run experiments on up to 8 Tesla V100 GPUs with 32 GB memory.

\paragraph{Figure \ref{fig:figure1_lin_1} \& \ref{fig:figure1_relu}.} We draw samples $\Xm\in\mathbb{R}^{n\times d}$ from a multivariate Gaussian distribution $\mathcal{N}(0,\Im_d)$, choosing $n=100$ and $d$ equal to the network width ($x$-axis). Targets $\Ym\in\mathbb{R}^{n\times d}$ are generated form a 1-hidden-layer network of width as on the $x$-axis. We use as mean-squared-error loss. The networks are simple MLPs with weight matrices in $d\times d$. As initialization we use LeCun uniform, i.e. $W^\ell_{i,j}\sim\mathcal{U}\left[-\tfrac{1}{\sqrt{d}},\tfrac{1}{\sqrt{d}}\right]$. As long as memory is sufficient we compute the full hessian, above we randomly sub-sample hessian blocks (layers).

\paragraph{Figure \ref{fig:width_effects_mlp} \& \ref{fig:mlp_vanishing_apx_BN}.}
As in Figure \ref{fig:figure1_lin_1} \& \ref{fig:figure1_relu} but using Xavier uniform initialization $W^\ell_{i,j}\sim\mathcal{U}\left[-\sqrt{\tfrac{3}{d}},\sqrt{\tfrac{3}{d}}\right]$ \citep{glorot2010understanding} for linear- and He uniform initialization $W^\ell_{i,j}\sim\mathcal{U}\left[-\sqrt{\tfrac{6}{d}},\sqrt{\tfrac{6}{d}}\right]$ \citep{he2015delving} for ReLU networks. The networks are simple MLPs with weight matrices in $\left\lceil\sqrt{d}\right\rceil\times d$ in the top and $d\times d$ in the bottom row. 

In Figure \ref{fig:mlp_vanishing_apx_BN} we add residual connections and batch normalization. Notably, the residual connections skip a set of \textit{three} layers.

\paragraph{Figure \ref{fig:train_mlp}.}
We train Fashion-MNIST \citep{xiao2017fashion} with the given train-test split, on a 32 hidden unit, 128 hidden layer MLP with ReLU activations. All optimizers are depicted with (individually) grid-searched learning rate (in terms of training accuracy) in the set ${1e-3, 5e-4, 1e-4, 5e-5, 1e-5, 5e-6, 1e-6, 5e-7, 1e-7, 5e-8, 1e-8 }$. Depicted are SGD and Momentum with learning rate $1e-4$ and RMSprop as well as Adam with learning rate $1e-5$. Batch size is 128 for all optimizers. The momentum factor for SGD was set to $0.9$.

\begin{figure}[ht]
\centering
\includegraphics[width=.33\textwidth]{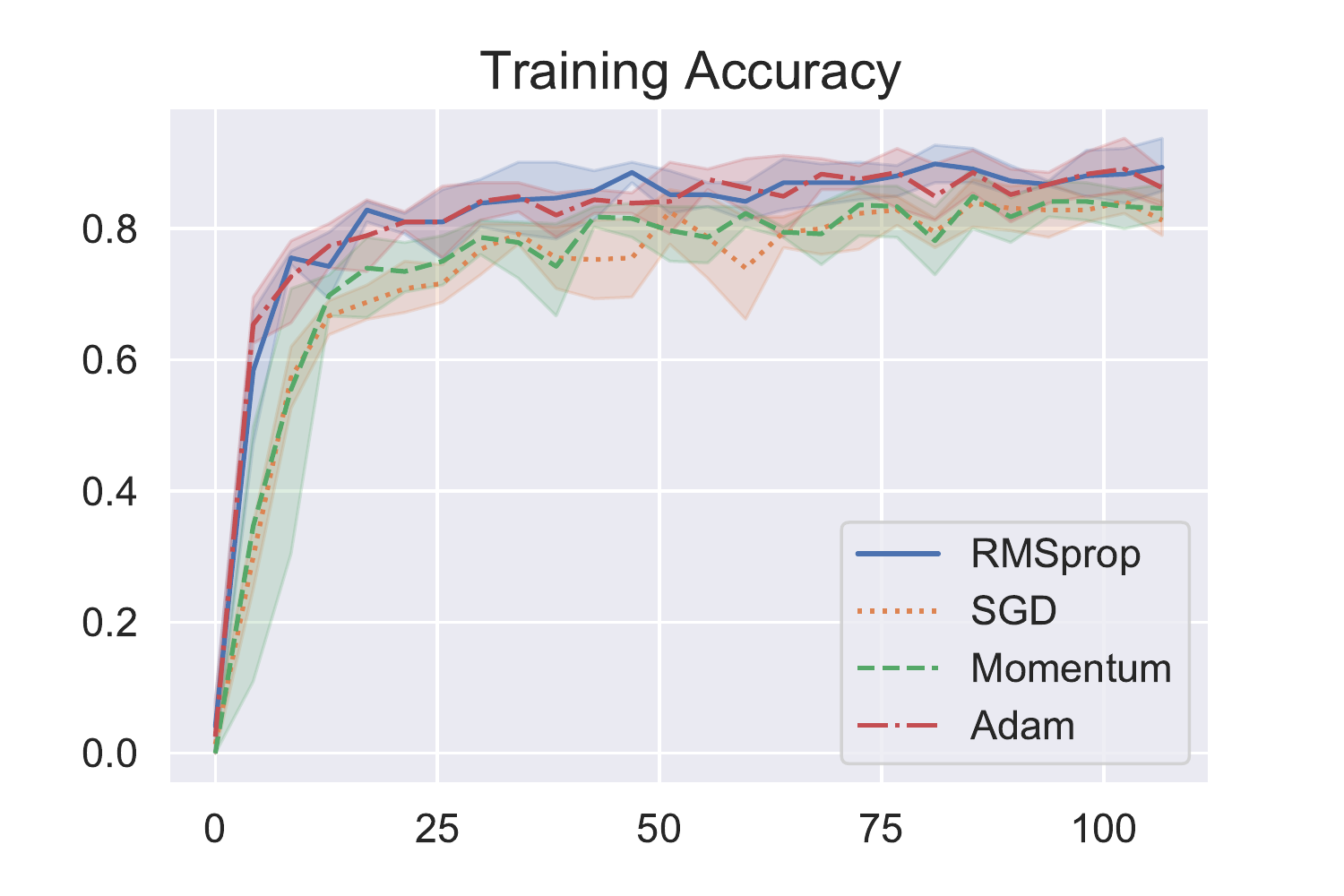}\hfill
\includegraphics[width=.33\textwidth]{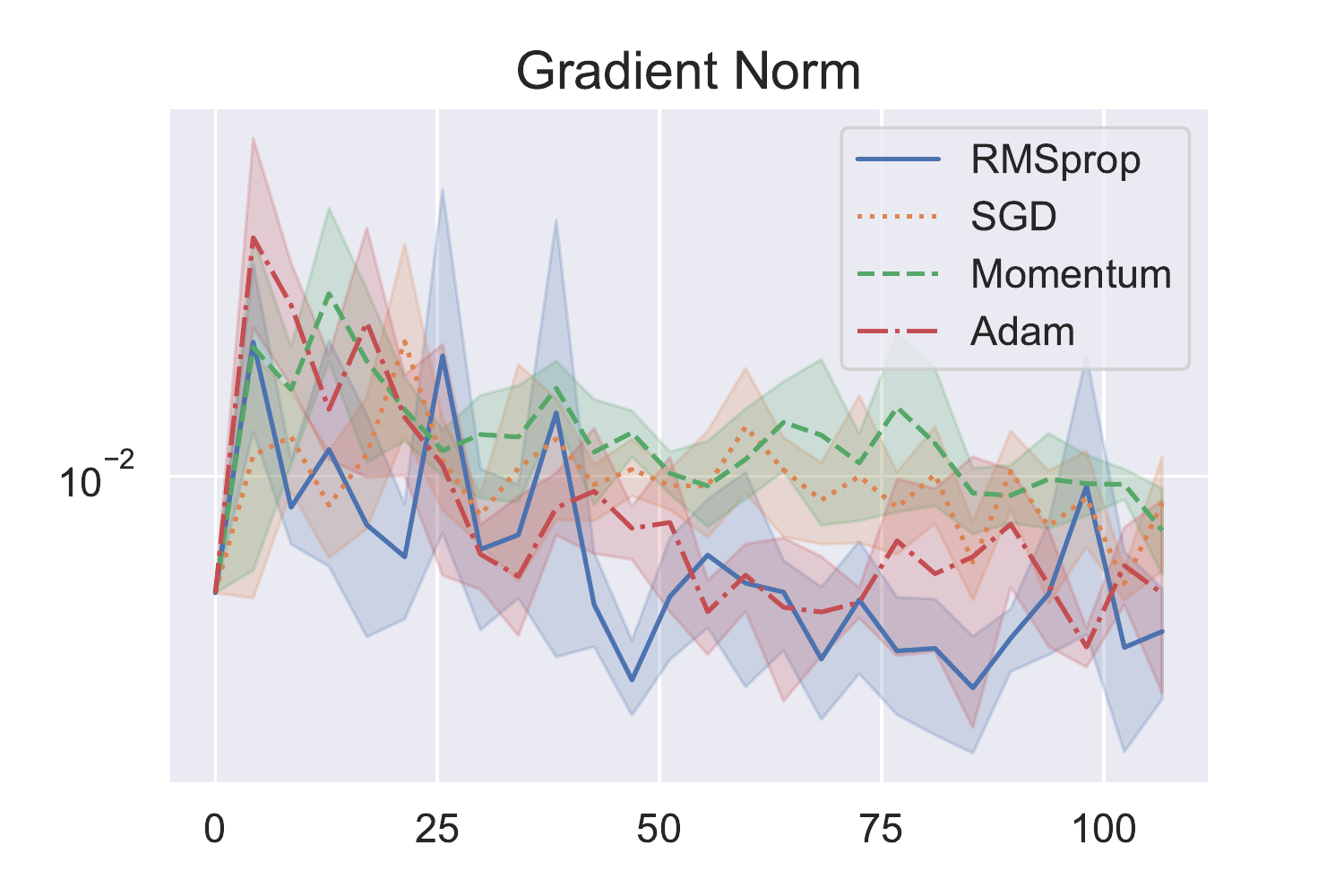}\hfill
\includegraphics[width=.33\textwidth]{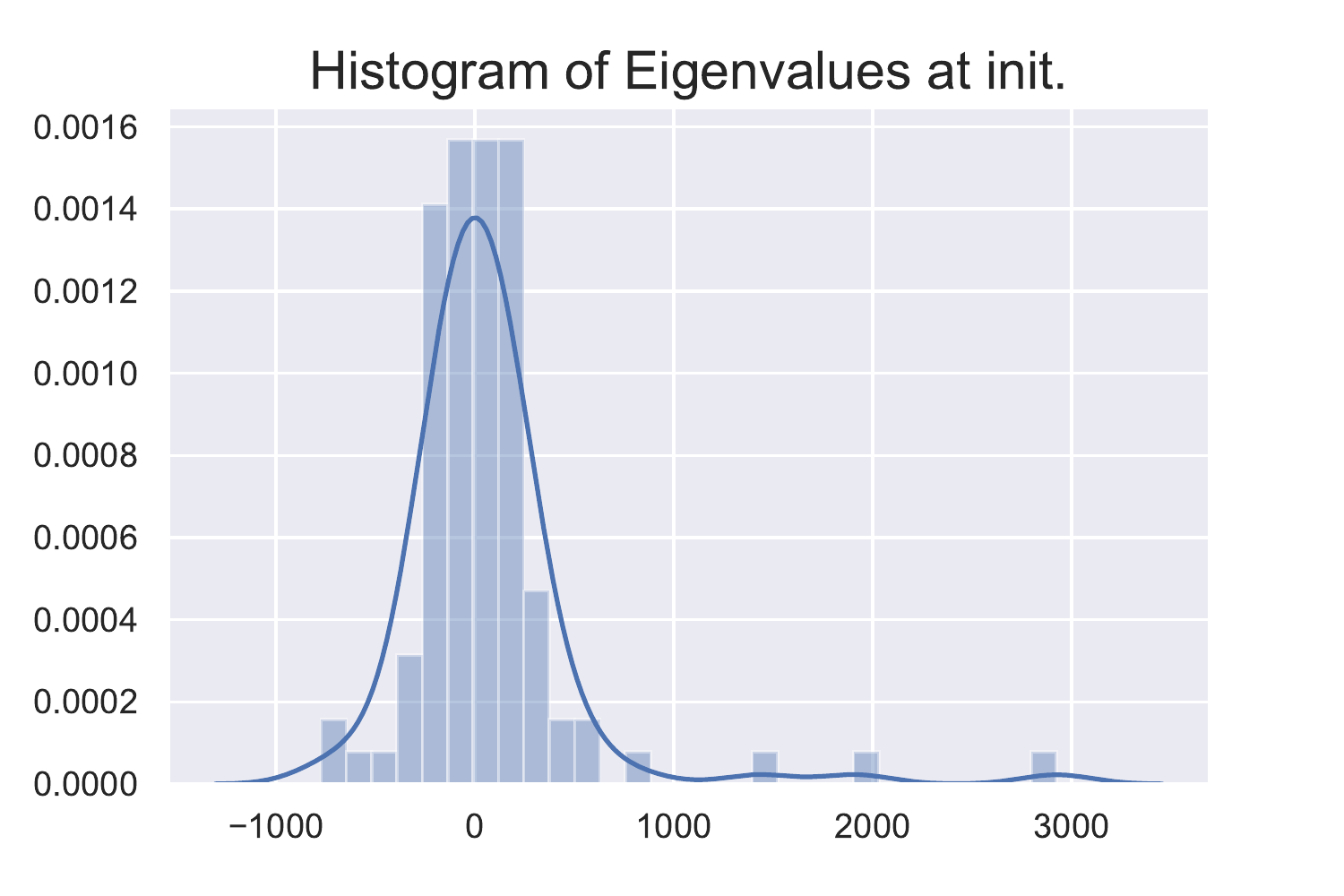}

\caption{Addendum to Fig. \ref{fig:train_mlp}: Training \textbf{Fashion-MNIST} on a \textbf{\textit{wide}} (128 units per layer) \textbf{128 layer ReLU MLP} (He init.) with batch size 64 and grid-searched learning rates. Training accuracy and gradient magnitude over epochs as well as eigenvalues at initialization. Mean and 95$\%$ CI of 10 random seeds. Clearly, increased width prevents vanishing and thus allows SGD to train.\looseness=-1}
\vspace{-3mm}
\label{fig:fm_apx_wide}
\end{figure}

\paragraph{Figure \ref{fig:fm_apx_wide}.}
Same setting as in Fig. \ref{fig:train_mlp} but with network width equal to network depth (128).

\paragraph{Figure \ref{fig:width_effects_cnn} \& \ref{fig:cnn_vanishing_apx_BN}.}
We here consider a fully convolutional image to image learning setting where each layer has $c$ kernels of size $3\times 3$ that operate with a padding of $1$. As a result, the image resolution does not change over depth. As inputs, we use a batch of 32 CIfAR-10 images \cite{krizhevsky2009learning}. We feed them trough the networks once at the original $32\times 32$ resolution and once down-sampled to $7\times 7$ images and compute a mean-squared-error loss at the end using the input image as target. We show plots for ReLU nets but note that the general picture is the same for linear networks (vanishing happens just a bit slower, compare MLPs).

In Figure \ref{fig:cnn_vanishing_apx_BN} we add residual connections and batch normalization. Notably, the residual connections skip a set of \textit{three} layers (similarly to the ResNet). In fact, we found exploding gradients/curvature when residual connections only skip one layer.

\paragraph{Figure \ref{fig:cnn_vanishing} \& \ref{fig:cnn_vanishing_apx}.}
In these figures we feed CIFAR-10 images \citep{krizhevsky2009learning} at the original $32\times 32$ resolution through convolutional networks that resemble the ResNet architecture but omit both Batch Normalization and residual connections. These networks have 4 main blocks of layers which operate at image resolutions: $56\times 56$, $28\times 28$, $14\times 14$ and $7\times 7$ and with $64$, $128$, $256$ and $512$ channels. Each of these blocks consists of 3 convolutional layers. We depict result on networks of depth $18,34,50,101,152,200,270,336,500$ which have the following block configurations
$$[2,2,2,2], [2,4,4,2], [3,4,6,3], [3,4,23,3], [3,8,36,3]$$ $$[3,24,36,3], [3,36,48,3], [3,44,62,3], [3,70,90,3].$$ 
Unless stated differently, these networks use ReLU activations. 

When neither Batch Norm nor residual connections are present we called the network \textit{stripped\_resnet}. The \textit{wide\_stripped\_resnet} has twice the number of channels in each block compared to the ones stated in the previous paragraph. The \textit{big\_stripped\_resnet} has the original number of channels but all $3\times 3$ filters are replaced by $5\times 5$ - and all $1\times 1$ filters are replaced by $3\times 3$ filters.

\paragraph{Figure \ref{fig:train_resnet}. }
We train CIFAR-10 on a \textit{stripped\_resnet} with 500 layers. All optimizers are depicted with (individually) grid-searched learning rate (in terms of training accuracy) in the set ${1e-3, 5e-4, 1e-4, 5e-5, 1e-5, 5e-6, 1e-6, 5e-7, 1e-7, 5e-8, 1e-8 }$. Batch size is 64 for all optimizers. For RMSprop, the best found learning rate was $1e-6$.

\subsection{Additional results}
\label{app:additional_results}

\begin{figure}[ht]
\includegraphics[width=.33\textwidth]{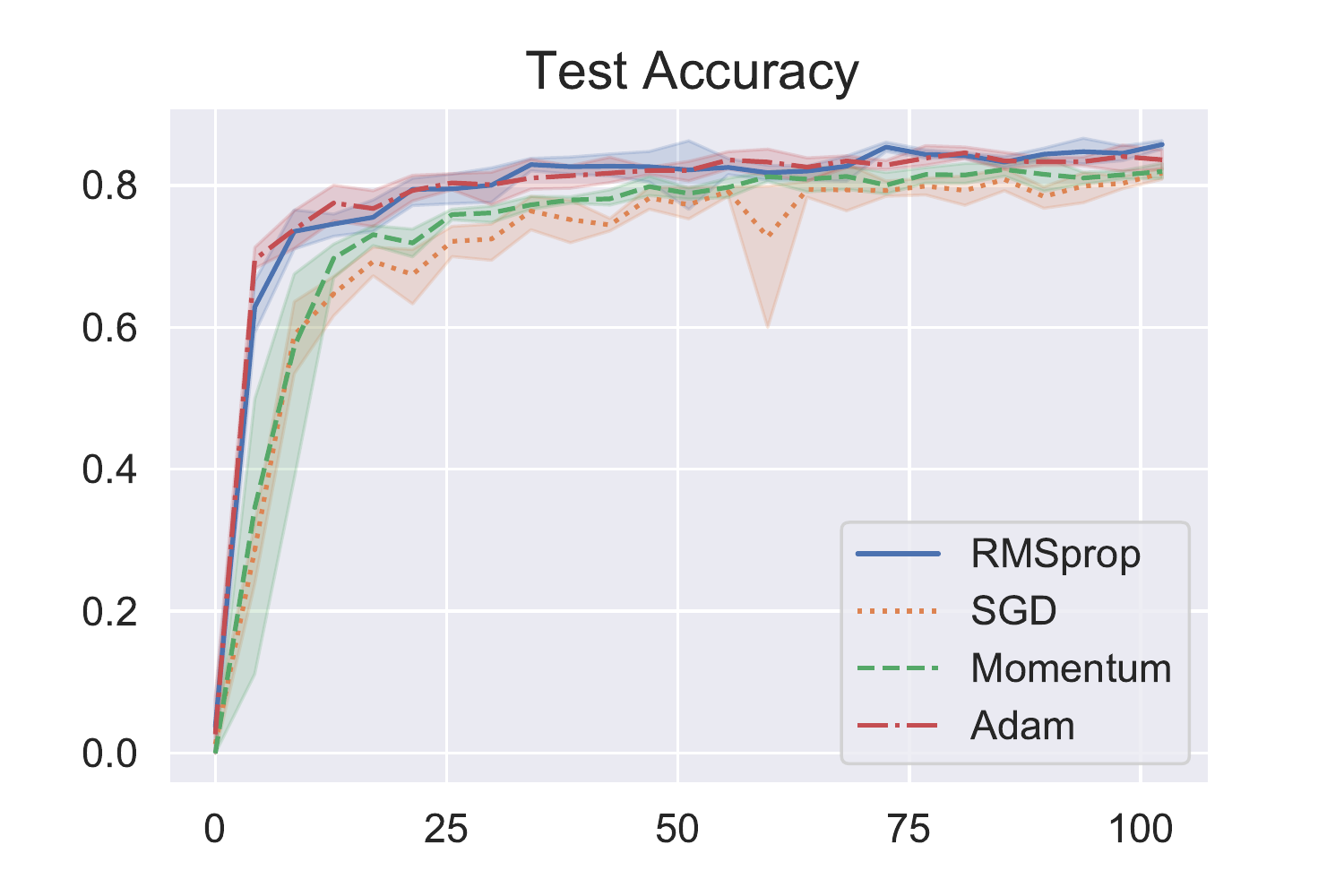}\hfill
\includegraphics[width=.33\textwidth]{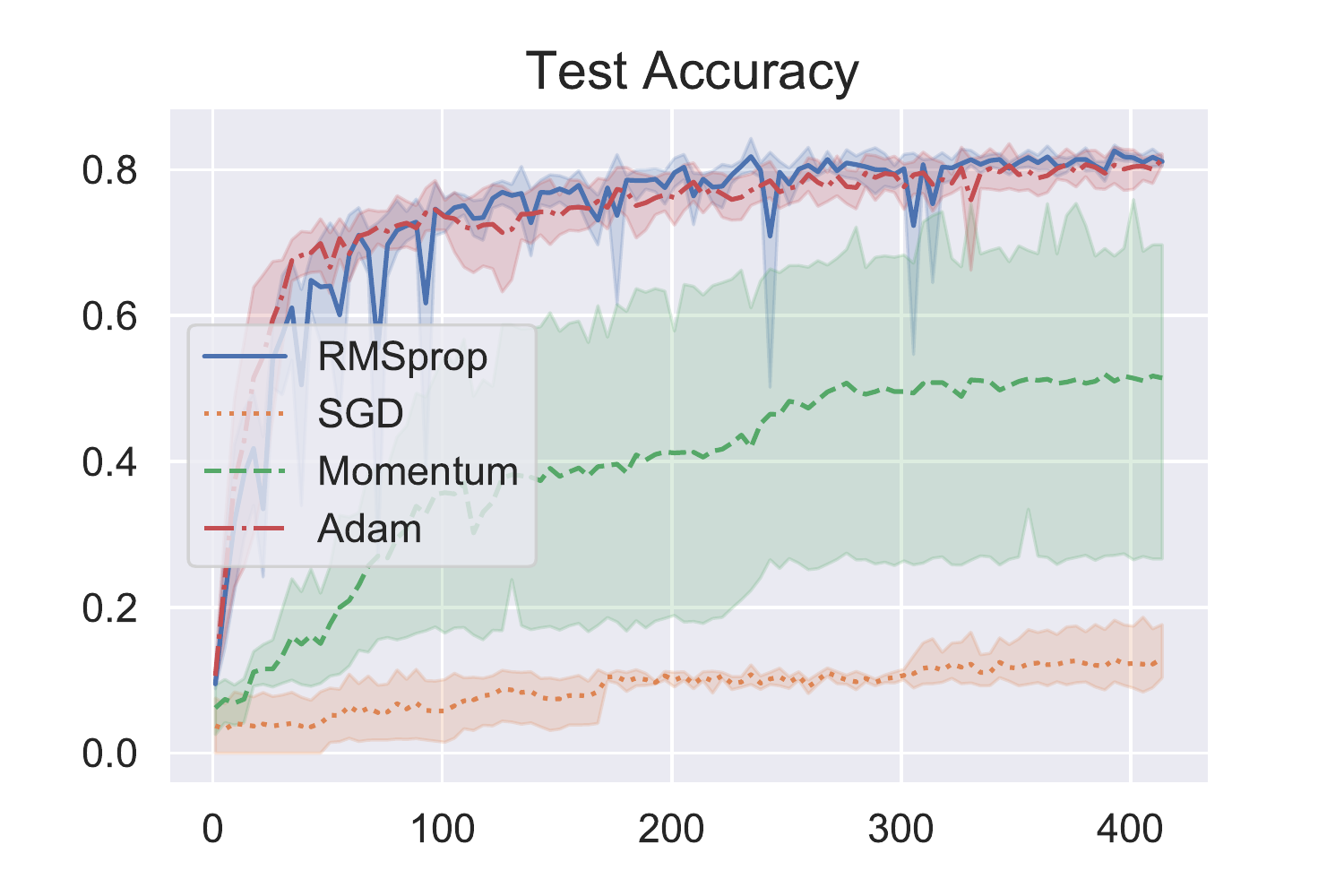}\hfill
\includegraphics[width=.33\textwidth]{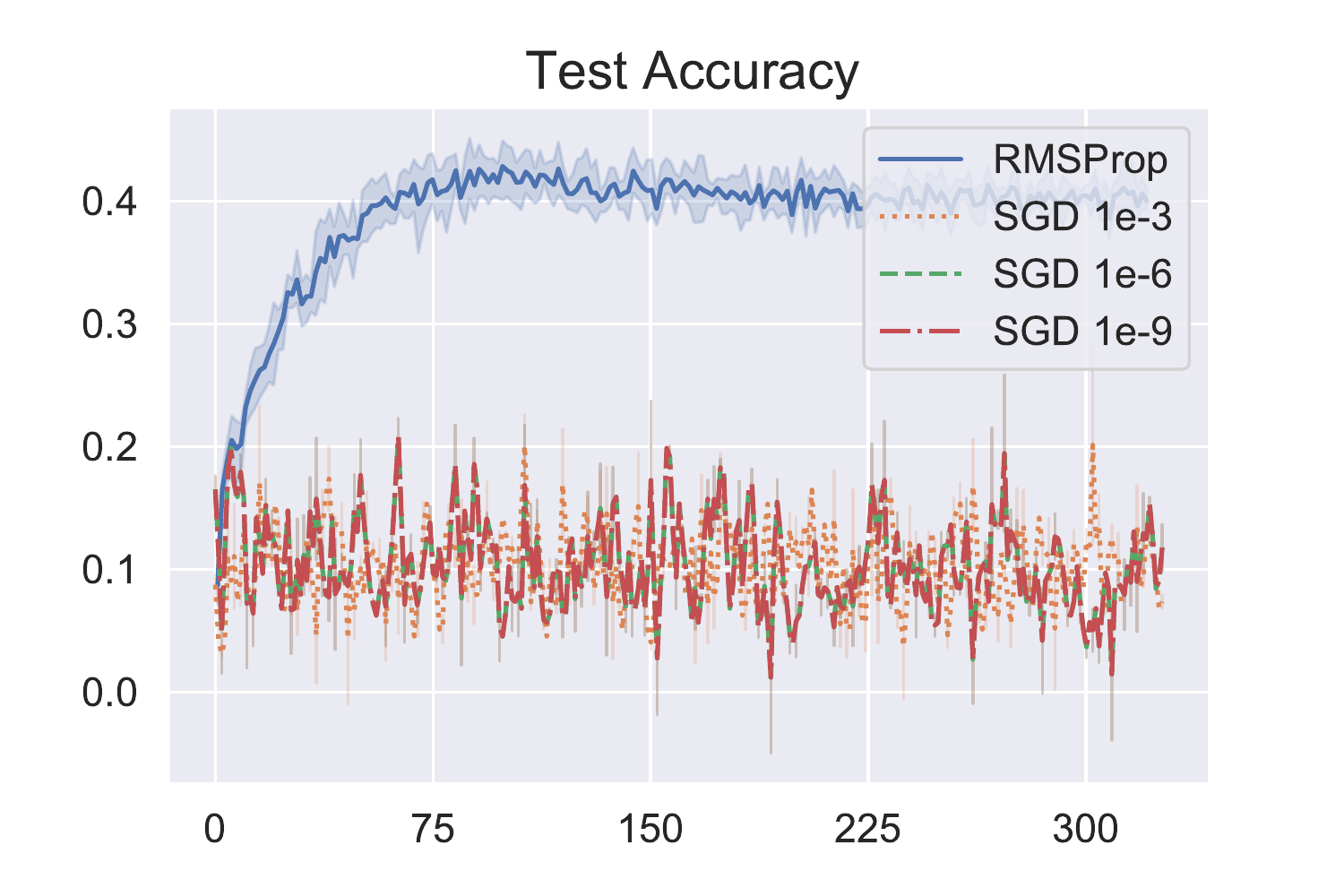}
\caption{\textbf{Test set performance} of the wide- and narrow Fashion-MNIST ReLU MLP (left and middle) as well as on the CIFAR-10 stripped ResNet500 (right). It becomes evident that RMSProp heavily overfits in the stripped ResNet, which is not surprising given that the network has no regularization what so ever and memorizing CIFAR-10 is comparatively easy for such large convnets.}
\vspace{-3mm}
\label{fig:test_accs}
\end{figure}

\begin{figure}[ht]
\centering
\begin{minipage}[b]{0.33\linewidth}
\includegraphics[width=1.05\textwidth]{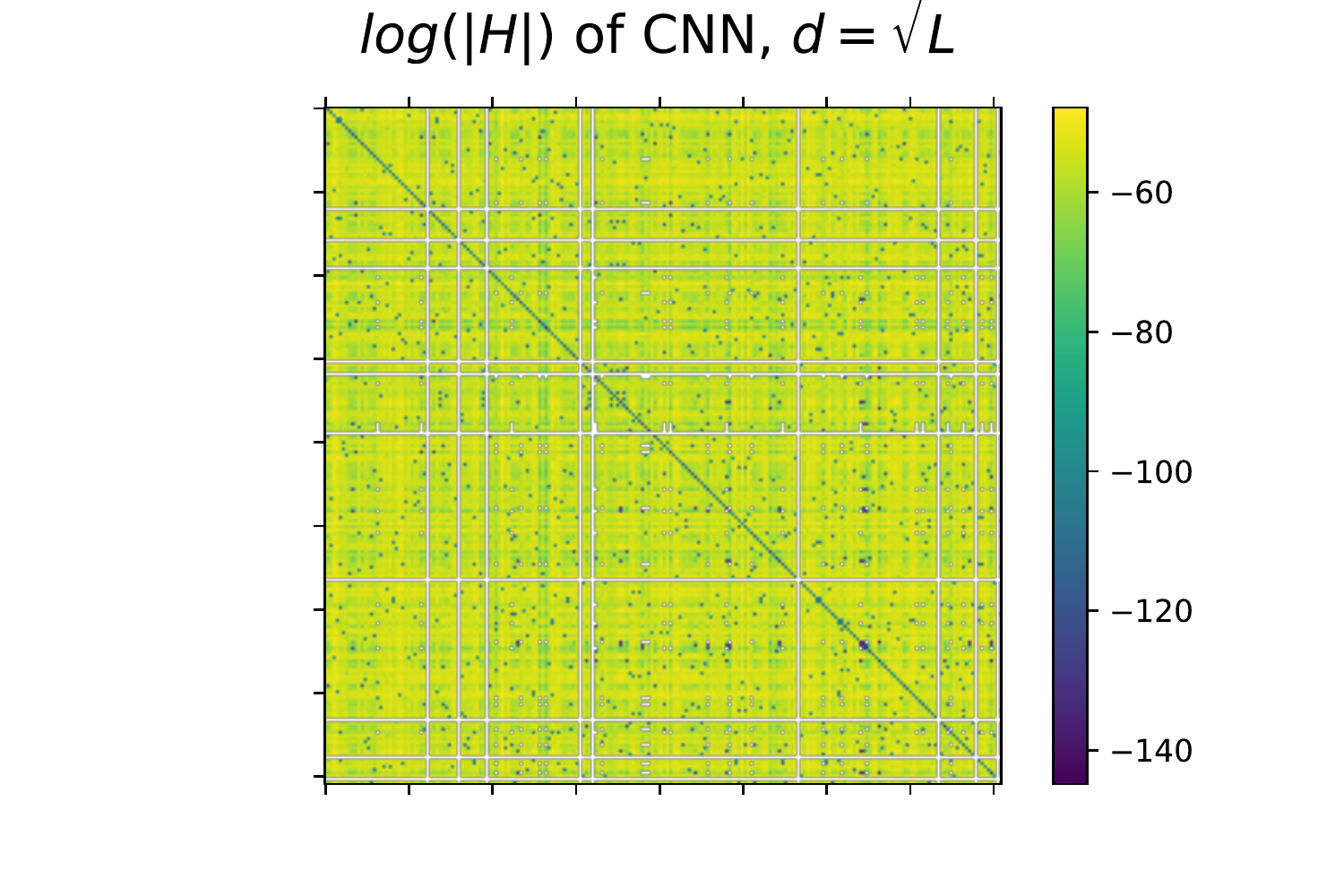}
\end{minipage}
\hspace{0.005cm}
\begin{minipage}[b]{0.33\linewidth}
\centering
\includegraphics[width=0.94\textwidth]{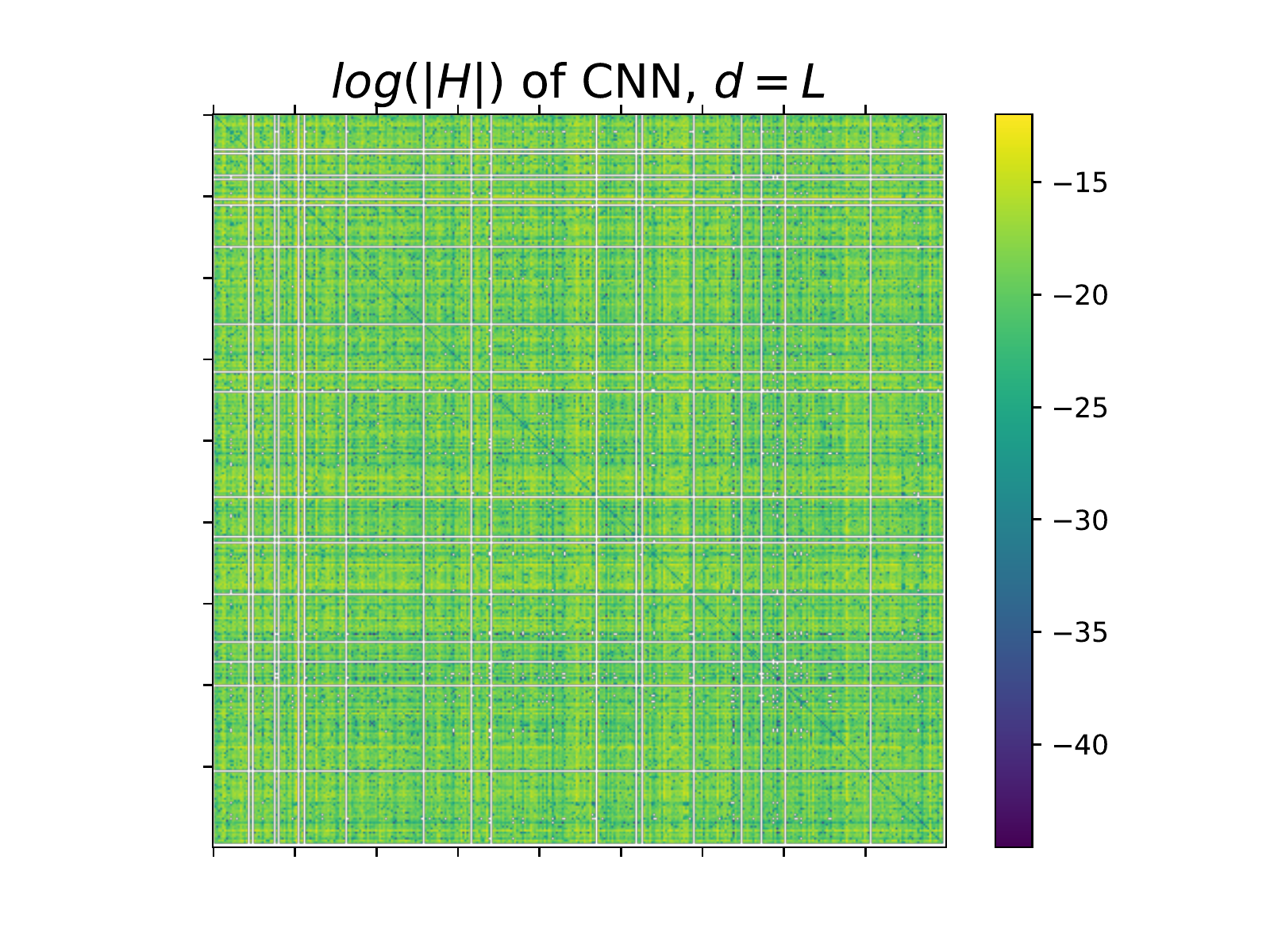}
\end{minipage}
\caption{Addendum to Fig. \ref{fig:width_effects_cnn}. \textbf{Hessians of fully convolutional ReLU networks} of depth $L=128$ on downscaled CIFAR-10 samples at random initializatio (He init.). Contrary to the MLP case in Fig. \ref{fig:figure1_lin_1}, the Hessians are no longer approximately block-hollow but approximately hollow. Furthermore, as expected, the hollowness decreases in network width (left to right). }
\label{fig:FCN_Hessian}
\end{figure}

\begin{figure}[ht]

\centering

\begin{minipage}[b]{0.33\linewidth}
\includegraphics[width=1.05\textwidth]{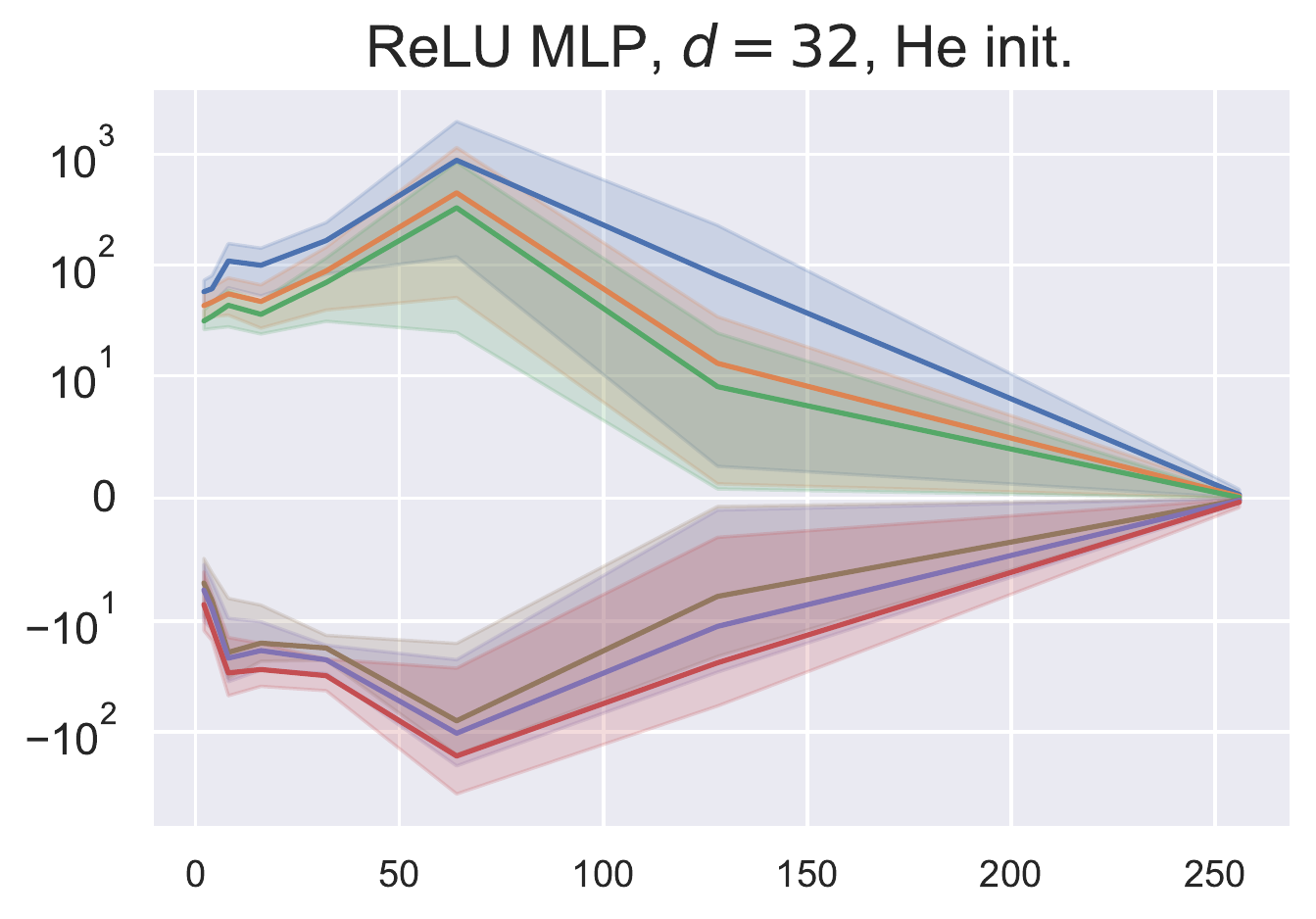}
\end{minipage}
\hspace{0.05cm}
\begin{minipage}[b]{0.33\linewidth}
\centering
\includegraphics[width=1.35\textwidth]{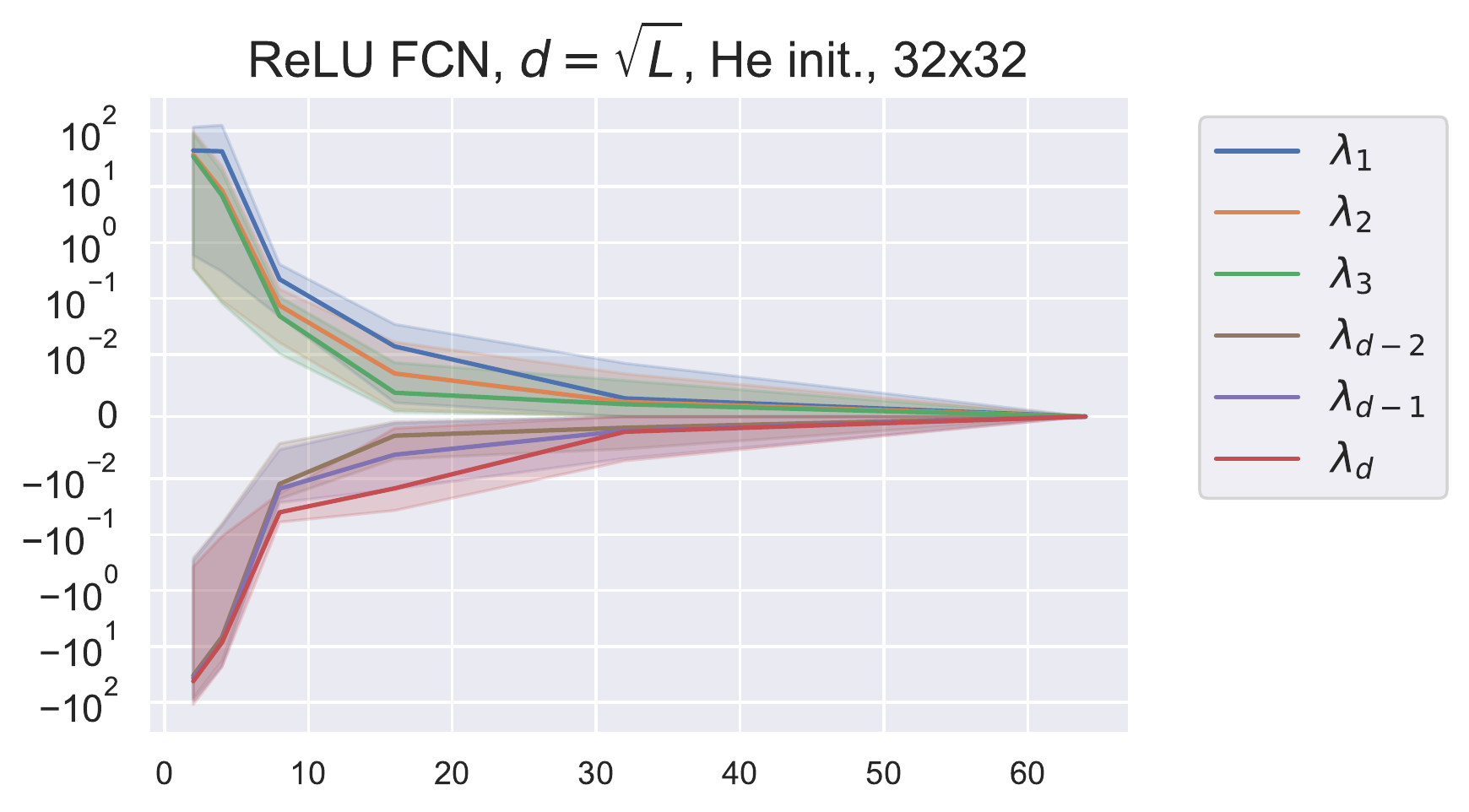}
\end{minipage}

\hspace{0.005cm}

\caption{\textbf{Eigenvalues over depth:} Largest ($\lambda_1,\lambda_2,\lambda_3$) and smallest ($\lambda_{d-2},\lambda_{d-1},\lambda_{d}$) three eigenvalues on a symlog scale over depth. Left: ReLU MLPs of width 32. Right: ReLU FCNs of $d=\sqrt{L}$. Both on Fashion-MNIST. Mean and 95$\%$ CI of 5 random seeds. }
\label{fig:eigenvalues_over_depth}
\end{figure}

\begin{figure}[ht]
\centering
\includegraphics[width=0.45\textwidth]{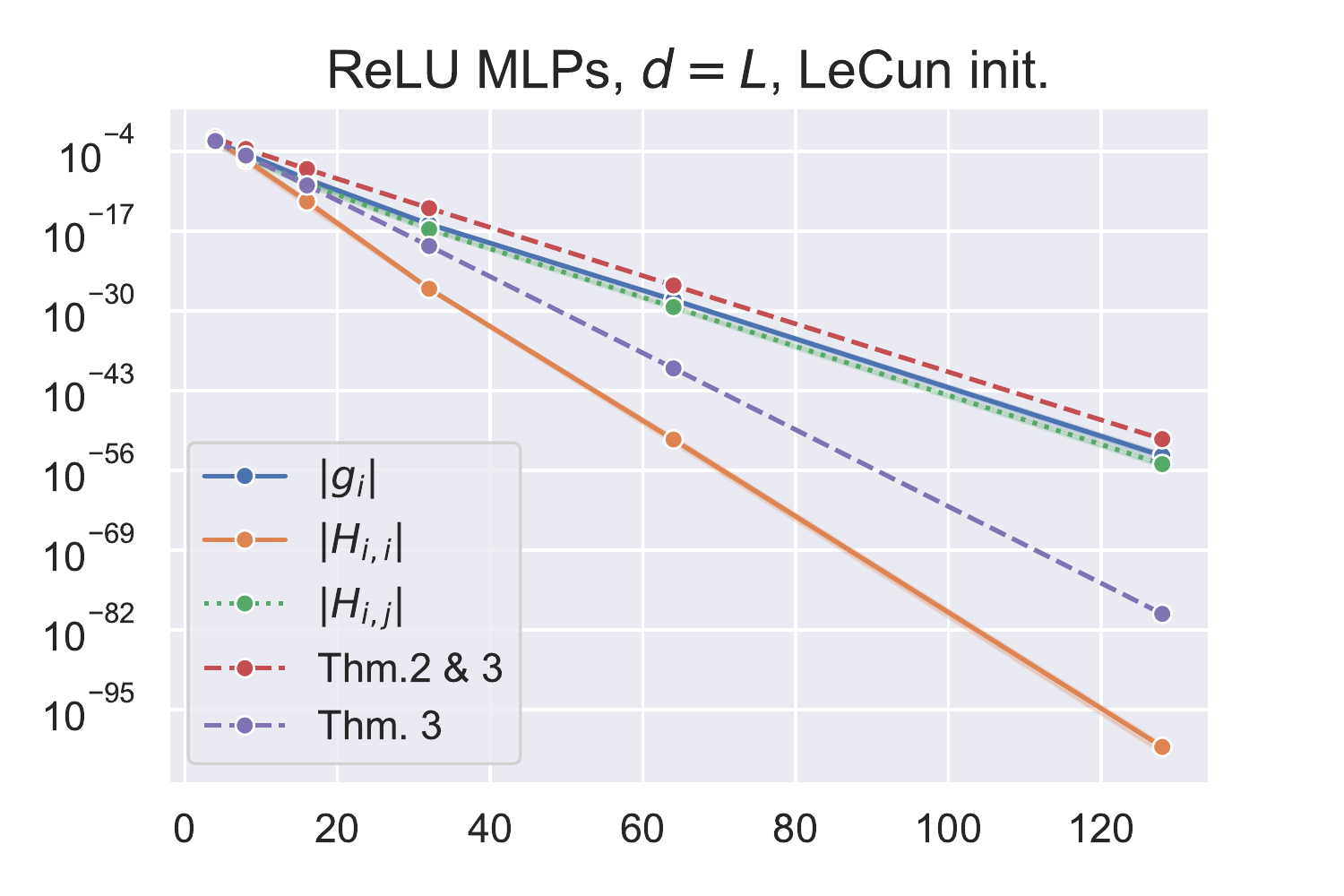}

\caption{Addendum to Fig. \ref{fig:figure1_lin_1}. Vanishing gradient and Hessian for deep \textbf{ReLU MLPs with LeCun init.}. x-axis depicts depth \textit{and} width of the networks. Mean and 95$\%$ CI of 10 random seeds. }
\label{fig:figure1_relu}
\end{figure}

\begin{figure}[ht]
 \centering

\begin{minipage}[b]{0.33\linewidth}
\includegraphics[width=1.05\textwidth]{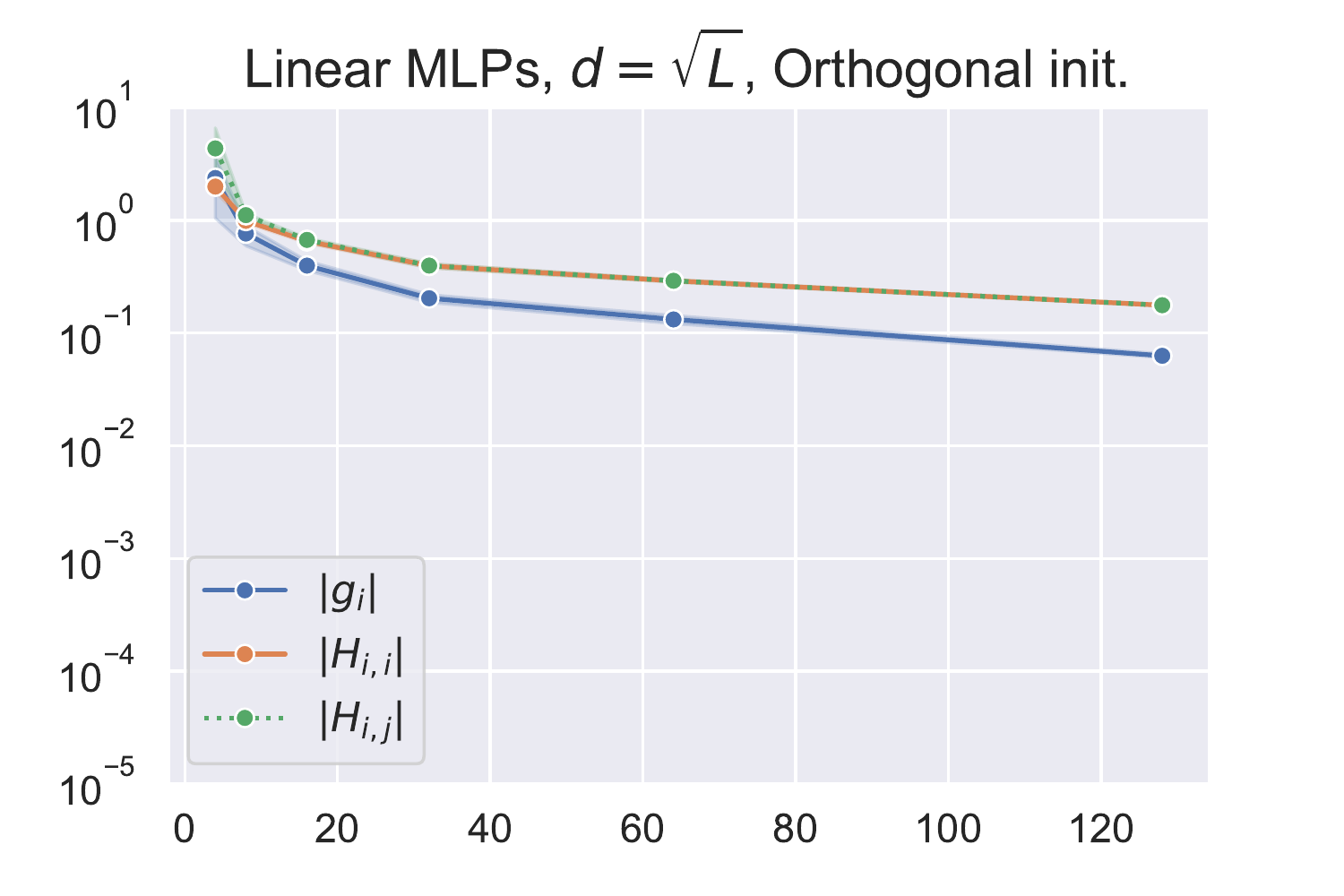}
\end{minipage}
\hspace{0.005cm}
\begin{minipage}[b]{0.33\linewidth}
\centering
\includegraphics[width=1.05\textwidth]{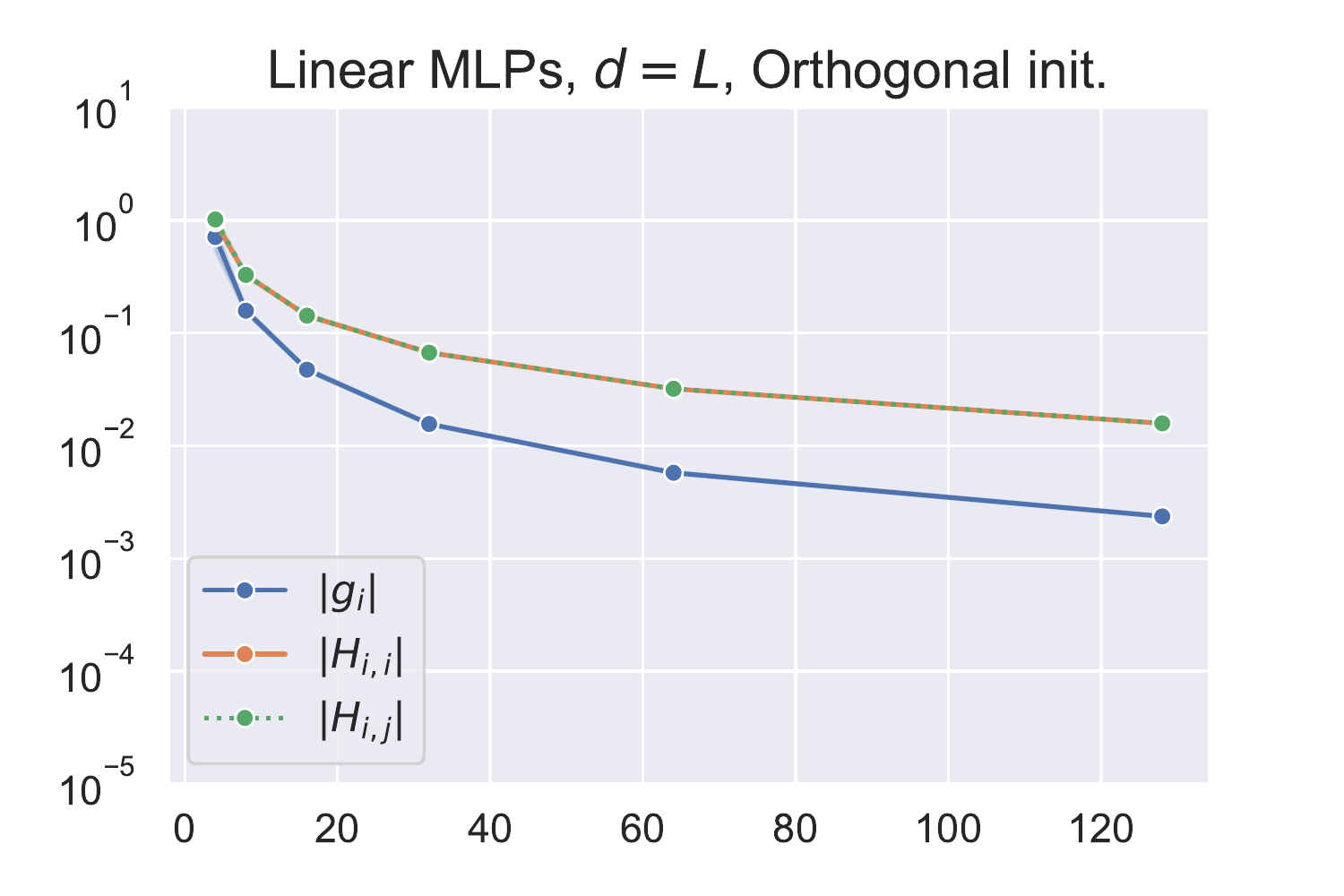}
\end{minipage}
\caption{Addendum to Fig. \ref{fig:width_effects_mlp}: Gradient/curvature on \textbf{linear MLPs} over depth initialized with \textbf{orthogonal initialization} \citep{saxe2013exact}. This strategy is very robust towards gradient/curvature vanishing on linear MLPs but quickly yields absolute zeros on ReLU MLPs (PyTorch implementation \citep{paszke2019pytorch}, not shown). Mean and 95\% CI of 15 runs.}
\label{fig:orthogonal_init}
\end{figure}

 \begin{figure}
 \begin{center}

\begin{minipage}[b]{\figsize\linewidth}
\centering
\includegraphics[width=1.05\textwidth]{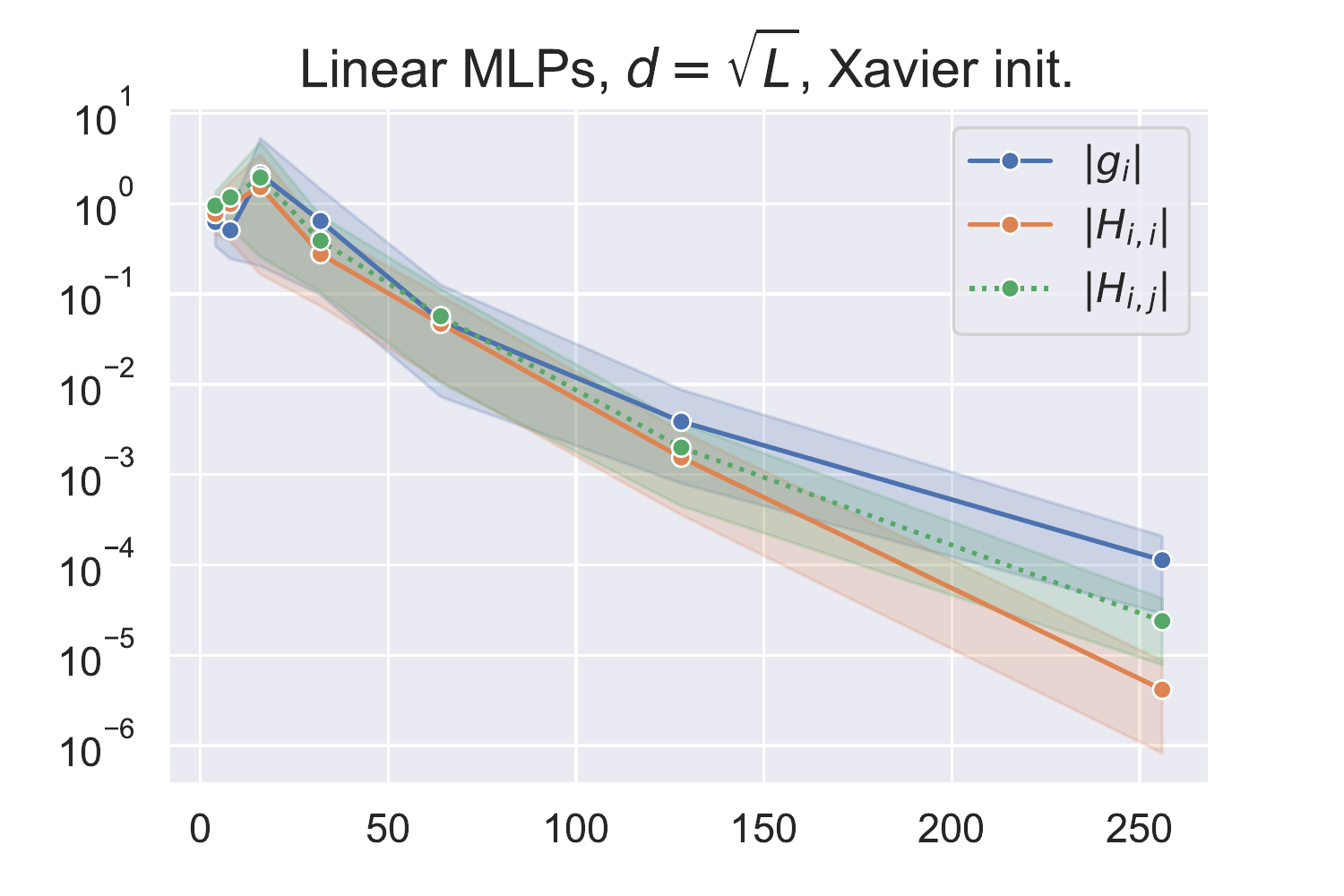}
\hspace*{0.75cm}\tiny{depth}
\end{minipage}
\hspace{0.005cm}
\begin{minipage}[b]{\figsize\linewidth}
\centering
\includegraphics[width=1.05\textwidth]{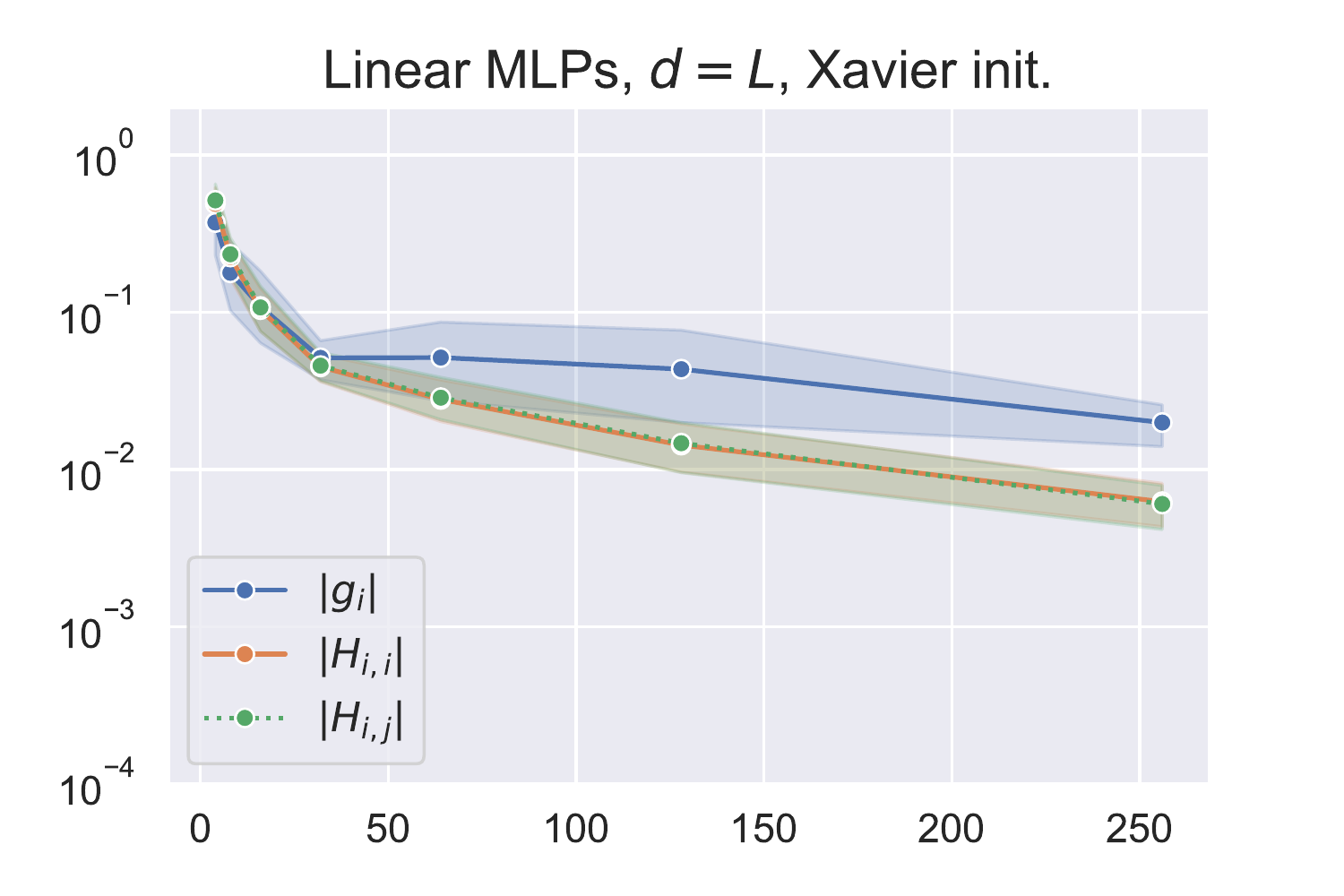}
\hspace*{0.75cm}\tiny{depth}
\end{minipage}
\end{center}

\caption{Addendum to Fig. \ref{fig:width_effects_mlp}: Effect of width in \textbf{linear} MLPs: Gradient and curvature scaling on Fashion-MNIST over depth. While quantities vanish on the left, where the width scales only as square-root of depth, the right shows stable behaviour. Mean and 95\% CI of 15 runs.}
\label{fig:width_effects_mlp_lin}
\end{figure}

\begin{figure}[t]
 \begin{center}

  \begin{minipage}[b]{\figsize\linewidth}
\centering
\includegraphics[width=1.05\textwidth]{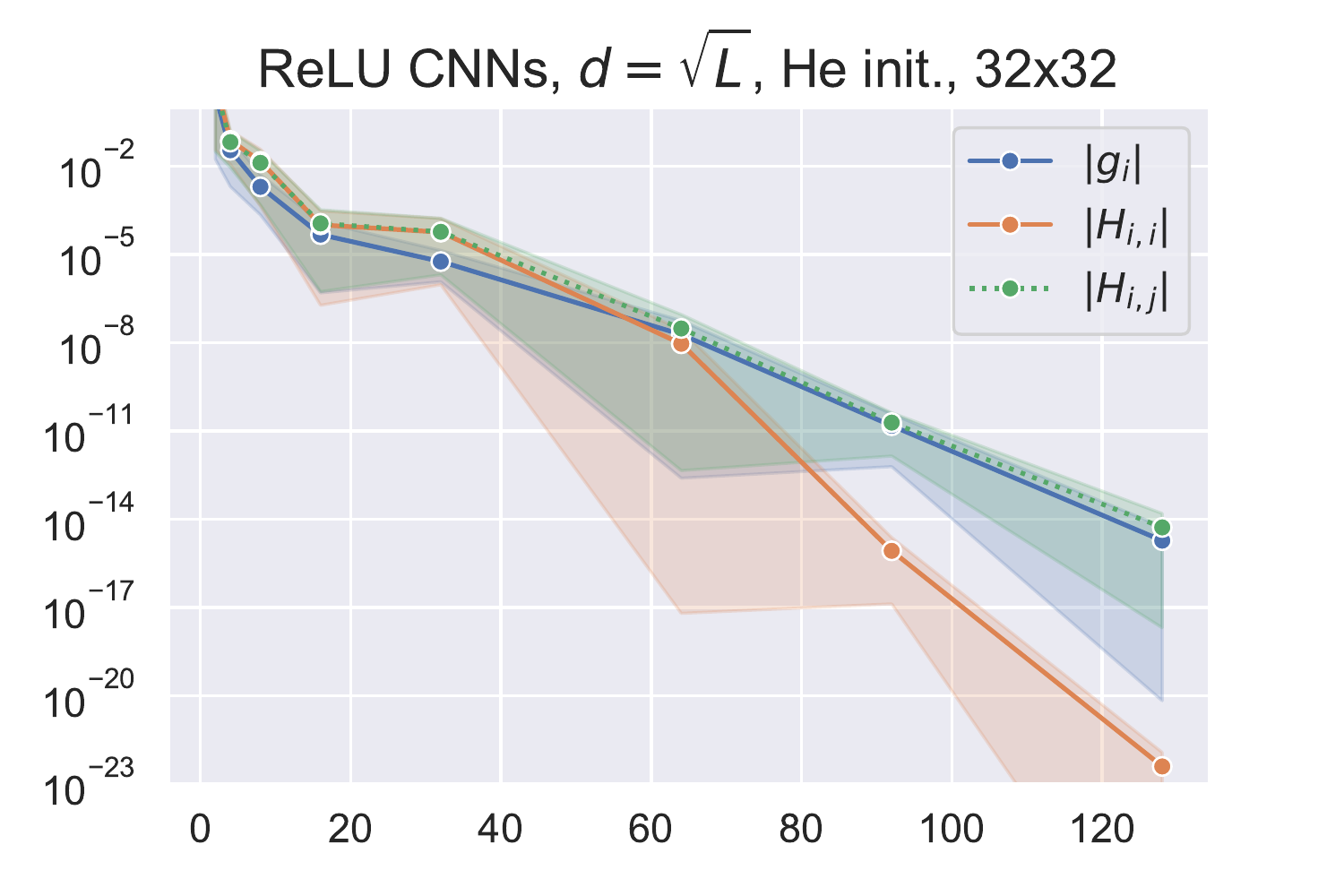}
\end{minipage}
\hspace{0.005cm}
\begin{minipage}[b]{\figsize\linewidth}
\centering
\includegraphics[width=1.05\textwidth]{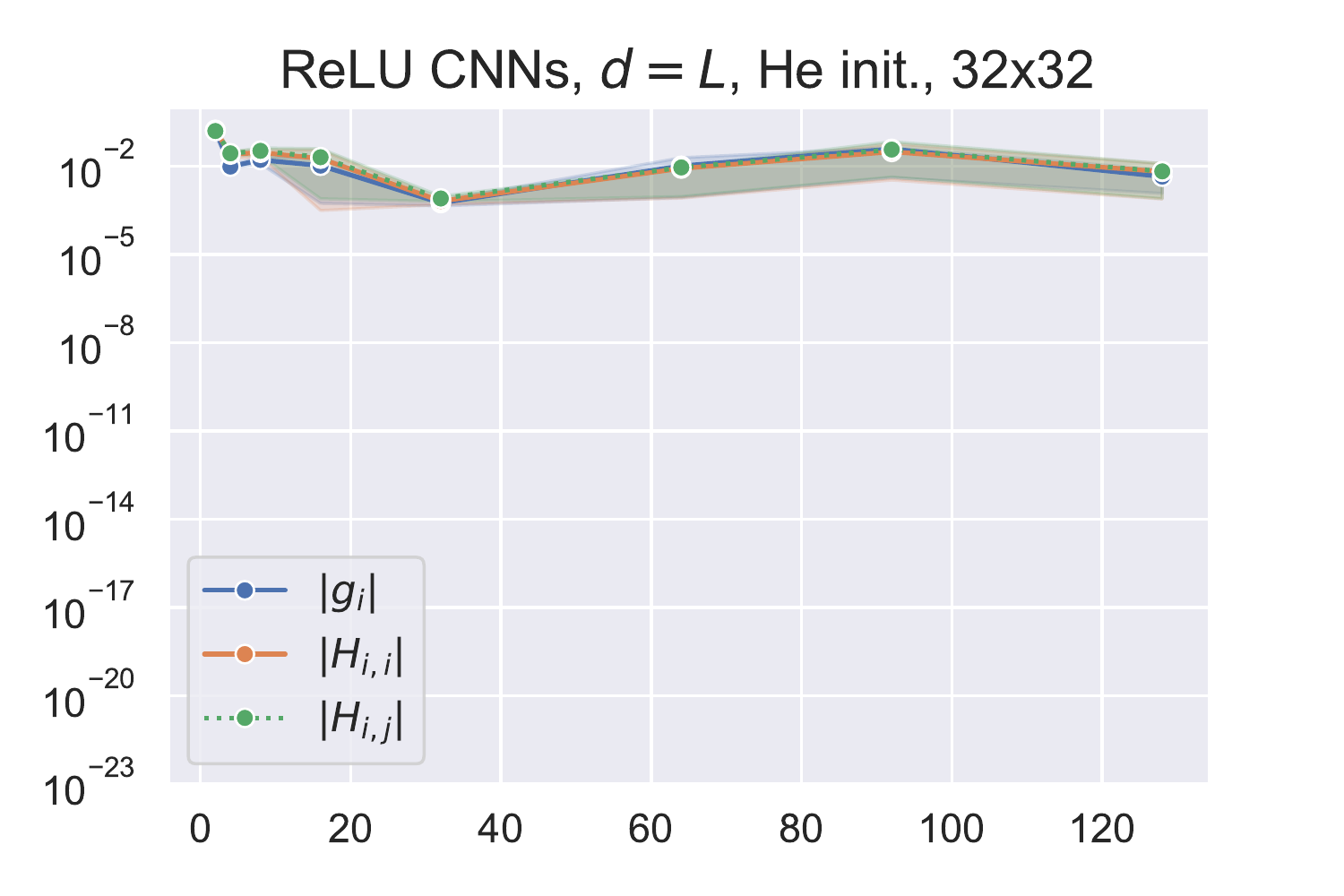}

\end{minipage}

\end{center}

\caption{Addendum to Figure \ref{fig:width_effects_cnn}: Gradient/curvature on \textbf{FCNs} over depth at large resolution (32x32). Mean and 95\% CI of 15 runs.}
\label{fig:width_effects_cnn_apx}
\vspace{-2mm}
\end{figure}

\begin{figure}[ht]
\centering
\includegraphics[width=0.32\textwidth]{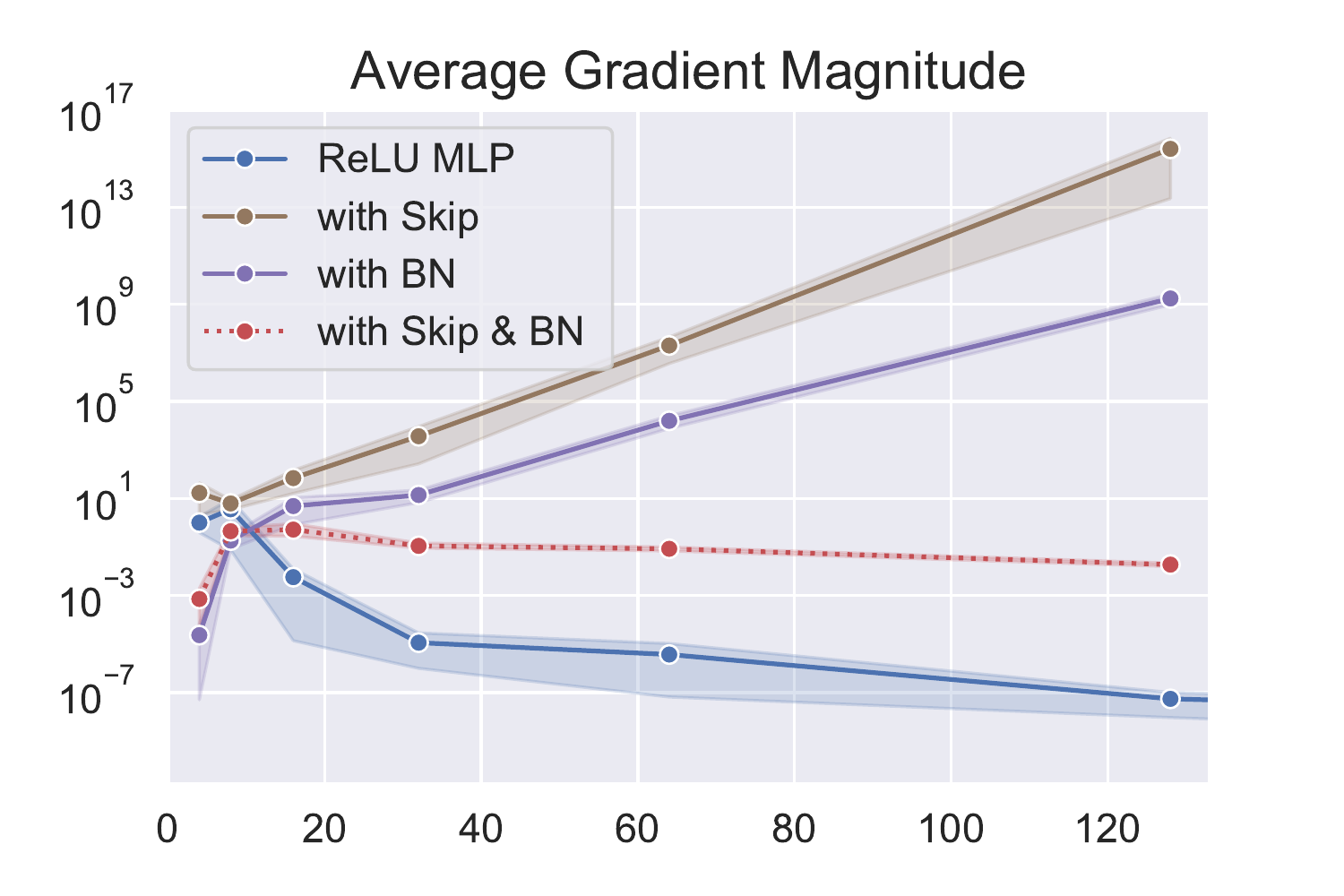}
\includegraphics[width=0.32\textwidth]{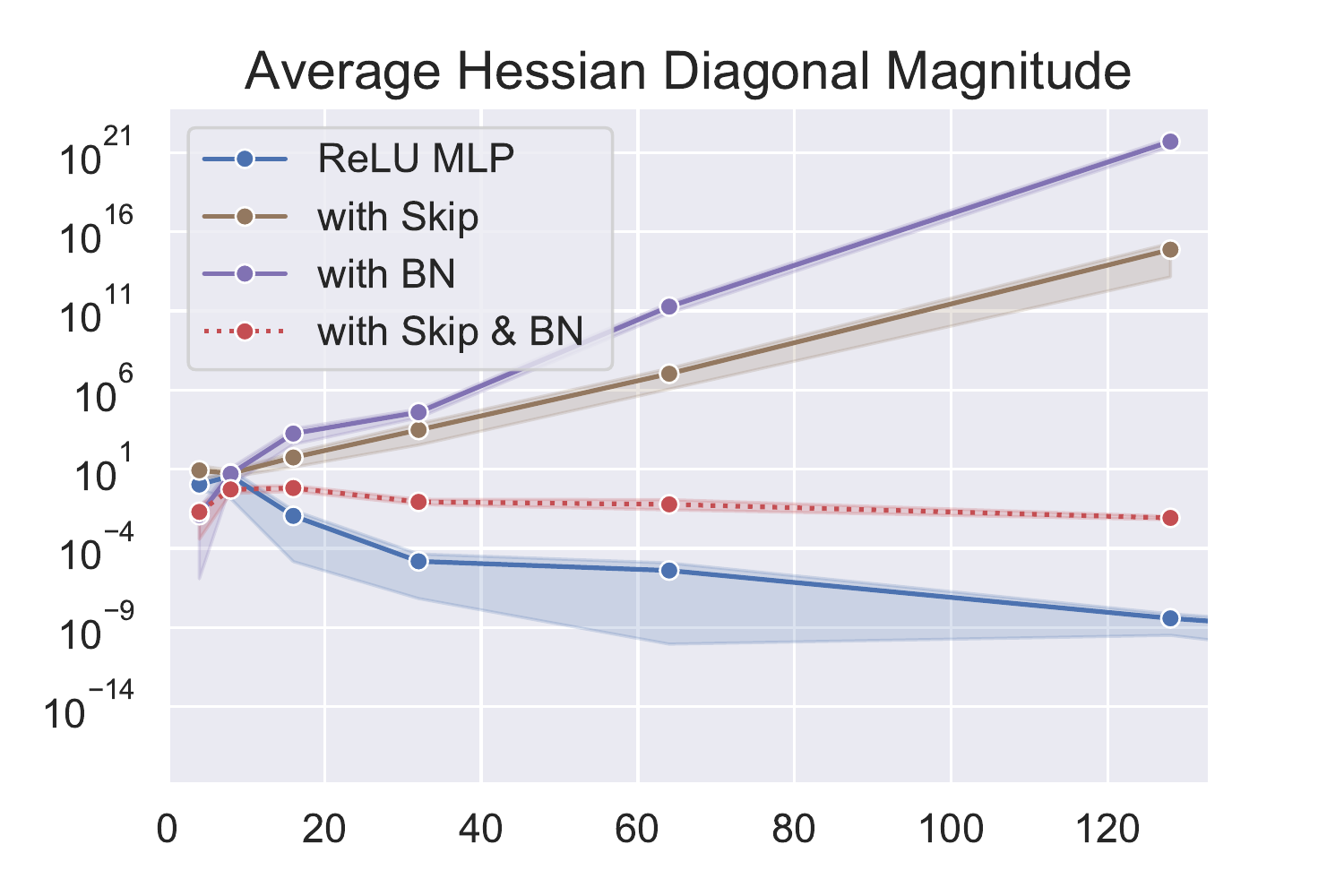}
\includegraphics[width=0.32\textwidth]{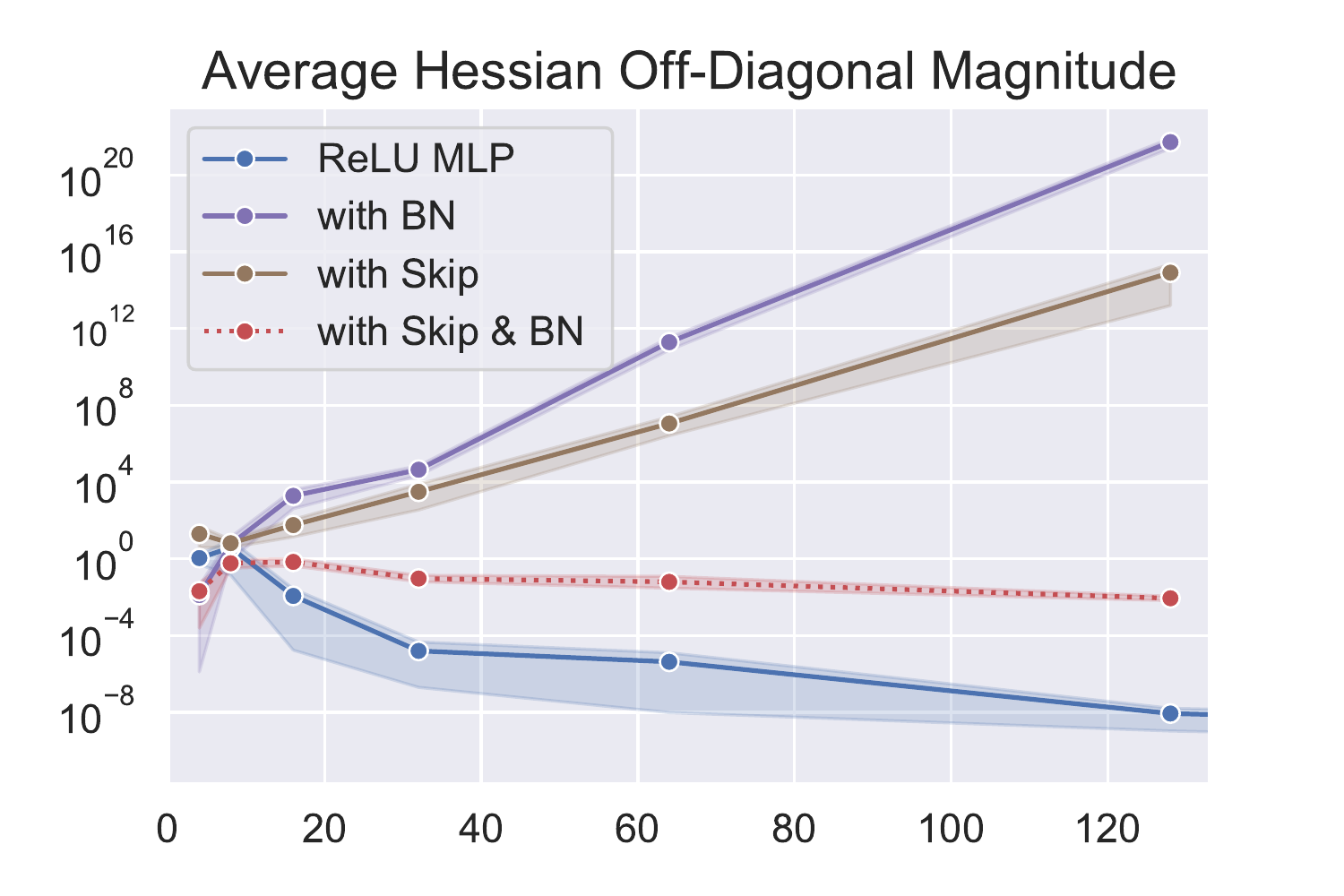}

\caption{Addendum to Fig. \ref{fig:width_effects_mlp}: Gradient and Hessian magnitudes in \textbf{ReLU MLPs} over depth. Depicted is the ReLU MLP with $d=\sqrt{L}$ and He init. Here, we add Batch Normalization and skip connections to the network. }\label{fig:mlp_vanishing_apx_BN}
\end{figure}

\begin{figure}[ht]
\centering
\includegraphics[width=0.32\textwidth]{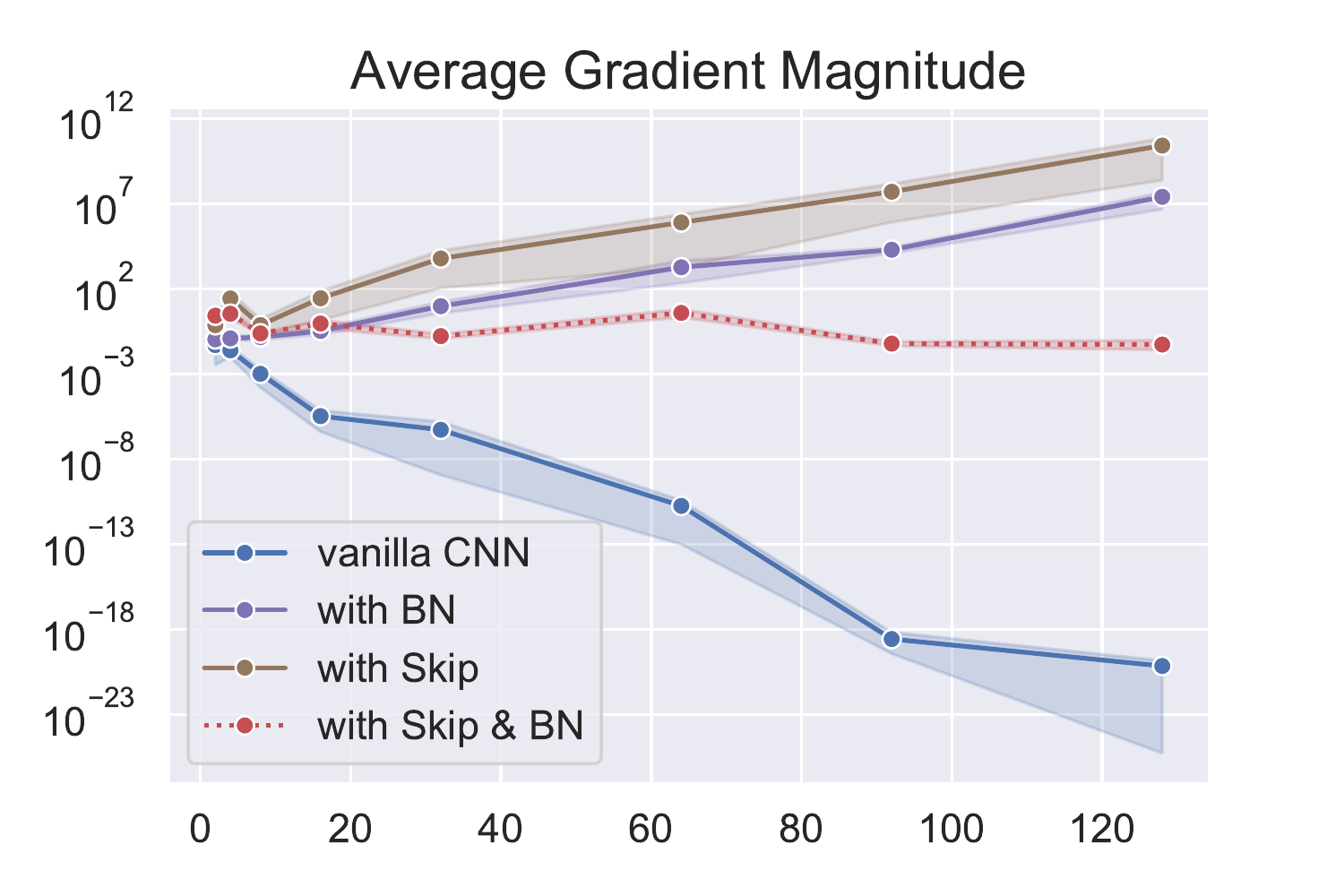}
\includegraphics[width=0.32\textwidth]{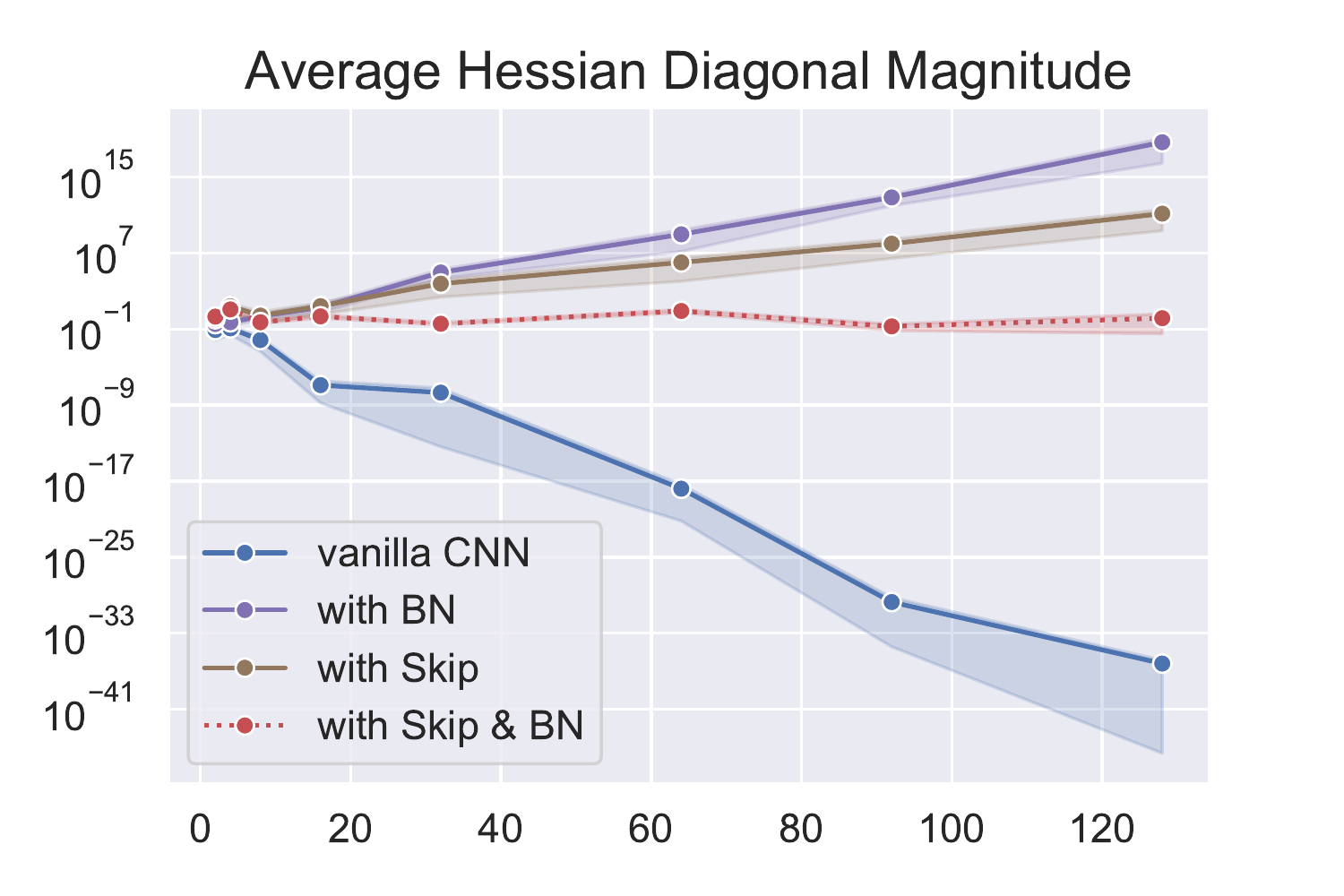}
\includegraphics[width=0.32\textwidth]{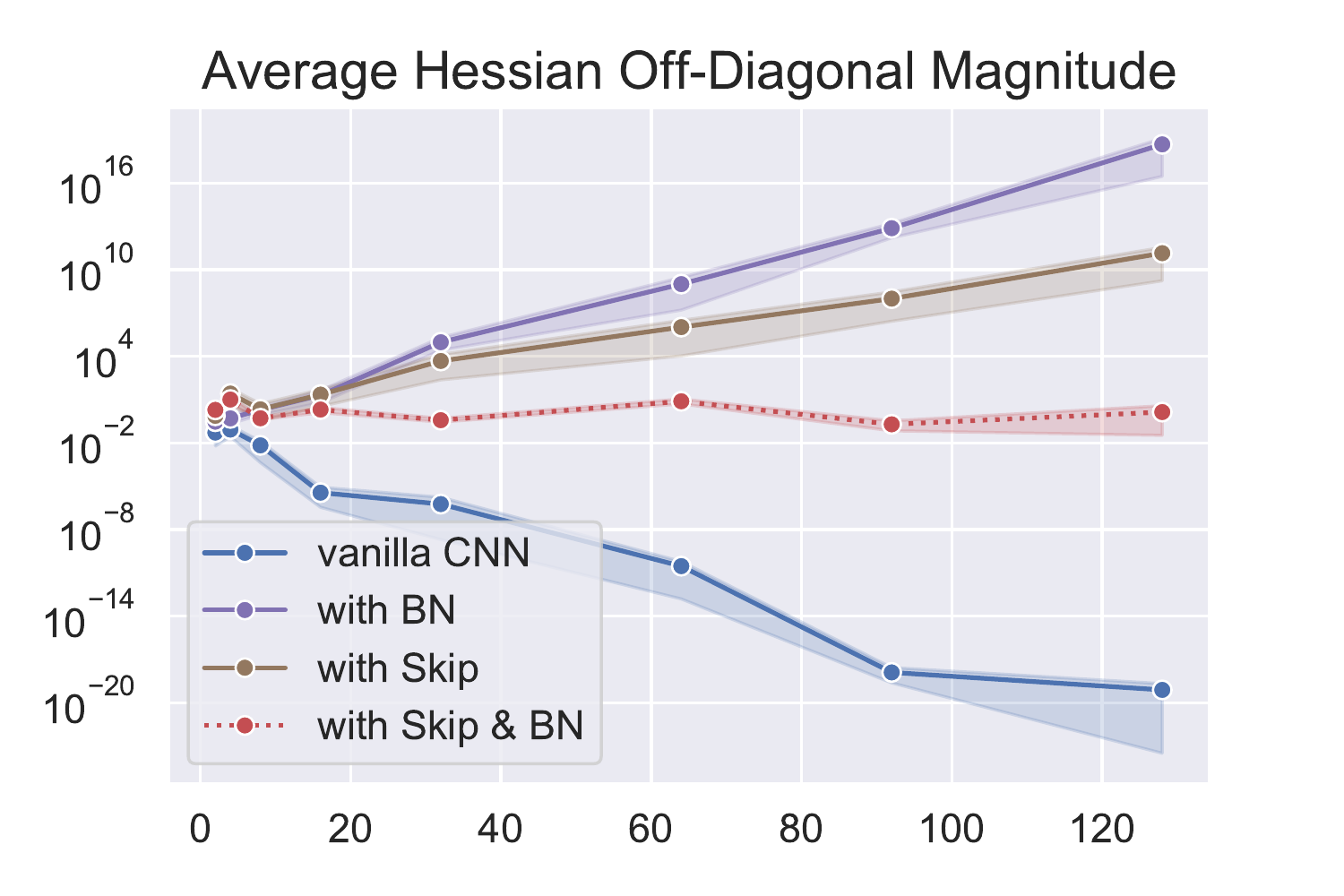}

\caption{Addendum to Fig. \ref{fig:width_effects_cnn}: Gradient and Hessian magnitudes in \textbf{FCNs} over depth. Depicted is the ReLU FCN with $d=\sqrt{L}$ at $7\times 7$ resolution initialized with He init. (Compare \ref{fig:width_effects_cnn} top right). Here, we add Batch Normalization and skip connections to the network.}\label{fig:cnn_vanishing_apx_BN}
\end{figure}

\begin{figure}[ht]
\centering
\includegraphics[width=0.32\textwidth]{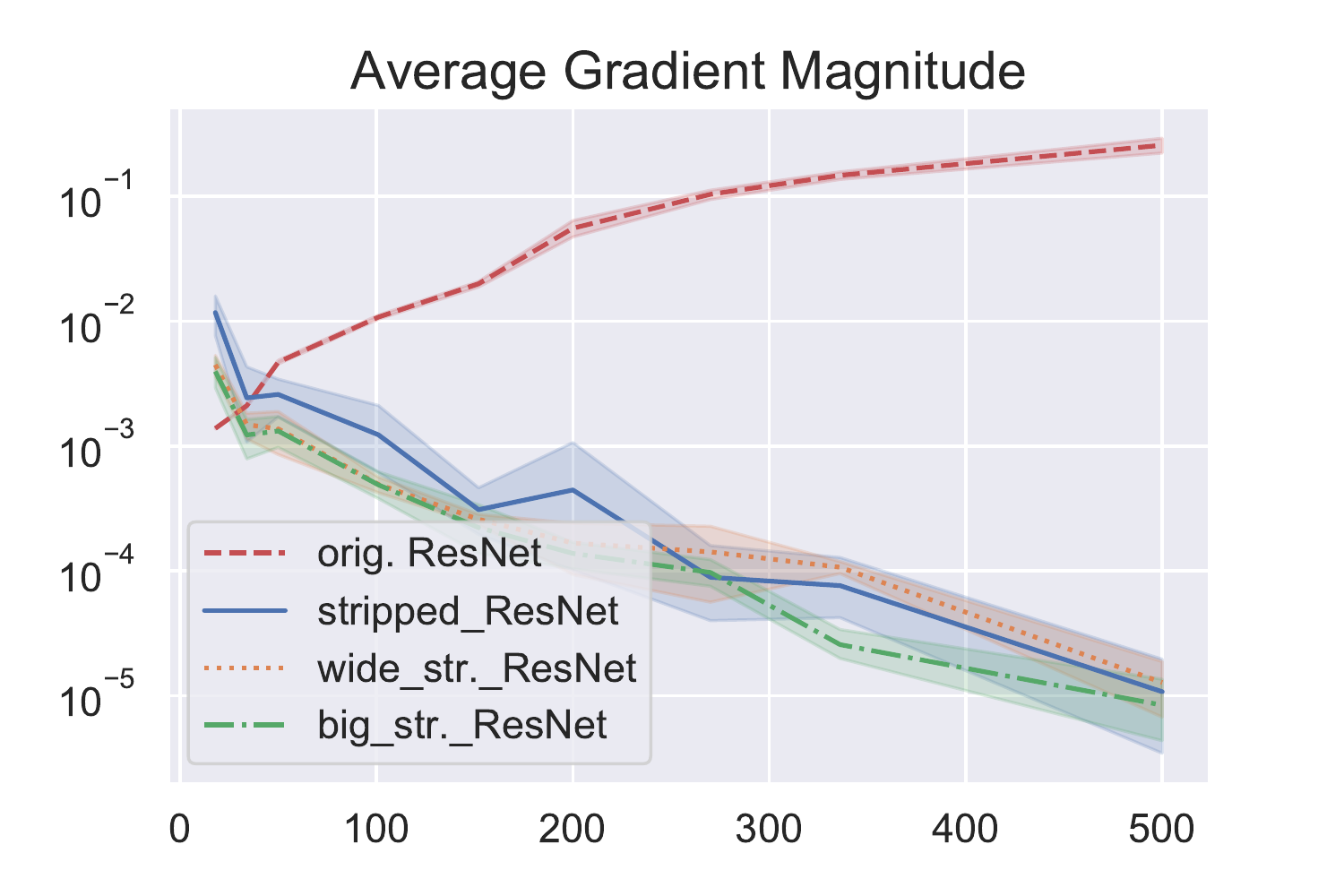}
\includegraphics[width=0.32\textwidth]{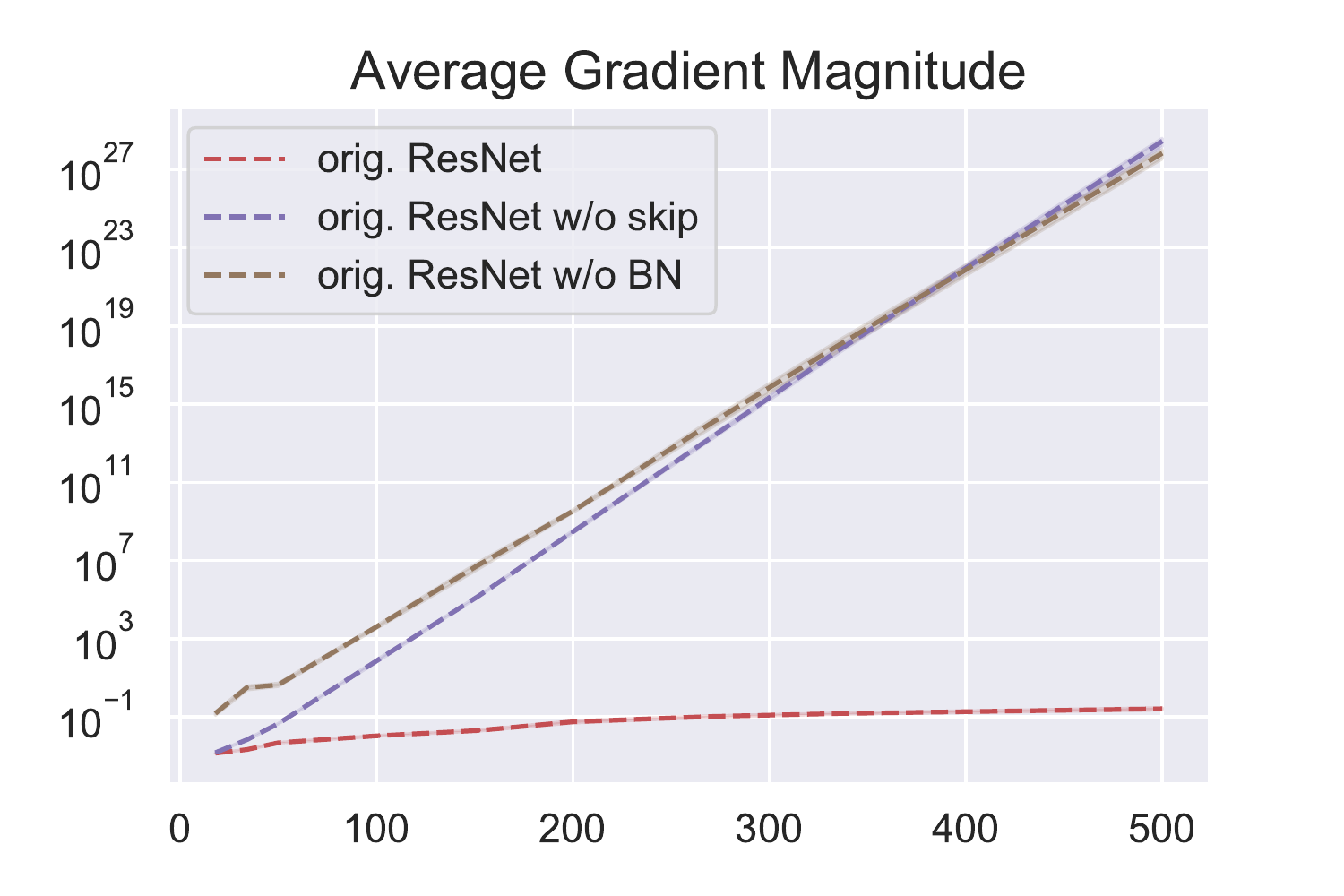}
\includegraphics[width=0.32\textwidth]{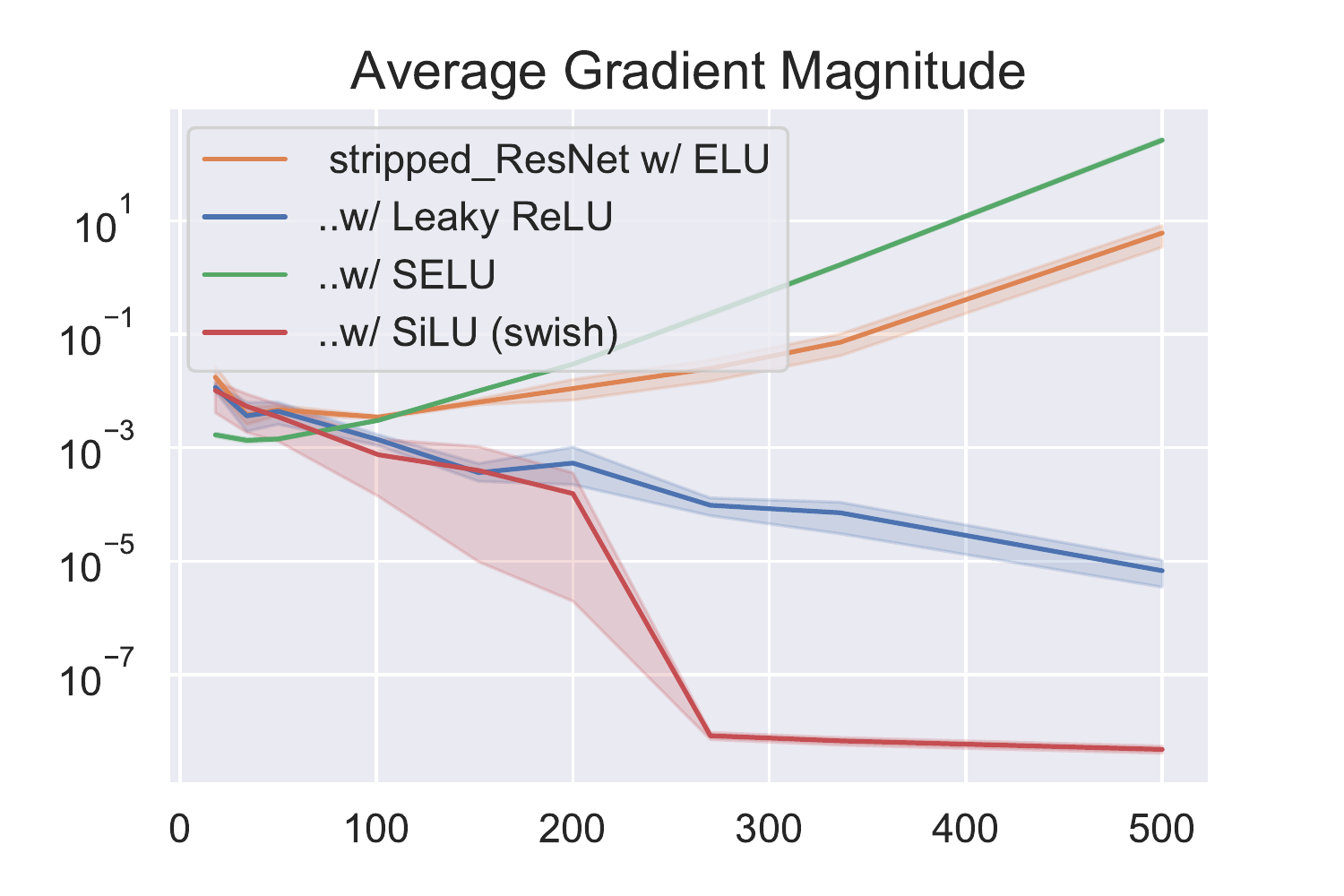}
\caption{Addendum to Fig. \ref{fig:cnn_vanishing}: \textbf{Left: Including original ResNet} architecture (i.e. with BN and skip connections). \textbf{Middle:} original ResNet plus version with just BN or Skip connections.\textbf{Right:} Effect of different activations functions in \textbf{stripped ResNets}.} \label{fig:cnn_vanishing_apx}
\end{figure}

 \begin{figure}[ht]
\centering
\includegraphics[width=\textwidth]{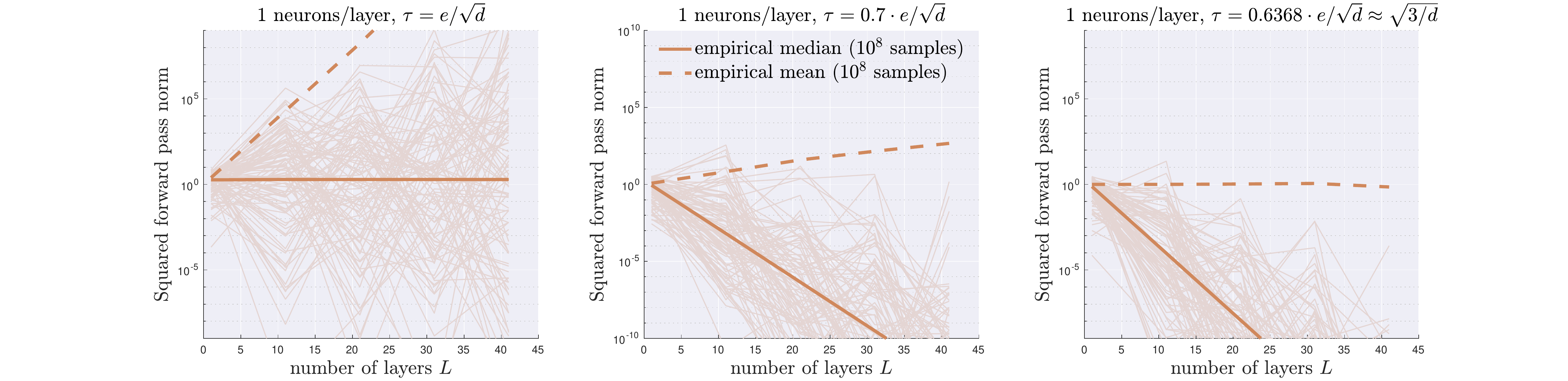}
\includegraphics[width=\textwidth]{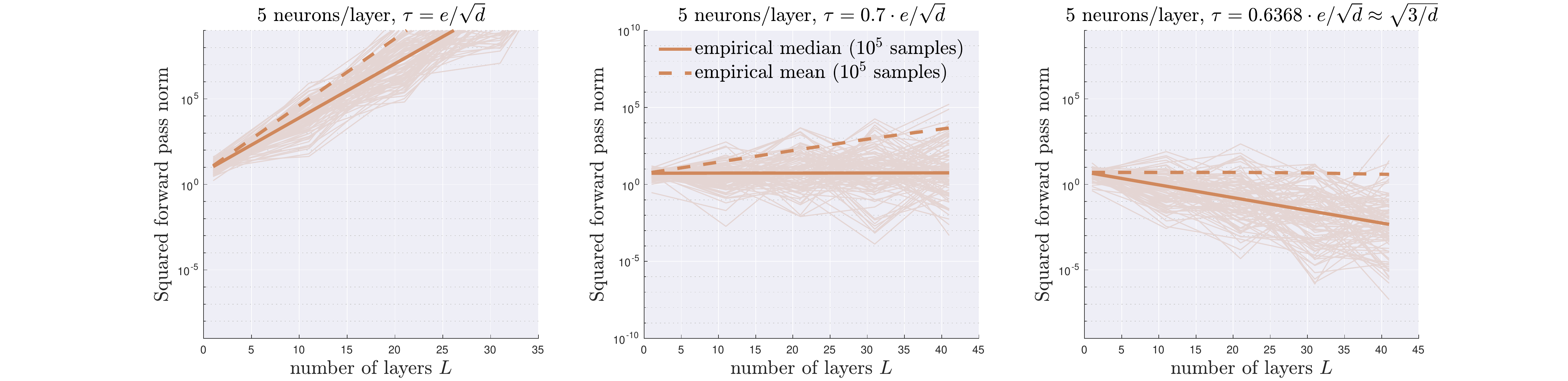}
\includegraphics[width=\textwidth]{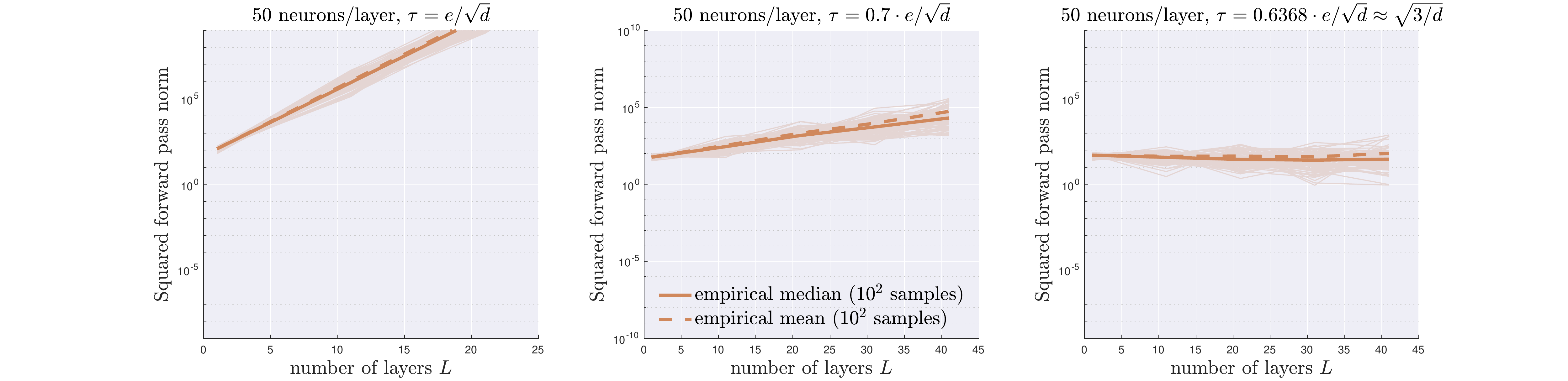}
\caption{Behavior of the variable $\|\Wm^L\Wm^{L-1}\cdots \Wm^1\mathbf{1}\|^2_2$. Entries randomly sampled $\mathcal{U}([-\tau,\tau])$. Only $50$ samples are shown, but $10^6$ are used to approximate population quantities. The expectation is by rare events, and is drastically different from the median if $d\ll L$, as shown in Thm.~\ref{thm:d_inf}.}
\label{fig:many_widths_effect}
\end{figure}

\begin{figure}[ht]
\includegraphics[width=.25\textwidth]{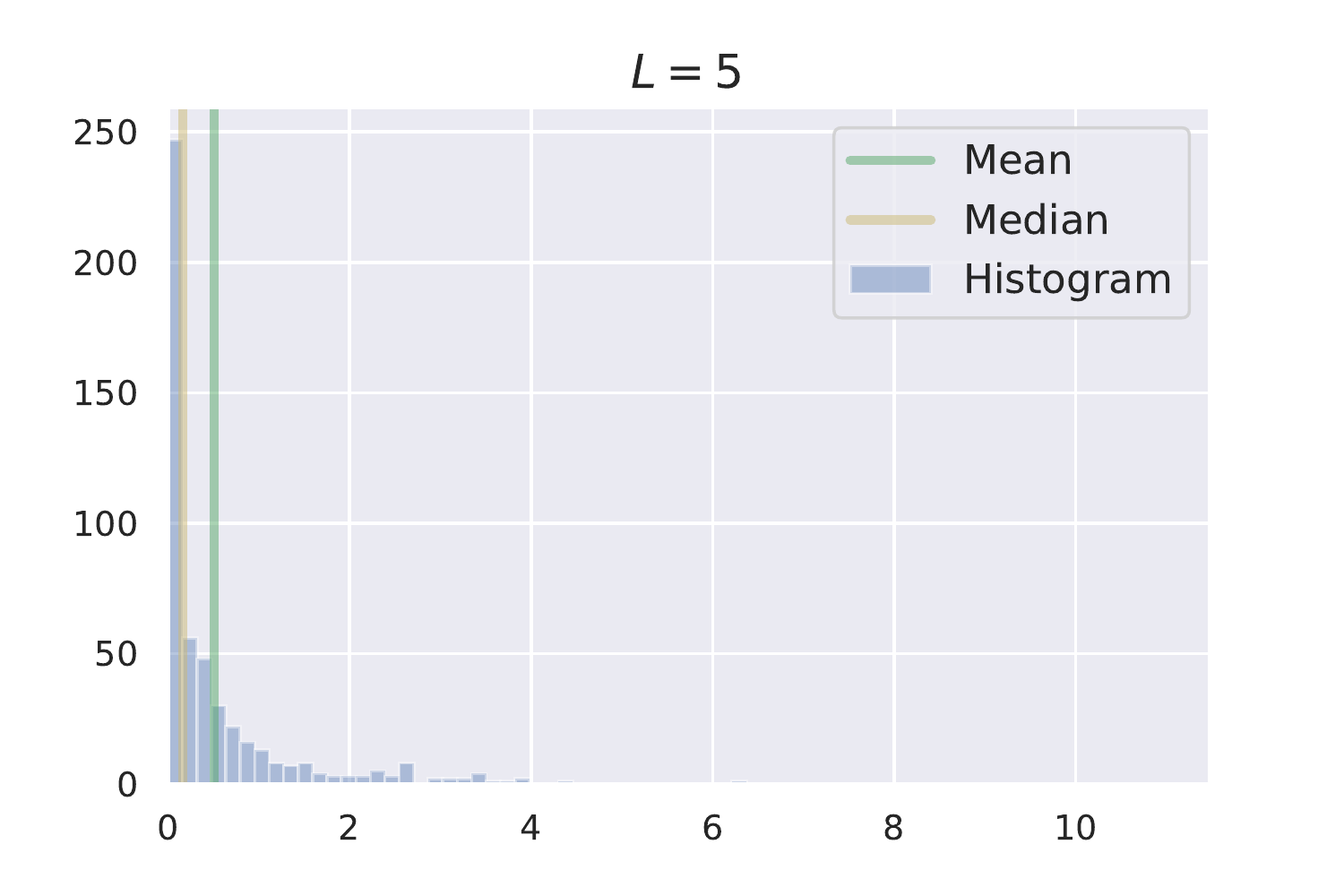}\hfill
\includegraphics[width=.25\textwidth]{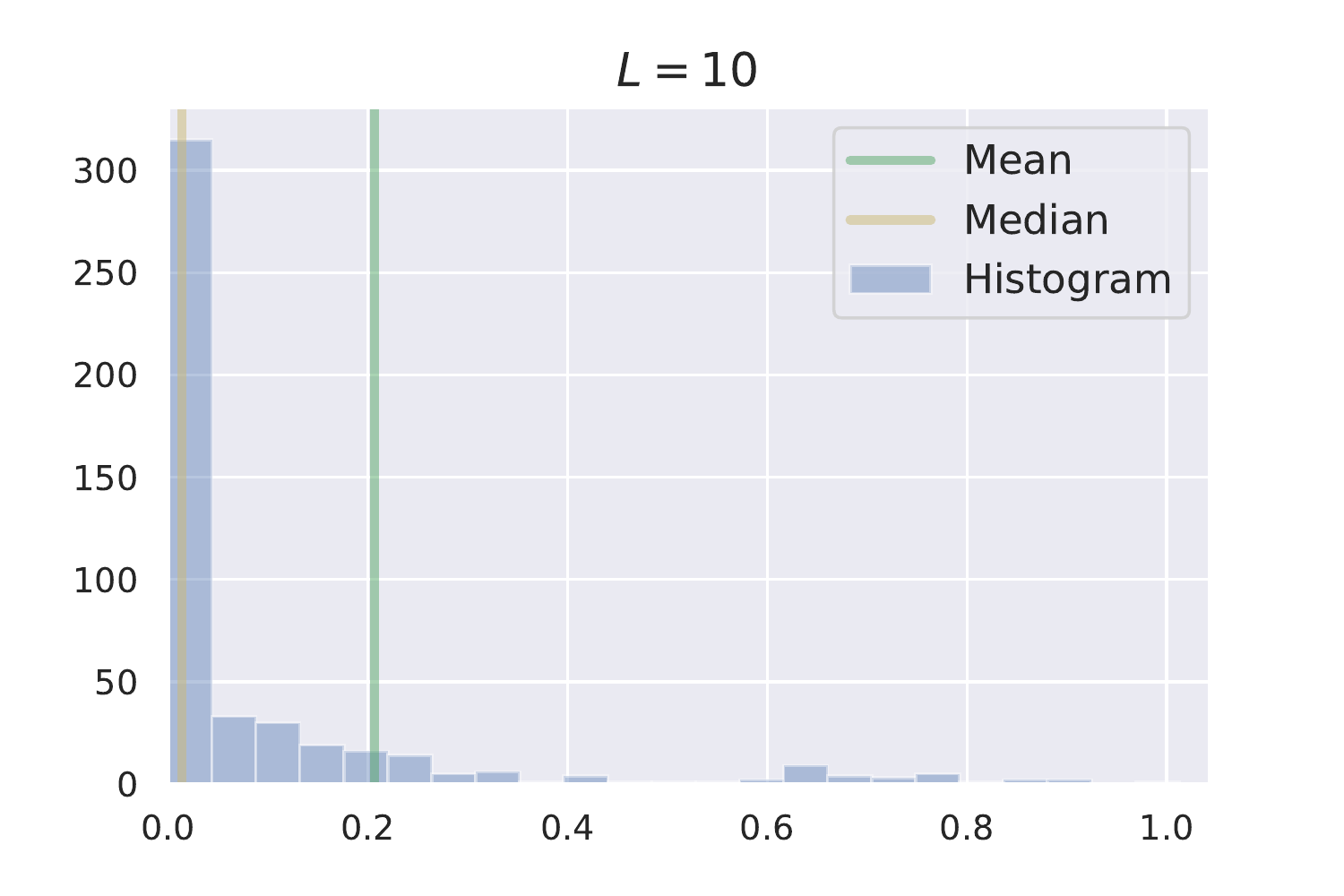}\hfill
\includegraphics[width=.25\textwidth]{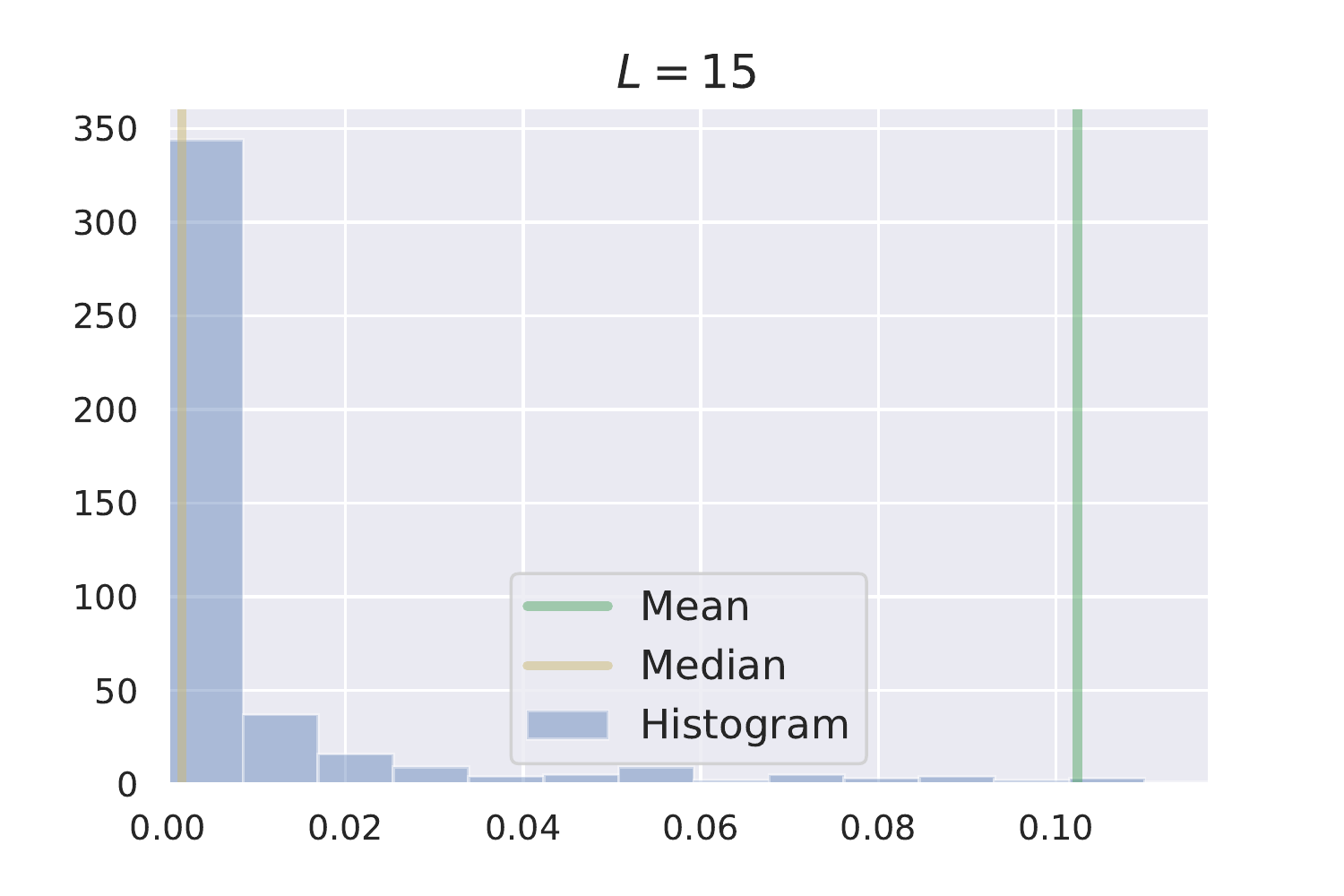}\hfill
\includegraphics[width=.25\textwidth]{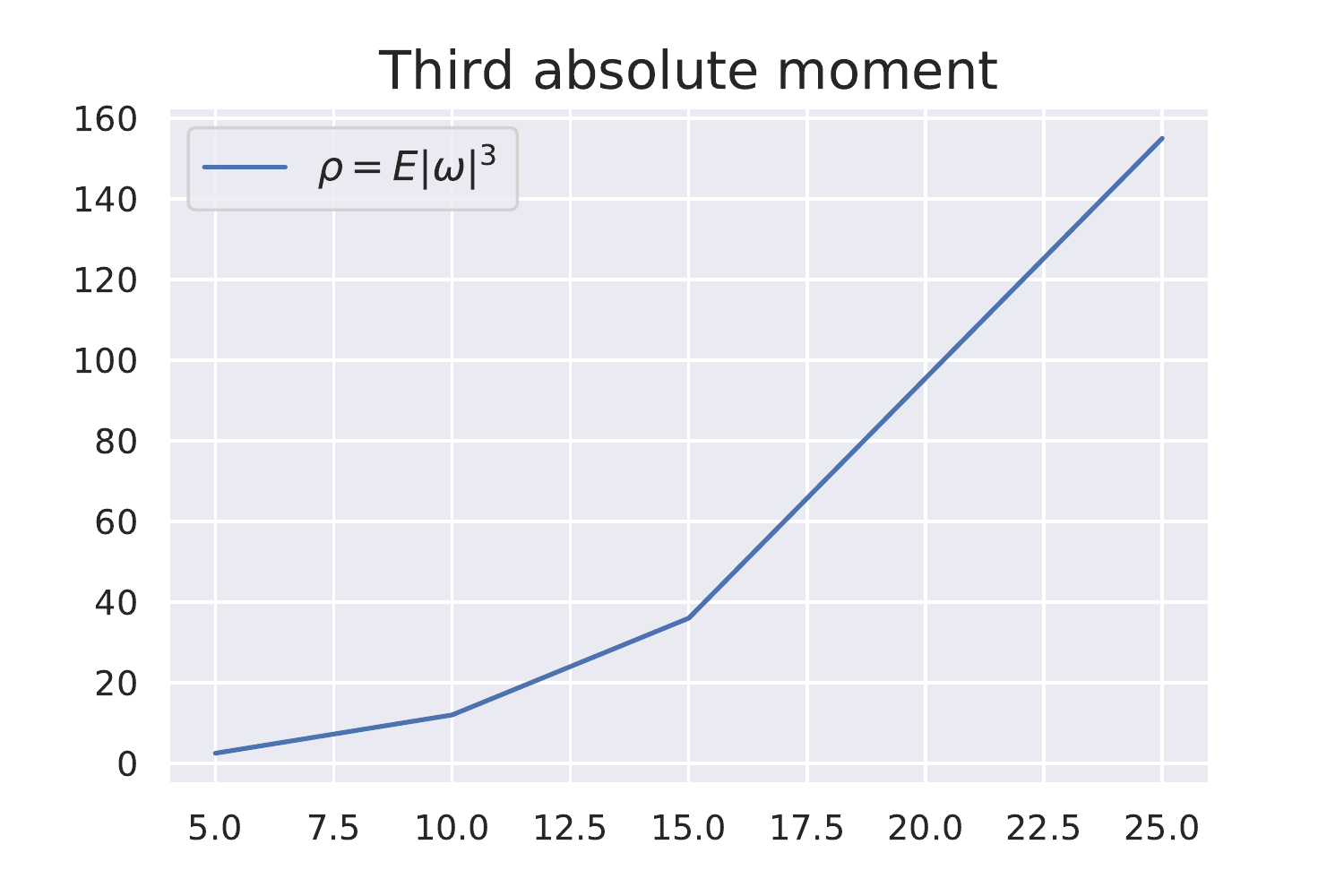}
\caption{Empirical distribution of products of an increasing number ($L$) of random uniforms. As can be seen the distribution becomes more fat-tailed as $L$ increases and the mean and median grow further apart ($x$-axis is capped at 75 times median in all plots). The last plot shows the increase in the third absolute moment over $L$.}
\label{fig:dist}
\end{figure}

\section{Analysis of deep linear and ReLU Networks}\label{sec:proof}
\textit{Note: the fundamental result of this Section~(Thm.~\ref{thm:forward_pass}) is checked numerically in Figure~\ref{fig:verif_num_forward_pass}.}

\subsection{Notation and fundamental properties}

\paragraph{Notation.}
\begin{figure}[h]
    \centering
    \includegraphics[width=0.6\textwidth]{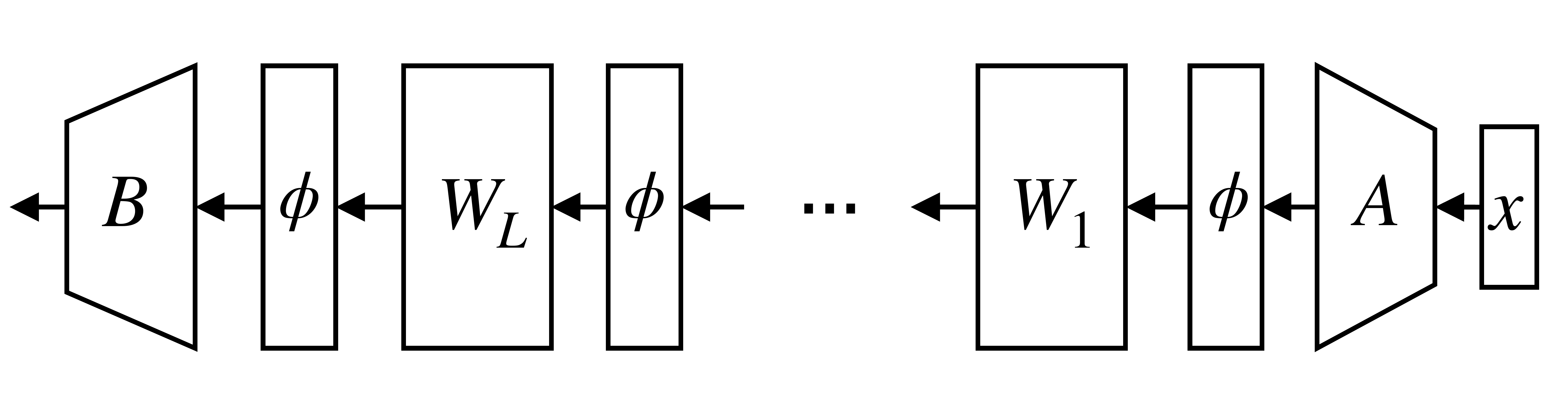}
\end{figure}
In this section, we use neural network notations similar to the one in~\citep{allen2019convergence}. In particular,
\begin{align}
F(\x) := \Bm \Dm^{L} \M^{L:1}_\phi\Am\x, \quad \M_\phi^{L:1} :=\M^L\Dm^{L-1}\M^{L-1} \cdots \M^2\Dm^1\M^1\Dm^0,
\end{align}
where $F(\x)$ constitutes the forward pass of a given input $\x\in\mathbb{R}^{d_{in}}$, $\Am \in \mathbb{R}^{d\times d_{in}}$,$\Bm \in \mathbb{R}^{d_{out}\times d}$, and $\M^\ell \in \R^{d \times d}, \forall \ell=1,\ldots,L$ . $\Dm^\ell$ is the diagonal matrix of activation gates w.r.t the non-linearity $\phi$ at layer $\ell$, which we consider to be either $\phi(x)=x$ (linear networks) or $\phi(x)=\max\{x,0\}$ (ReLU networks). Finally, we denote by $\a^\ell$  and $\h^\ell$ the pre- and post activations in layer $\ell$ respectively, i.e. for example $\a^1=\Am \x$ and $\h^1=\Dm^0\a^1$.

\textbf{Fundamental properties of activations and preactivations at each layer}\\

\begin{tcolorbox}
\begin{restatable}[Fundamental properties of activations and preactivations at each layer]{lemma}{symmetric_lemma}
\label{lemma:symmetric_lemma}
Let the entries of $\Am$, $\Bm$ and each $\W^\ell$ be i.i.d. samples from a symmetric distribution around 0 with finite moments, and variance $\sigma^2$. Then for any \underline{fixed} input $\x$, we have
\begin{enumerate}
    \item At each layer, entries of the preactivation vector are integrable and have a distribution symmetric around zero. 
    \item For ReLU networks, at each layer, the entries of the activation vector are non-zero with probability $1/2$.
    \item At each layer, in both the ReLU and the linear case, the preactivation and the activation vectors have uncorrelated squared entries.
\end{enumerate}
\end{restatable}
\end{tcolorbox}
\begin{proof}Recall that the preactivation at each layer is $\a^{\ell+1} = \W^\ell\h^\ell$. Clearly, since $\W^\ell = -\W^\ell$ in distribution, $\a^{\ell+1} = -\a^{\ell+1}$ in distribution. From this follows also that, if a ReLU is applied, $\h^{\ell+1}>0$ with probability $1/2$. The last property to show is that squared entries of activations and preactivations are uncorrelated. Let's drop the layer index $\ell$ and pick two neurons $i \ne j$, then
\begin{equation}
    \E[(g_i)^2 (g_j)^2]=\sum_{r,s,u,v} \E\left[w_{ir}w_{is}\right] \E\left[w_{ju} w_{jv}\right] \E\left[h_r h_s  h_{u}h_{v}\right] =\sigma^4\sum_{r,u} \E\left[(h_r)^2  (h_{u})^2\right].
\end{equation}
Instead, for the single squared variables we have
\begin{equation}
    \E[(g_i)^2]=\sum_{r,s} \E \left[w_{ir}w_{is}\right] \E\left[h_r h_s\right] = \sigma^2 \sum_r\E\left[(h_r)^2\right].
\end{equation}
Therefore $\E[(g_i)^2 (g_j)^2] = \E[(g_i)^2]\E[(g_j)^2]$ if and only if $(h_i)^2$ and $(h_j)^2$ are uncorrelated. As $\x$ is fixed, this is the case at the input layer and we conclude the proof by induction on $\ell$.

Last, we show the same properties for the activations in the ReLU case. Let $d_i=\phi(g_i)/g_i=\mathbbm{1}{( g_i>0)}$ and consider the new activation $h^+_i=d_ig_i$. We start by applying the law of total expectation:
\begin{equation}\begin{aligned}\label{eq:uncorrelated_squared_entries}
        \E[(h_i^+)^2 (h_j^+)^2]= \E\left[(d_i)^2(d_j)^2(g_i)^2 (g_j)^2\right]=\frac{1}{4}\E\left[(g_i)^2 (g_j)^2 \lvert d_i,d_j=1 \right] \\= \frac{1}{4}\E\left[(g_i)^2 (g_j)^2\right] = \frac{1}{2}\E\left[(g_i)^2\right] \frac{1}{2}\E\left[(g_j)^2\right] = \E[(h_i^+)^2] \E[(h_j^+)^2],
        \end{aligned}
\end{equation}
where the third and the last equalities follow from the fact that the value of the squared preactivation is independent on the sign of the preactivation.
\end{proof}

\textbf{Statistics for the propagation through one layer}\\

\begin{tcolorbox}
\begin{restatable}[Statistics after activation function]{lemma}{relu_lemma}
\label{lemma:relu_lemma}
Let $\x$ be a symmetric random vector with uncorrelated squared entries. Let $\phi(\x) = \Dx\x$. We have
\begin{align}
    \E\|\Dx \x\|_2^2 &= p \ \E\|\x\|_2^2;\\
    \E\|\Dx \x\|_2^4 &= p^2 \E\|\x\|_2^4 + (p-p^2) \E\|\x\|_4^4;\\
    \E\|\Dx \x\|_4^4 &= p \ \E\|\x\|_4^4.
\end{align}
where $p=1$ for linear nets and $p=1/2$ for ReLU nets.
\end{restatable}
\end{tcolorbox}

\begin{proof} The first property is based on the fundamental idea in~\citep{he2015delving}. Let $d_{i}$ be the entry $(i,i)$ of $\Dx$. Then, $d_{i}$ is independent from $x_i^2$. Hence, also noting that $(d_{i})^2=d_{i}$, we have
\begin{equation}
    \E\|\Dx \x\|_2^2 = \sum_{i} \E\left[d_{i}^2 x_i^2\right] = \sum_{i} \E\left[d_{i}\right] \E\left[x_i^2\right] = p \ \E\|\x\|_2^2.
\end{equation}
where $p=\E[d_{i}]$, which is $1/2$ for ReLU nets and $1$ for linear nets, as shown in Lemma~\ref{lemma:symmetric_lemma}. The last property can be proved in the same way by noting that $d_{i}$ is independent from $x_i^4$
\begin{equation}
    \E\|\Dx \x\|_4^4 = \sum_{i} \E\left[d_{i}^4 x_i^4\right] = \sum_{i} \E\left[d_{i}\right] \E\left[x_i^4\right] = p \ \E\|\x\|_4^4.
\end{equation}
The second property is a bit more involved to prove.
\begin{equation}
    \E\|\Dx \x\|_2^4 = \E\left(\sum_{i=1}^d (d_{i}x_i)^2\right)^2 = \sum_{i=1}^d \E\left[(d_{i}x_i)^4\right] +\sum_{i\ne j}\E\left[(d_{i}x_i)^2(d_{j}x_j)^2\right].
\end{equation}
From Lemma~\ref{lemma:symmetric_lemma}, third point, we have $\E\left[(d_{i}x_i)^2(d_{j}x_j)^2\right] = \E\left[(d_{i}x_i)^2\right]\E\left[(d_{j}x_j)^2\right]$ (compare Eq.\eqref{eq:uncorrelated_squared_entries}). Hence, noting again that $d_{i} = d_{i}^2$,
\begin{align}
\E\|\Dx \x\|_2^4 &= \sum_{i=1}^d \E\left[d_{i}\right] \E\left[x_i^4\right] +\sum_{i\ne j} \E\left[d_{i}\right]\E\left[d_{j}\right]\E\left[x_i^2 x_j^2\right]\\
&=p \ \E\|\x\|^4_4 + p^2 \E\left[\|\x\|^4_2-\|\x\|^4_4\right].
\end{align}
This concludes the proof.
\end{proof}

The following corollary is of fundamental importance of understanding the properties of ReLU nets: if the input of the net is modified, the ReLU gates act as purely random Bernoulli gates. This comment can be also found in the proof of Lemma A.8 in~\citep{zou2020gradient}.
\begin{tcolorbox}
\begin{restatable}[Statistics after activation function with changed input]{corollary}{relu_lemma_spiked}
\label{lemma:relu_lemma_spiked}
In the context of Lemma~\ref{lemma:relu_lemma}, let $\va$ be a fixed vector. we have
\begin{align}
    \E\|\Dx \va\|_2^2 &= p \ \|\va\|_2^2;\\
    \E\|\Dx \va\|_2^4 &= p^2 \|\va\|_2^4 + (p-p^2) \|\va\|_4^4;\\
    \E\|\Dx \va\|_4^4 &= p \ \|\va\|_4^4.
\end{align}
where $p=1$ for linear nets and $p=1/2$ for ReLU nets.
\end{restatable}
\end{tcolorbox}

\begin{proof}
Just note that since $\Dx$ and $\va$ are independent we can basically follow the proof of Lemma~\ref{lemma:relu_lemma}, but simplified:
\begin{equation}
    \E\|\Dx \va\|_m^m = \sum_{i} \E\left[d_{i}^m \alpha_i^m\right] = \sum_{i} \E\left[d_{i}\right] \alpha_i^m = p \ \|\va\|_m^m.
\end{equation}
The second property is also easy to show compared to Lemma~\ref{lemma:relu_lemma}:
\begin{align}
    \E\|\Dx \va\|_2^4 &= \E\left(\sum_{i=1}^d (d_{i}\alpha_i)^2\right)^2\\ &= \sum_{i=1}^d \E\left[(d_{i}\alpha_i)^4\right] +\sum_{i\ne j}\E\left[(d_{i}\alpha_i)^2(d_{j}\alpha_j)^2\right]\\
    &= \sum_{i=1}^d \E\left[d_{i}\right]\alpha_i^4 +\sum_{i\ne j}\E[d_{i}]\E[d_{j}]\alpha_i^2\alpha_j^2.
\end{align}
This concludes the proof.
\end{proof}

Next, we study the change in statistics after multiplication with a random matrix.

\begin{tcolorbox}
\begin{restatable}[Statistics after multiplication with a random matrix]{lemma}{recursion_lemma}
\label{lemma:recursion_lemma}
Let $\M$ be an iid random matrix, with zero mean entries that have variance $\sigma^2$ and kurtosis $\kappa$. Let $\vxi \in \R^d$ be an arbitrary random vector (not necessarily symmetric or with uncorrelated squared entries). Then 
\begin{align}
&\E\|\W\vxi\|_2^2 = d \sigma^2 \E\|\vxi\|^2;\\
&\E \|\M \mb \vxi\|_2^4  =  d(d+2) \sigma^4 \E\|\vxi\|_2^4 + (\kappa-3)d \sigma^4 \E\|\vxi\|^4_4;\\
&\E\|\M \vxi\|_4^4   = 3 d \sigma^4 \E\|\vxi\|_2^4 + (\kappa -3) d \sigma^4 \E\|\vxi\|_4^4.
\end{align}
\end{restatable}
\end{tcolorbox}
\begin{proof} The first property is easy to show:
\begin{equation}
    \E\|\W\vxi\|_2^2 = \E \sum_{r} \left(\sum_{s, u} w_{rs}\xi_s w_{ru}\xi_u\right) = \E \sum_{r} \left(\sum_{s} w_{rs}^2 \xi_s^2 \right) = d \sigma^2 \E\|\vxi\|^2.
\end{equation}

The second and the third properties need computations.
\begin{align}
\| \M \vxi\|_2^4 & = 
\Big( \sum_{i} \Big(\sum_{r} w_{ir} \xi_{r}\Big)^2 \Big)^2 
\\
& =\sum_{i,j}
\sum_{r} w_{ir} \xi_{r}
\sum_{s} w_{is} \xi_{s}
\sum_{u} w_{ju} \xi_{u}
\sum_{v} w_{jv} \xi_{v},
\\
\| \M \vxi\|^4_4 & = 
\sum_{i} \Big( \sum_r w_{ir} \xi_r\Big)^4\\
& = 
\sum_i 
\sum_{r} w_{ir} \xi_{r}
\sum_{s} w_{is} \xi_{s}
\sum_{u} w_{iu} \xi_{u}
\sum_{v} w_{iv} \xi_{v}.
\end{align}
Taking expectations, yields
\begin{align}
\frac{\E\|\M \vxi\|_4^4}{\sigma^4} & = \underbrace{\kappa \sum_i}_{\kappa d}  \E\|\mb \vxi\|^4_4 + \underbrace{3\sum_i}_{3d} \underbrace{\sum_{r \neq s} \E\left[\xi_r^2 \xi_s^2\right]}_{\E\|\vxi\|^4_2 -\E\|\vxi\|_4^4} = 3 d \ \E\|\vxi\|_2^4 + (\kappa-3) d \  \E\|\vxi\|_4^4;
\\[2mm]
\frac{\E \|\M \vxi\|^4_2}{\sigma^4}  
& = \underbrace{\sum_{i \neq j}}_{d(d-1)} \underbrace{\E\left[\Big(\sum_{r=s} \xi_r^2\Big) \Big(\sum_{u=v} \xi_u^2 \Big)\right]}_{\E\|\vxi\|_2^4}
 + 
\underbrace{3 \sum_{i=j}}_{3d} \underbrace{\sum_{r \neq s} \E\left[\xi_r^2 \xi_s^2\right]}_{\E\|\vxi\|^4_2 -\E\|\vxi\|_4^4} 
+  \underbrace{\kappa \sum_i}_{\kappa d} \underbrace{\sum_{r=s} \E[\xi_r^4]}_{\E\|\vxi\|_4^4}
\\[1mm]
& = d(d+2) \E \|\vxi\|_2^4 + (\kappa-3)d \ \E \|\vxi\|^4_4.
\nonumber
\end{align}
The factor 3 appears because, if $4$ indices are paired in groups of two, we have a total of $3$ disjoint choices: $\{i=j, u=v\}$, $\{i=u, j=v\}$, $\{i=v, j=u\}$.
\end{proof}

\begin{tcolorbox}
\begin{restatable}[Preactivations after one (ReLU) layer]{corollary}{preactivations_relu}
\label{cor:preact_relu}
Let $\x$ be a symmetric random vector with uncorrelated squared entries and $\M$ be an iid random matrix, entries having mean zero, variance $\sigma^2$ and kurtosis $\kappa$. Then, the following formulas hold:
\begin{align}
    &\E\|\W\Dx \x\|_2^2 = d\sigma^2 p \E\|\x\|_2^2;\\
    &\E\|\W\Dx \x\|_2^4 = \left[d(d+2)\sigma^4p^2\right]\E\|\x\|_2^4 + \left[d\sigma^4 p \left(\kappa-3+(1-p)(d+2)\right)\right]\E\|\x\|^4_4;\\
    &\E\|\W\Dx \x\|_4^4 = \left[3 d \sigma^4 p^2\right]\E\|\x\|_2^4 + \left[d\sigma^4 p (\kappa -3p)\right]\E\|\x\|^4_4.
\end{align}
\end{restatable}
\end{tcolorbox}

\begin{proof}
Follows directly from Lemma~\ref{lemma:recursion_lemma} by plugging in $\vxi = \Dx \x$. The effect of the potential ReLU is then integrated out using Lemma~\ref{lemma:relu_lemma}.
\end{proof}

\subsection{Forward pass}
\begin{figure}
    \centering
    \includegraphics[width=0.2\textwidth]{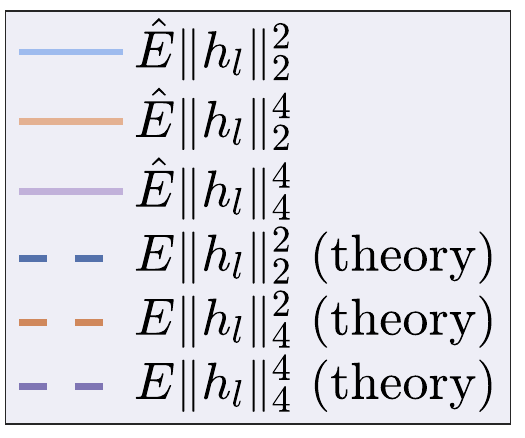}\\
    \includegraphics[width=0.81\textwidth]{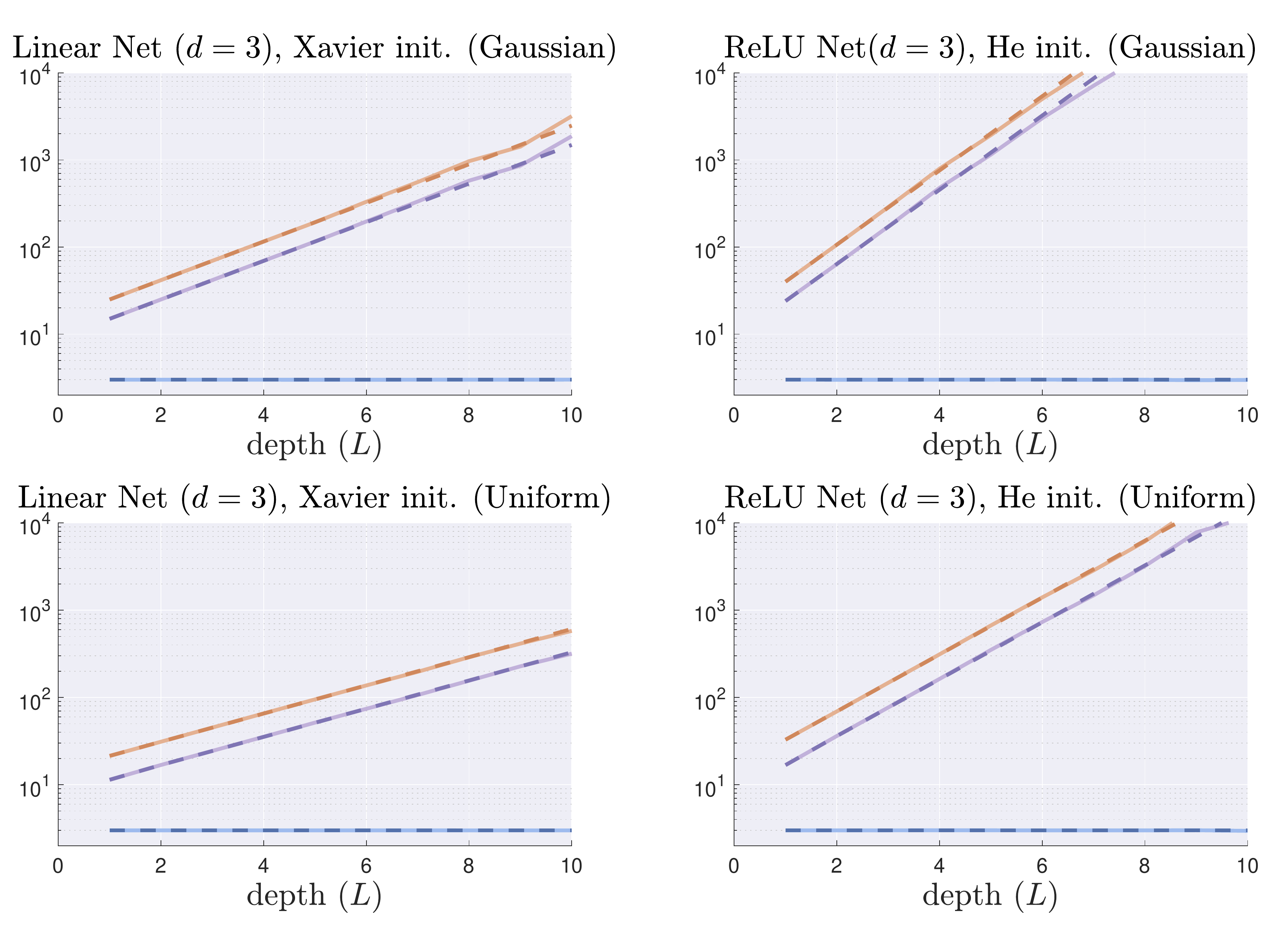} \\
    \includegraphics[width=0.81\textwidth]{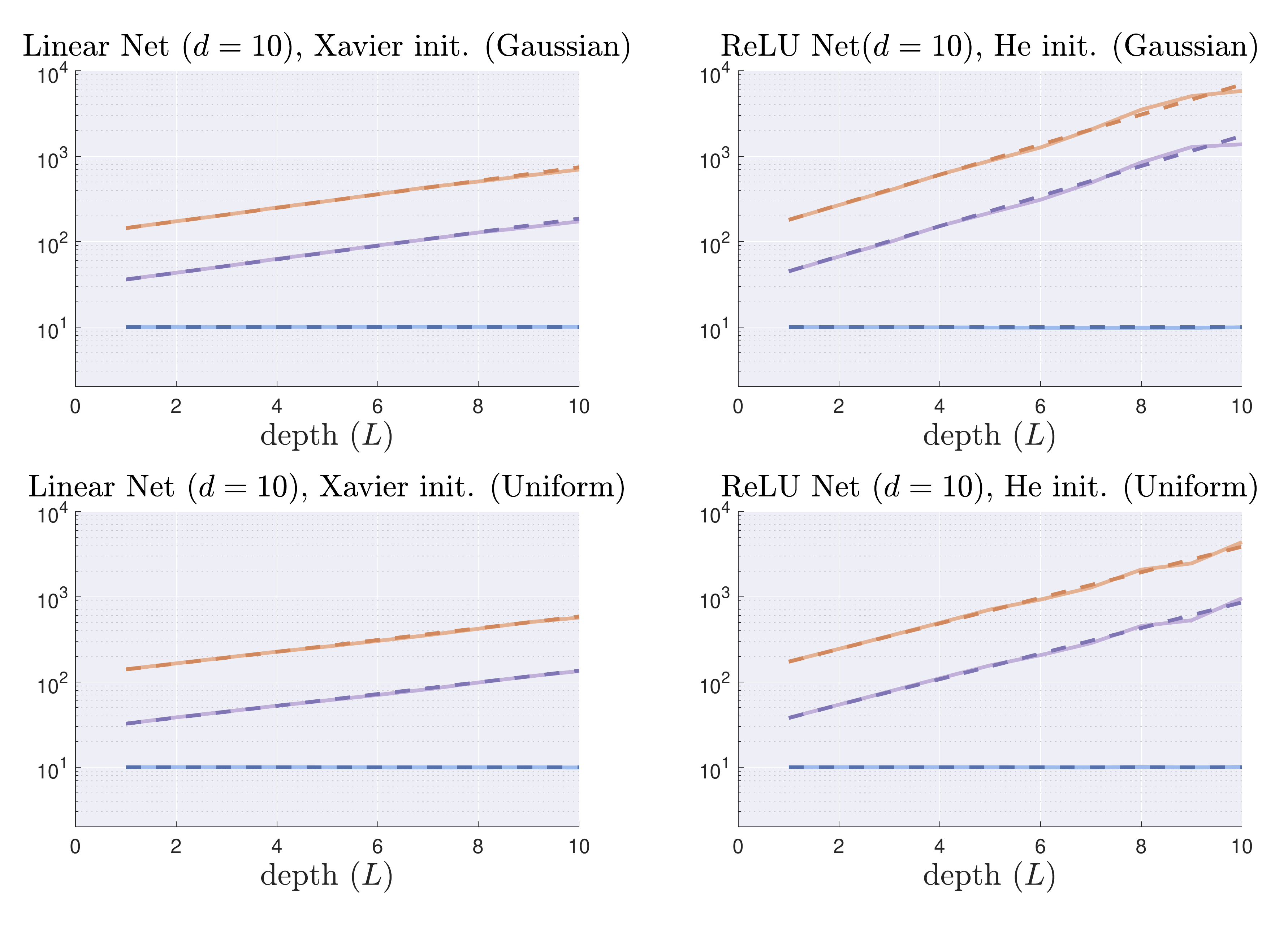} 
    \caption{Numerical validation of Theorem~\ref{thm:forward_pass} using the classical stabilizing initializations~\citep{glorot2010understanding,he2015delving}. The variance of the weights is set to $1/\sqrt{d}$ for linear nets and $2/\sqrt{d}$ for ReLU nets~(here, we used $d=3,10$). We consider a random Gaussian input and initialization of the weights with either Gaussian or uniform distribution. The theory matches the experiment~(empirical mean denoted as $\hat\E$ --- $1e5$ runs for $d=10$, $1e7$ runs for $d=3$). The results for the two initializations are similar, yet Gaussian case explodes a bit faster due to the effect of the kurtosis, which for the Gaussian is $3$ while for the uniform is $3-\frac{6}{5}$. The formula in Thm~\ref{thm:forward_pass} also perfectly predicts this tiny shift in the population quantities, \textit{confirming the correctness of our calculations}.} 
    \label{fig:verif_num_forward_pass}
\end{figure}

The next theorem, which is the fundamental tool for our analysis, is checked numerically in Figure~\ref{fig:verif_num_forward_pass}.
\begin{tcolorbox}
\begin{restatable}[Forward pass statistics]{theorem}{forward_pass}\label{thm:forward_pass}
Let $\z = \Am \x$, which is, for $\x$ fixed, symmetrically distributed and with uncorrelated squared entries. Then,
\begin{align}
\E\|\wMp{k:1}\z\|_2^2 &= (d\sigma^2 p)^k \E\|\z\|^2_2;\\
\begin{pmatrix}\E\|\wMp{k:1}\z\|_2^4\\
\E\|\wMp{k:1}\z\|_4^4\end{pmatrix} &= 
    \left(p^2 d\sigma^4\right)^k 
    \Qm^k \begin{pmatrix}\E\|\z\|_2^4\\
\E\|\z\|_4^4\end{pmatrix},\quad \label{eq:moment_recursion}
    \Qm:=\begin{pmatrix}
    d+2&\frac{\kappa-3+(1-p)(d+2)}{p}\\ 3 & \frac{\kappa-3p}{p}
    \end{pmatrix}.
\end{align}
\end{restatable}
\end{tcolorbox}

\begin{proof}
From Lemma~\ref{lemma:symmetric_lemma} we know that at initialization the entries of layer preactivations are symmetrically distributed and with uncorrelated squared entries. Hence, we can apply Corollary ~\ref{cor:preact_relu} recursively to preactivations. 
\begin{align}
    &\E\|\wMp{k:1}\z\|_2^2 = d\sigma^2 p \E\|\wMp{k-1:1}\z\|_2^2;\\
    &\E\|\wMp{k:1}\z\|_2^4 = \left[d(d+2)\sigma^4p^2\right]\E\|\wMp{k-1:1}\z\|_2^4 \\& \quad \quad \quad \quad \quad \quad \quad \quad + \left[d\sigma^4 p \left(\kappa-3+(1-p)(d+2)\right)\right]\E\|\wMp{k-1:1}\z\|^4_4;\\
    &\E\|\wMp{k:1}\z\|_4^4 = \left[3 d \sigma^4 p^2\right]\E\|\wMp{k-1:1}\z\|_2^4 + \left[d\sigma^4 p (\kappa -3p)\right]\E\|\wMp{k-1:1}\z\|^4_4.
\end{align}
The formula for $\E\|\wMp{k:1}\z\|_2^2$ directly follows, while the other two statistics evolve as a coupled dynamical system.
\end{proof}

\vspace{-2mm}
\paragraph{Simple take away.}
Consider a Gaussian initialization $\kappa=3$ in linear neural networks ($p=1$). Then the recursion matrix $\Qm$ in Eq.\eqref{eq:moment_recursion} simplifies to
\begin{align}
    \Qm:=\begin{pmatrix}
    d+2&0\\ 3 & 0
    \end{pmatrix},
\end{align}

from which follows that $
    \E\|\wMp{k:1}\z\|_2^4=  \left(d\sigma^4\right)^k \left(d+2\right)^k\E\|\z\|^2_4$. Hence, similar to the case of the neural chain depicted in Eq.~\ref{eq:chain_moments} the pair of moments $ \E\|\wMp{k:1}\Am\z\|_2^2$ and  $\E\|\wMp{k:1}\z\|_2^4$ cannot be stabilized jointly. Hence, as shown in Figure~\ref{fig:many_widths_effect}, the mean is allowed to be very different from the median by Mallows inequality~\cite{mallows1969inequalities} --- see also Thm.~\ref{thm:d_inf}. Finally, we note that the effect we just discussed grows stronger if $d\ll L$. This is also confirmed by the simulation in Figure~\ref{fig:verif_num_forward_pass}.

\begin{tcolorbox}
\begin{restatable}[Asymptotic forward pass statistics]{corollary}{forward_pass_asy}\label{cor:asy_forward}
In the context of Theorem \ref{thm:forward_pass}, as $d\to\infty$
\begin{align}
    &\E\|\wMp{k:1}\Am\x\|_2^4 \lesssim (d\sigma^2 p)^{2k}, \\
    &\E\|\wMp{k:1}\Am\x\|_4^4 \lesssim (d\sigma^2 p)^{2k},
\end{align}
where ``$\lesssim$'' denotes ``asymptotically less of equal than a multiple of''~(same as $\mathcal{O}$). 
\end{restatable}
\end{tcolorbox}
\begin{proof}
For the linear and the ReLU case we respectively have
\begin{equation}
    \Qm_{\text{linear}}=\begin{pmatrix}
    d+2&\kappa-3\\ 3 & \kappa-3
    \end{pmatrix},\quad \Qm_{\text{ReLU}}=\begin{pmatrix}
    d+2&d+2\kappa-4\\ 3 & 2\kappa-3
    \end{pmatrix}.
\end{equation}
As $d\to\infty$, these matrices behave as follows:
\begin{equation}
    \Qm^\infty_{\text{linear}}=\begin{pmatrix}
    \mathcal{O}(d)&\mathcal{O}(1)\\\mathcal{O}(1) & \mathcal{O}(1)
    \end{pmatrix},\quad \Qm^\infty_{\text{ReLU}}=\begin{pmatrix}
    \mathcal{O}(d)&\mathcal{O}(d)\\\mathcal{O}(1) & \mathcal{O}(1)
    \end{pmatrix}.
\end{equation}
The result follows.
\end{proof}

\begin{tcolorbox}
\begin{restatable}[Forward pass form a different vector]{proposition}{fp_alpha}
\label{prop:fp_alpha}
Assume that, instead of $\x$, we feed into the random net~(with activation gates computed using $\x$) a different vector $\va$. Then, all the forward pass statistics still hold.
\end{restatable}
\end{tcolorbox}

\begin{proof}
Direct consequence of Corollary~\ref{lemma:relu_lemma_spiked}, when applied iteratively as done above.
\end{proof}

\textbf{Statistics for network-dependent matrices from the forward pass}\\

\begin{tcolorbox}
\begin{restatable}[From network matrices to propagations of canonical vectors]{proposition}{matrix_to_vec}
\label{prop:matrix_to_vec}
Let $\eb_1$ be the first vector in the canonical basis. We have
\begin{align}
    &\E\|\wMp{\ell:k+1}\|^2_F = d \ \E\|\wMp{\ell:k+1} \eb_1\|^2_2\sim (d\sigma^2 p)^{\ell-k};\\
    &\E\|\wMp{\ell:k+1}\|^4_F = d^2 \ \E\|\wMp{\ell:k+1} \eb_1\|^4_2\sim (d\sigma^2 p)^{2(\ell-k)}.
\end{align}
where ``$\lesssim$'' denotes ``asymptotically less of equal than a multiple of''~(same as $\mathcal{O}$). 
\end{restatable}
\end{tcolorbox}

\begin{proof}By direct calculation
\begin{align}
    \E\|\wMp{\ell:k+1}\|^2_F = \E\left[\sum_{i=1}^d\|\wMp{\ell:k+1} \eb_i\|^2_2 \right] = d \ \E\|\wMp{\ell:k+1} \eb_1\|^2_2\sim (d\sigma^2 p)^{\ell-k}.
\end{align}
The second last equality follows from the isotropic structure of $\wMp{\ell:k}$ and the last from Prop.~\ref{prop:fp_alpha}.
\begin{align}
    \E\|\wMp{\ell:k+1}\|^4_F = \E\left[\left(\sum_{i=1}^d\|\wMp{\ell:k+1} \eb_i\|^2_2\right)^2 \right] = \E\left[\sum_{ij}\|\wMp{\ell:k+1} \eb_i\|^2_2\|\wMp{\ell:k+1} \eb_j\|^2_2\right].
\end{align}
Since $\|\wMp{\ell:k+1} \eb_j\|^2_2 = \|\wMp{\ell:k+1} \eb_i\|^2_2$ in distribution, $\E\|\wMp{\ell:k+1}\|^4_F = d^2 \E\|\wMp{\ell:k+1} \eb_i\|^4_2$. The asymptotic statements hold thanks to Proposition~\ref{prop:fp_alpha}.
\end{proof}

\subsection{Gradient}
We consider the loss
\begin{align} 
\Ls_{\x,\y}(\W) = \frac 12 \|\y - \Bm \Dm^L \wMp{L:1}\Am\x\|^2.
\end{align}
As in~\citep{allen2019convergence}, by noting $\tilde\Bm:=\Bm \Dm^L$ and $\z = \Am \x$ we get
\begin{align}
\jacobi{\Ls}{\wM k} &=  \wMp{k+1:L} \tilde\Bm^\top [\tilde\Bm \wMp{L:1}\z- \y] \z^\top \wM{1:k-1}\\
&=  \underbrace{\wMp{k+1:L} \tilde\Bm^\top \tilde\Bm \wMp{L:1}\z\z^\top \wM{1:k-1}}_{\boldsymbol\partial\Ls_k^1} - \underbrace{\wMp{k+1:L}\tilde\Bm^\top \y \z^\top \wM{1:k-1}}_{\boldsymbol\partial\Ls_k^2},
\end{align}
with $\wMp{k+1:L}:=\left(\wMp{L:k+1}\right)^\top$. By the triangle inequality, we have 
\begin{align}
    \left\|\jacobi{\Ls}{\wM k}\right\|_F\le \|\boldsymbol\partial\Ls_k^1\|_F + \|\boldsymbol\partial\Ls_k^2\|_F, 
\end{align}
therefore we can bound each term individually. Let $\E^p[\cdot]$ be the $p$-th power of $\E[\cdot]$.
\begin{tcolorbox}
\begin{restatable}[Bounding gradients with forward passes]{proposition}{grad_pass}\label{prop:grad_bound}
We have
\begin{align}
    \E\|\boldsymbol\partial\Ls_k^1\|_F &\le \E^{1/2}\left[\|\tilde\Bm\wMp{L:k+1}\|^2_2\right] \E^{1/4}\left[\|\tilde\Bm\wMp{L:1}\z\|^4_2\right] \E^{1/4}\left[\|\wMp{k-1:1} \z\|^4_2\right];\\
    \E\|\boldsymbol\partial\Ls_k^2\|_F &\le \E\left[\|\y^\top\tilde\Bm\wMp{L:k+1}\|^2_2\right] \E\left[\|\wMp{k-1:1} \z\|^2_2\right].
\end{align}
\end{restatable}
\end{tcolorbox}

\begin{proof}The bounds follow from submultiplicativity of Frobenius norm and Cauchy-Schwarz inequality --- applied possibly twice.
\begin{align}
    \E\|\boldsymbol\partial\Ls_k^1\|_F &= \E\|\wMp{k+1:L} \tilde\Bm^\top \tilde\Bm \wMp{L:1}\z\z^\top  \wMp{1:k-1}\|_F\\
    &\le \E\left[\|\wMp{k+1:L} \tilde\Bm^\top \|_F \|\tilde\Bm \wMp{L:1}\z \|_F \|\z^\top \wMp{1:k-1}\|_F\right]\\
    &= \E\left[\|\tilde\Bm\wMp{L:k+1}\|_F \|\tilde\Bm\wMp{L:1}\z\|_F \|\wMp{k-1:1} \z\|_F\right]\\
    &\le \E^{1/2}\left[\|\tilde\Bm\wMp{L:k+1}\|^2_F\right] \E^{1/2}\left[\|\tilde\Bm\wMp{L:1}\z\|^2_F \|\wMp{k-1:1} \z\|^2_F\right]\\
    &\le \E^{1/2}\left[\|\tilde\Bm\wMp{L:k+1}\|^2_F\right] \E^{1/4}\left[\|\tilde\Bm\wMp{L:1}\z\|^4_F\right] \E^{1/4}\left[\|\wMp{k-1:1} \z\|^4_F\right].\\
    &\nonumber\\
     \E\|\boldsymbol\partial\Ls_k^2\|_F &= \E\|\wMp{k+1:L} \tilde\Bm^\top \y \z^\top \wMp{1:k-1}\|_F\\
     &\le \E\left[\|\wMp{k+1:L} \tilde\Bm^\top \y\|_F\|\z^\top \wMp{1:k-1}\|_F\right]\\
     &\le \E^{1/2}\left[\|\wMp{k+1:L} \tilde\Bm^\top \y\|^2_F\right] \E^{1/2}\left[\|\z^\top \wMp{1:k-1}\|^2_F\right]\\
     &= \E^{1/2}\left[\|\y^\top\tilde\Bm\wMp{L:k+1}\|^2_F\right] \E^{1/2}\left[\|\wMp{k-1:1} \z\|^2_F\right].
\end{align}
We conclude by noting that the Frobenius norm is the $2$-norm for vectors.
\end{proof}

\begin{tcolorbox}
\begin{restatable}[Bounding gradients with forward passes, wide net]{proposition}{grad_pass_inf} As $d\to\infty$, 
\begin{align}
    \E\|\boldsymbol\partial\Ls_k^1\|_F &\lesssim (p\sigma^2 d)^{\frac{2L-1}{2}},\\
    \E\|\boldsymbol\partial\Ls_k^2\|_F &\lesssim (p\sigma^2 d)^{\frac{L-1}{2}},
\end{align}
where ``$\lesssim$'' denotes ``asymptotically less of equal than a multiple of''~(same as $\mathcal{O}$).  Hence, we have
\begin{align}
    \E\left\|\jacobi{\Ls}{\wM k}\right\|_F &\lesssim (p\sigma^2 d)^{\frac{L-1}{2}} \quad \quad \text{if }\quad  (p\sigma^2 d)\le 1 \quad \text{(vanishing-stable regime)}.\\
    \E\left\|\jacobi{\Ls}{\wM k}\right\|_F &\lesssim (p\sigma^2 d)^{\frac{2L-1}{2}} \quad \quad \text{if }\quad  (p\sigma^2 d)\ge 1 \quad \text{(exploding regime)}.
\end{align}
\end{restatable}
\end{tcolorbox}

\begin{proof}
Simple application of Corollary~\ref{cor:asy_forward} and Proposition~\ref{prop:matrix_to_vec} to the bounds in Proposition~\ref{prop:grad_bound}.
\end{proof}

\subsection{Hessian}
The Hessian of a linear DNN can be split into two block matrices, where each block has a Kronecker product structure. We can apply the product rule to the gradient and consider $\wM \ell$ and $\wMt \ell$ ($\ell>k$) as two distinct matrices: the block $(k,\ell)$ of the Hessian matrix is:
$$\underbrace{\frac{\partial^2 \Ls}{\partial \wM k \partial \wM \ell}}_{\mb H^{k\ell}_1}+ \underbrace{\frac{\partial^2 \Ls}{\partial \wM k {\partial \wMt \ell}}}_{\mb H^{k\ell}_2}\underbrace{\frac{\partial(\wMt \ell)}{\partial\wM \ell}}_{\mathbb{T}},$$
where $\mathbb T$ is the matrix transpose tensor. Recall that
\begin{align}
\jacobi{\Ls}{\wM k} =  \wMp{k+1:L} \tilde\Bm^\top [\tilde\Bm \wMp{L:1}\z- \y] \z^\top \wM{1:k-1}.
\end{align}
By using the simple rule $\frac{\partial \mb E \W \mb F}{\partial \W} = \mb F^\top \otimes \mb E$, we get
\begin{align}
\mb H^{k\ell}_1  &= \wMp{k-1:1}\z\z^\top \wMp{1:\ell-1} \; \otimes \; \wMp{k+1:L} \tilde\Bm^\top\tilde\Bm \wMp{L:\ell+1};\\
\mb H^{k\ell}_2 &= \wMp{k-1:1} \z\left(\z^\top \wMp{1:L}\tilde\Bm^\top -\y^\top\right)\tilde\Bm\wMp{L:\ell+1}\otimes \wMp{k+1:\ell-1}\\
&= \underbrace{\wMp{k-1:1} \z\z^\top \wMp{1:L}\tilde\Bm^\top\tilde\Bm\wMp{L:\ell+1}\otimes \wMp{k+1:\ell-1}}_{\mb H^{k\ell}_{21}} - \underbrace{\wMp{k-1:1} \z\y^\top\tilde\Bm\wMp{L:\ell+1}\otimes \wMp{k+1:\ell-1}}_{\mb H^{k\ell}_{22}},\nonumber
\end{align}
Note that if instead $k=\ell$, $\mb H^{kk}=\mb H^{kk}_1$.
\begin{tcolorbox}
\begin{restatable}[Bounding the Hessian with forward passes]{proposition}{hess_pass}\label{prop:hess_to_forward}
It is possible to bound the Hessian with statistics only on the forward pass.
\end{restatable}
\end{tcolorbox}
\vspace{-2mm}
\begin{proof} We simply apply the Cauchy-Schwarz inequality twice for each term, using also the Frobenius norm formula for the Kronecker product and norm submultiplicativity.
\begin{align}
    \E\|\mb H^{k\ell}_1\|_F  &= \E\left\|\wMp{k-1:1} \z\z^\top \wMp{1:\ell-1} \; \otimes \; \wMp{k+1:L} \tilde\Bm^\top\tilde\Bm \wMp{L:\ell+1}\right\|_F\\
    &\le \E^{1/2}\left\|\wMp{k-1:1}\z\z^\top \wMp{1:\ell-1}\right\|^2_F \E^{1/2}\left\|\wMp{k+1:L} \tilde\Bm^\top\tilde\Bm \wMp{L:\ell+1}\right\|^2_F\\
    &\le \E^{1/4}\left\|\wMp{k-1:1}\z\right\|^4_F \E^{1/4}\left\|\wMp{\ell-1:1}\z\right\|^4_F \E^{1/4}\left\|\tilde\Bm \wMp{L:\ell+1}\right\|^4_F \E^{1/4}\left\|\tilde\Bm \wMp{L:k+1}\right\|^4_F.
\end{align}

\begin{align}
\E\|\mb H^{k\ell}_{22}\|_F &\le \E^{1/2}\left\|\wMp{k-1:1} \z\y^\top\tilde\Bm\wMp{L:\ell+1}\right\|^2_F\E^{1/2}\left\|\wMp{k+1:\ell-1}\right\|^2_F\\
&\le \E^{1/4}\left\|\wMp{k-1:1} \z\right\|^4_F\E^{1/4}\left\|\y^\top\tilde\Bm\wMp{L:\ell+1}\right\|^4_F\E^{1/2}\left\|\wMp{\ell-1:k+1}\right\|^2_F.
&\\
&\nonumber\\
\E\|\mb H^{k\ell}_{21}\|_F &\le \E^{1/2}\left\|\wMp{k-1:1} \z\z^\top \wMp{1:L}\tilde\Bm^\top\tilde\Bm\wMp{L:\ell+1} \right\|^2_F\E^{1/2}\left\|\wMp{k+1:\ell-1}\right\|^2_F\\
&\le \E^{1/4}\left\|\wMp{k-1:1} \z\z^\top \wMp{1:\ell}\wMp{\ell+1:L}\tilde\Bm^\top\tilde\Bm\wMp{L:\ell+1} \right\|^2_F\E^{1/4}\left\|\wMp{\ell-1:k+1}\right\|^2_F.
\end{align}
 Note that the last term is not simplified completely, but unfortunately a simple iterated Cauchy-Schwarz splitting would lead to quantities with high exponents (eighth moment). Hence, we need to take a more complex approach. First, we split between terms which do not share weights.
 \begin{align}
     \E\left[\left\|\wMp{k-1:1} \z\z^\top \wMp{1:\ell}\right\|^2_F\left\|\wMp{\ell+1:L}\tilde\Bm^\top\tilde\Bm\wMp{L:\ell+1} \right\|^2_F\right].
 \end{align}
 Using the law of total expectation, the last expression becomes
 \begin{align}
     \E\left[ \ \ \E\left[ \ \ \left\|\wMp{k-1:1} \z\z^\top \wMp{1:\ell}\right\|_F\left\|\wMp{\ell+1:L}\tilde\Bm^\top\tilde\Bm\wMp{L:\ell+1} \right\|^2_F \ \ \bigg\lvert  \ \ \mathcal{F}_\ell \ \ \right] \ \ \right],
 \end{align}
 where $\mathcal{F}_\ell$ is the information until layer $\ell$. Using the same reasoning as Corollary~\ref{lemma:relu_lemma_spiked} and Proposition~\ref{prop:fp_alpha}, it is easy to realize that fixing the preactivation a separate integration of the second term in the product. In particular, the expression becomes
  \begin{align}
     \E\left[\left\|\wMp{k-1:1} \z\z^\top \wMp{1:\ell}\right\|_F \right] \ \cdot \ \E\left[ \ \ \left\|\wMp{\ell+1:L}\tilde\Bm^\top\tilde\Bm\wMp{L:\ell+1} \right\|^2_F \ \ \bigg\lvert  \ \ \mathcal{F}_\ell \ \ \right].
 \end{align}
 As usual, we drop the filtration notation and plug this back into the expression for $\E\|\mb H^{k\ell}_{21}\|_F$:
 \begin{align}
     \E\|\mb H^{k\ell}_{21}\|_F&\le \E^{1/2}\left\|\wMp{k-1:1} \z\z^\top \wMp{1:\ell}\right\|^2_F\E^{1/2}\left\|\wMp{\ell+1:L}\tilde\Bm^\top\tilde\Bm\wMp{L:\ell+1} \right\|^2_F\E^{1/2}\left\|\wMp{\ell-1:k+1}\right\|^2_F\nonumber\\
     &\le \E^{1/4}\left\|\wMp{k-1:1} \z\right\|^4_F \E^{1/4}\left\|\wMp{\ell:1}\z\right\|^4_F + \E^{1/2}\left\|\tilde\Bm\wMp{L:\ell+1} \right\|^4_F\E^{1/2}\left\|\wMp{\ell-1:k+1}\right\|^2_F.\nonumber
 \end{align}
\end{proof}

\begin{tcolorbox}
\begin{restatable}[Bounding Hessians with forward passes, wide net]{proposition}{hess_pass_inf} As $d\to\infty$, for $k\ne\ell$
\begin{align}
    \E\|\mb H^{k\ell}\|_F &\lesssim (p\sigma^2 d)^{\frac{L-2}{2}} + (p\sigma^2 d)^{L-1}, \\
    \E\|\mb H^{kk}\|_F &\lesssim (p\sigma^2 d)^{L-1},
\end{align}
where ``$\lesssim$'' denotes ``asymptotically less of equal than a multiple of''~(same as $\mathcal{O}$). Hence, we have
\begin{align}
    \E\|\mb H^{k\ell}\|_F\lesssim(p\sigma^2 d)^{\frac{L-2}{2}},\quad \E\|\mb H^{kk}\|_F\lesssim (p\sigma^2 d)^{L-1} \quad \quad \text{if }\quad  (p\sigma^2 d)\le 1 \quad \text{(vanishing regime)}.\nonumber\\
    \E\|\mb H^{k\ell}\|_F\lesssim (p\sigma^2 d)^{L-1},\quad \E\|\mb H^{kk}\|_F\lesssim (p\sigma^2 d)^{L-1}  \quad \quad \text{if }\quad  (p\sigma^2 d)\ge 1 \quad \text{(exploding regime)}.\nonumber
\end{align}
\end{restatable}
\end{tcolorbox}

\begin{proof}
Follows from the triangle inequality. The only element left to bound is the transpose tensor $\mathbb{T}$, which however has only polynomial frobwnius norm in $d$ and $L$.
\end{proof}

\onecolumn

\section{Curvature Adaption of RMSProp}
\label{app:RMS}
\subsection{Literature review}
\paragraph{Role of adaptive methods in modern-day deep learning.} Adam and RMSprop~(as well as explicitly regularized variants such as AdamW~\citep{loshchilov2017decoupled}) are known to perform extremely well compared to SGD when training attention models~\citep{zhang2019adam,liu2019variance,wolf-etal-2020-transformers,brown2020language}, generative models \citep{karras2020training} and RNNs~\citep{hochreiter1997long}. In convolutional neural networks, many works \citep{loshchilov2017decoupled,balles2018dissecting,chen2018closing,liu2019variance} show that slight variations of adaptive methods (e.g. AdamW) can close a suspected generalization gap \citep{wilson2017marginal} and recently, \citep{tan2019efficientnet, tan2019mnasnet} achieved state of the art on ImageNet classification training with RMSprop~(Adam with $\beta_1=0$). Finally, in the context of generative adversarial nets~\citep{goodfellow2014generative}, RMSprop is often preferred over Adam, leading to fast convergence~\citep{park2019semantic,brock2018large,karras2020analyzing}, as opposed to SGD with momentum~(see e.g.~\citep{gempunreasonable} and references within). 
\vspace{-3mm}
\paragraph{The stochastic non-convex optimization approach to Adam.} The first correct\footnote{Quite interestingly, the original proof in~\cite{kingma2014adam} contains a few mistakes due to the problematic non-monotonic decrease of the effective stepsize.} proof of convergence for (a modified variant of) Adam in the stochastic nonconvex case was given in~\cite{reddi2019convergence}, subject to a few assumptions~(e.g. bounded gradients, decreasing stepsizes) and under the framework of online optimization. Perhaps the most recent simple, up-to-date, complete, and elegant proof of convergence of Adam was recently given in~\citep{defossez2020convergence}, where the authors show that a rate $\mathcal{O}(\log(k)/\sqrt{k})$ can be achieved in expectation with iterate averaging yet no additional assumption on the maximum gradient norm~(as opposed to most previous work). The same exact rate is also achieved by vanilla SGD~\citep{ghadimi2013stochastic}, which is well-known to be optimal for non-convex stochastic programming with bounded variance \citep{arjevani2019lower}, if one does not rely on variance reduction~(else, one can achieve slightly faster convergence~\citep{cutkosky2019momentum}). Hence, it is clear that worst-case first-order complexity bounds which can be derived using the standard non-convex optimization methodology~(at least for first order stationary points) are not yet able to explain the superiority of Adam in the context of optimization of deep neural networks. However, for Padam~\citep{chen2018closing} a slight variation of Adam, it is possible to show a better dependency on the problem dimension, compared to (the known upper bound for) SGD.
\vspace{-3mm}
\paragraph{Geometry adaptation, noise rescaling, gradient clipping and other conjectures.} Some papers on Adam do not take the standard non-convex optimization approach discussed above. Chronologically, the first was~\citep{balles2018dissecting}, that ``dissects'' Adam highlighting a variance-dependent preconditioning effect. This insight was taken one step further by~\citep{staib2019escaping}, that shows how the variance-adaptation of \textit{Adam effectively scales the gradient variance to be isotropic}; the authors claim this effect leads to fast escape from saddle points, since all eigendirections are equally excited by the stochastic perturbation.

Other papers take instead a geometric approach and motivate how Adam and RMSprop provide a cheap diagonal approximation of the empirical Fisher preconditioner, which is sometimes related to the Hessian~\citep{martens2020new}. As a result, \textit{Adam can be thought of as an approximate Gauss-Newton method}~\citep{nocedal2006numerical}. While this interpretation could in principle explain the fast convergence of Adam, it was recently shown~\citep{kunstner2019limitations} that very often the Adam preconditioner can be very far away from the true Hessian, which is instead provably related to the non-empirical~(a.k.a. true) Fisher. However, in~\citep{dauphin2015equilibrated}, the authors show that RMSprop effectively regularizes the landscape, making it more ``equilibrated'' (well-conditioned).

Finally,~\cite{zhang2019gradient} relate the success of adaptive methods to the underlying gradient clipping effect: if sporadic big stochastic gradients are encountered, those are smoothed out by the effect of the variables $m$ and $v$. The authors are able to show that clipped SGD, under quite uncommon assumptions on the cost function, is able to slightly improve~(by a constant factor) over the rate of SGD, in some particular cases. While this does provide a quantitative improvement on SGD~(not the case e.g. for the rates in~\citep{defossez2020convergence}), the result is arguably not strong enough to motivate the success of adaptive methods.

\subsection{Behaviour of RMSprop on the chain}
\begin{figure}[ht!]
    \centering
    \begin{tabular}{c|c}
    \includegraphics[width=0.4\textwidth]{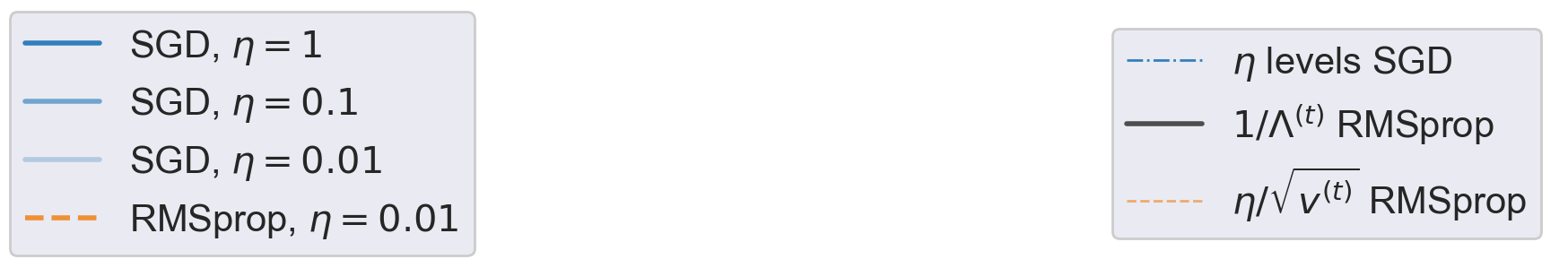} &\\
    Run \#1 $(\tau=0.2)$ & Run \#2 $(\tau=0.2)$\\
    \includegraphics[width=0.48\textwidth]{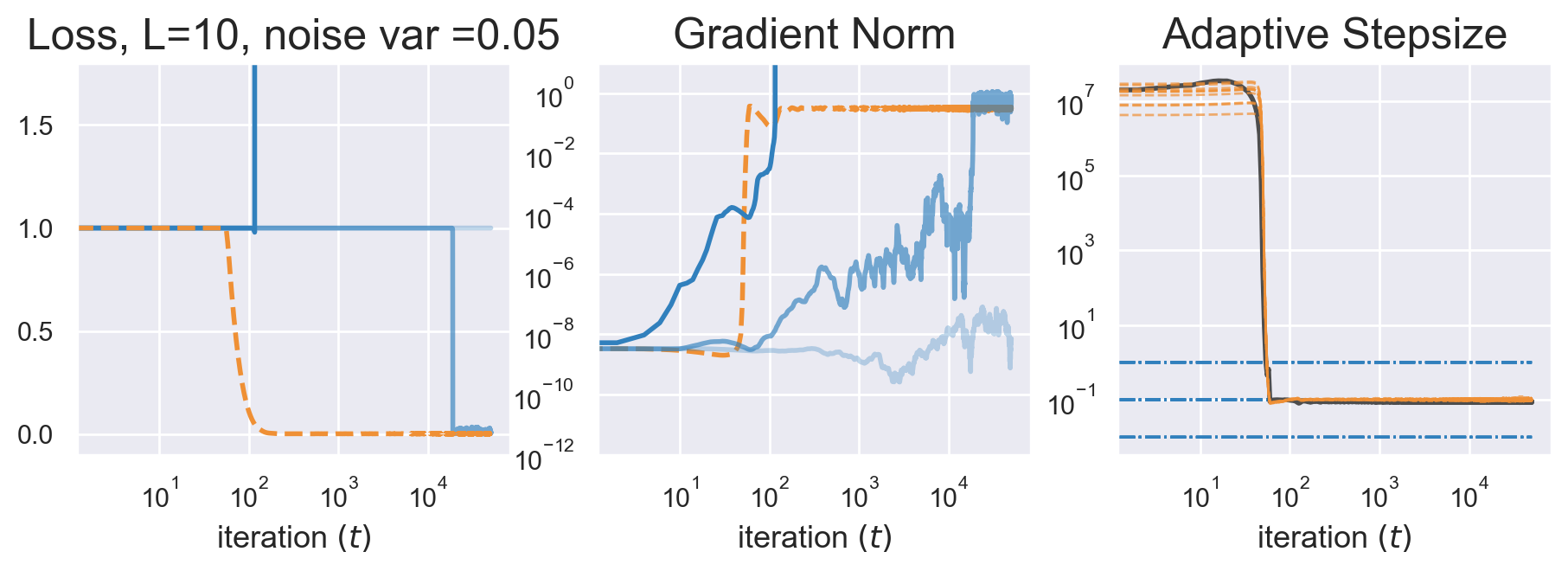} &
    \includegraphics[width=0.48\textwidth]{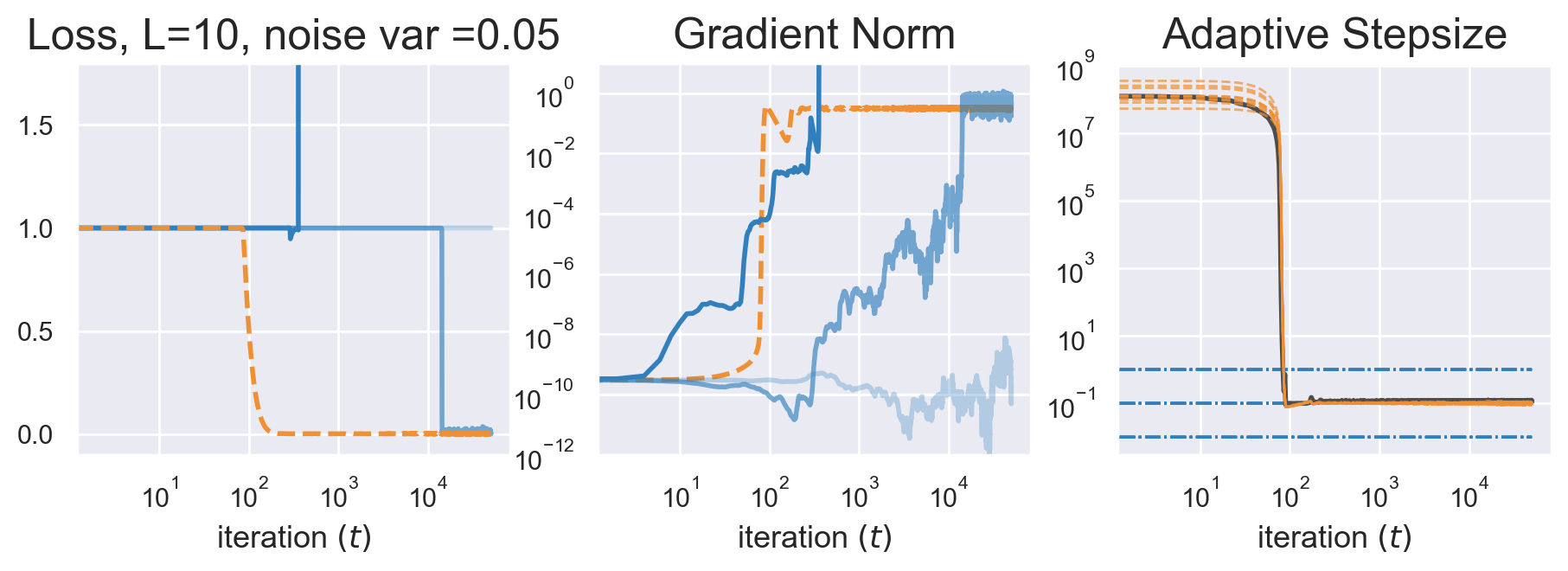}\\
    \hline
    \includegraphics[width=0.48\textwidth]{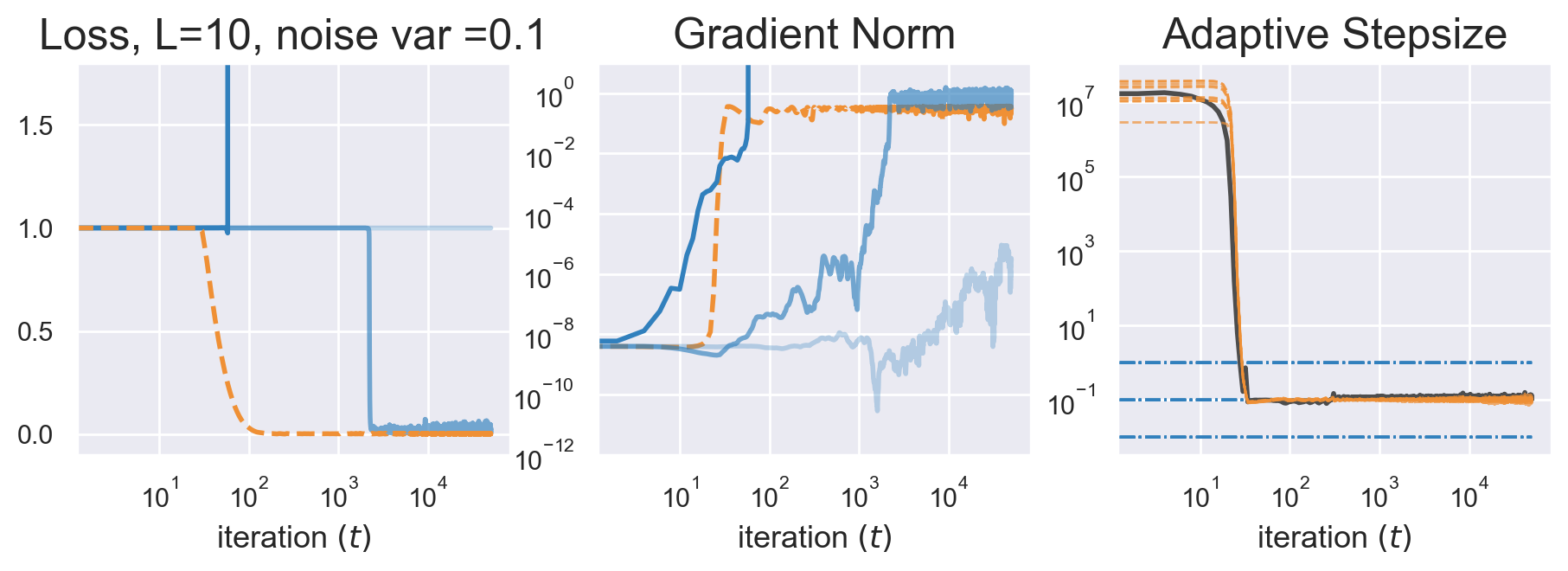}&
    \includegraphics[width=0.48\textwidth]{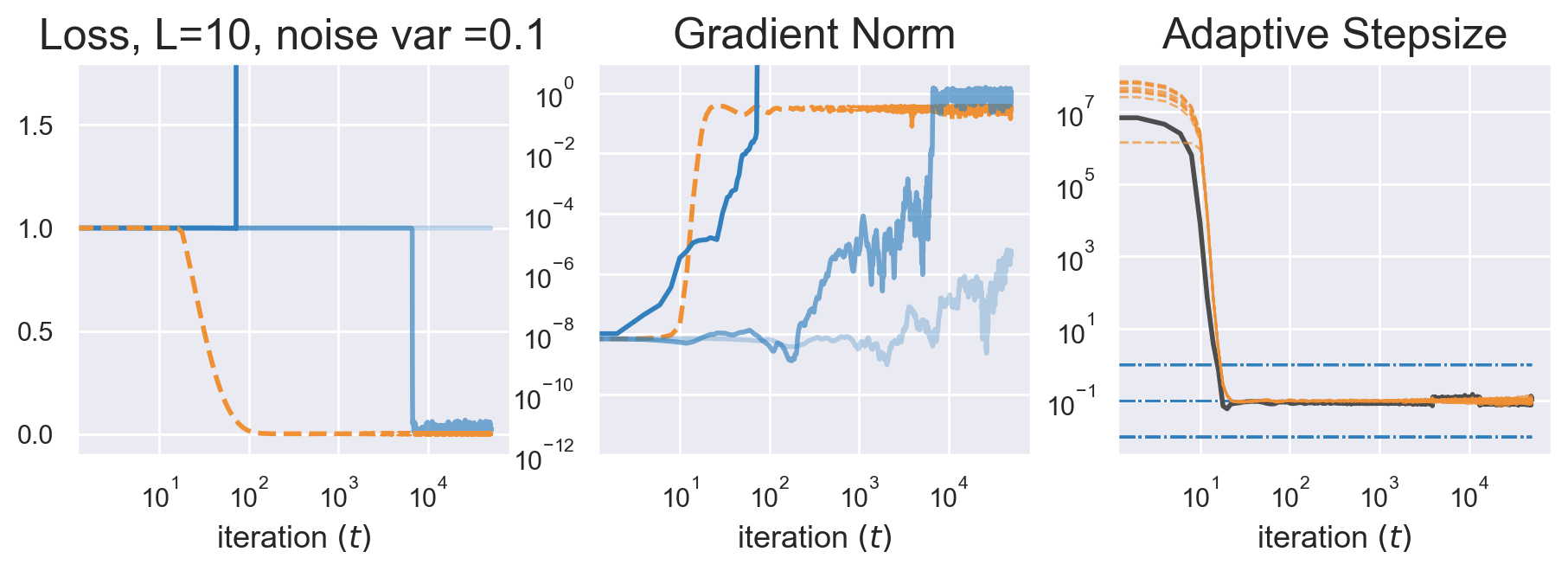}\\
    \hline
    \includegraphics[width=0.48\textwidth]{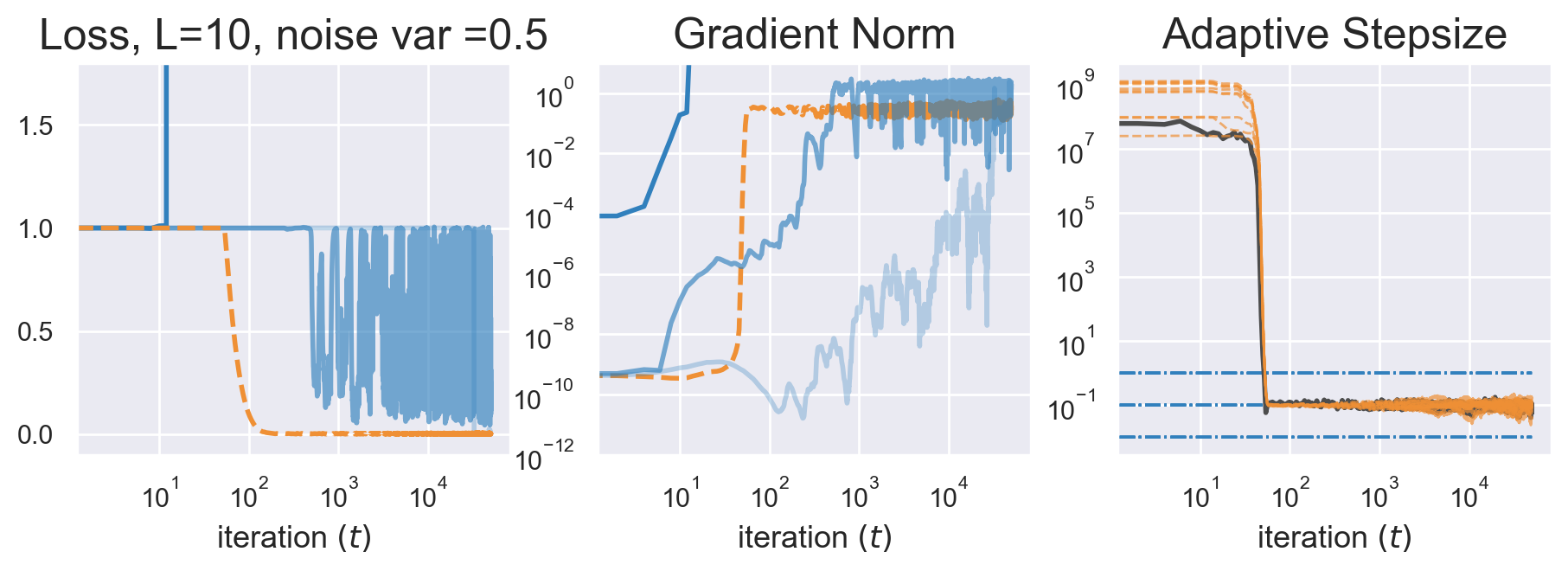}&
    \includegraphics[width=0.48\textwidth]{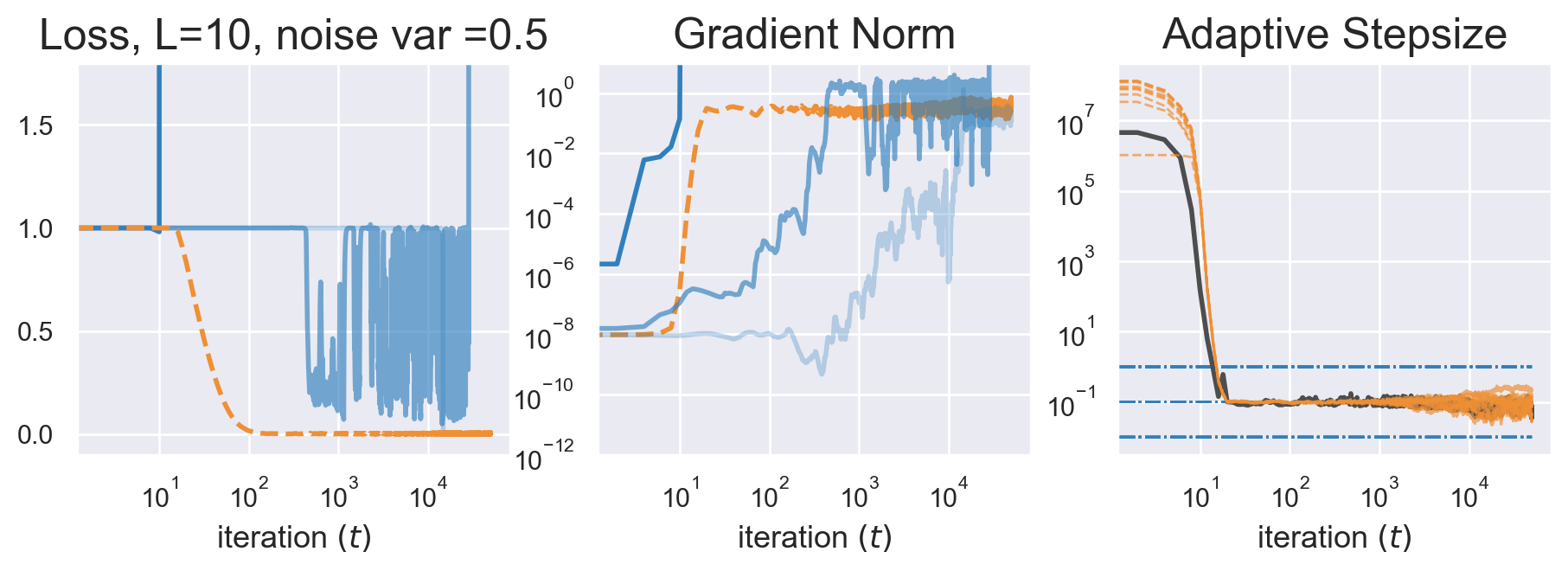}\\
    \end{tabular}
    \caption{Optimization of a ten-dimensional chain, $w_i(0)\sim \mathcal{U}[-0.2,0.2]$. Injected is an isotropic Gaussian noise of standard deviations $0.05, 0.1, 0.5$ (two additional runs in Fig.~\ref{fig:RMSprop_app_d10_true_2}). While moderate noise helps GD~(blue), no choice of stepsize is able to provide a performance comparable to RMSprop, which escapes after less than 100 iterations. Notably, the effective RMSprop stepsize matches the inverse curvature and is robust to noise.}
    \label{fig:RMSprop_app_d10_true_1}
\end{figure}

We empirically study the behavior of RMSprop on the chain loss with one data point $(x,y)=(1,1)$:
\begin{equation}
    \Ls_{\text{chain}}(\w) = \frac{1}{2}(1-w_Lw_{L-1}\cdots w_1)^2.
\end{equation}
While this cost function is very simple, it showcases the adaptiveness of RMSprop in an extremely clean and concrete way. We consider $L= 10$ and initialize each weight to $w_i(0)\sim \mathcal{U}[-0.2,0.2]$, to induce vanishing gradients~(order $10^{-8}$) and curvature~(order order $10^{-7}$), as it can be seen from Figure~\ref{fig:RMSprop_app_d10_true_1} and is predicted by Theorem~\ref{thm:as} and Corollary~\ref{corr:ev}. This initialization is close to $\w=\mathbf{0}$, which is clearly a saddle point because all partial derivatives vanish, but there exist directions of both increases and decreases: increasing all $w_i$ simultaneously to $\epsilon>0$ makes the loss decrease, while increasing half~(i.e. five) of them to $\epsilon$ and decreasing the other half to $-\epsilon$ makes the loss increase.

\begin{figure}[ht!]
    \centering
    \begin{tabular}{c|c}
    Run \#3 $(\tau=0.2)$ & Run \#4 $(\tau=0.2)$\\
    \includegraphics[width=0.48\textwidth]{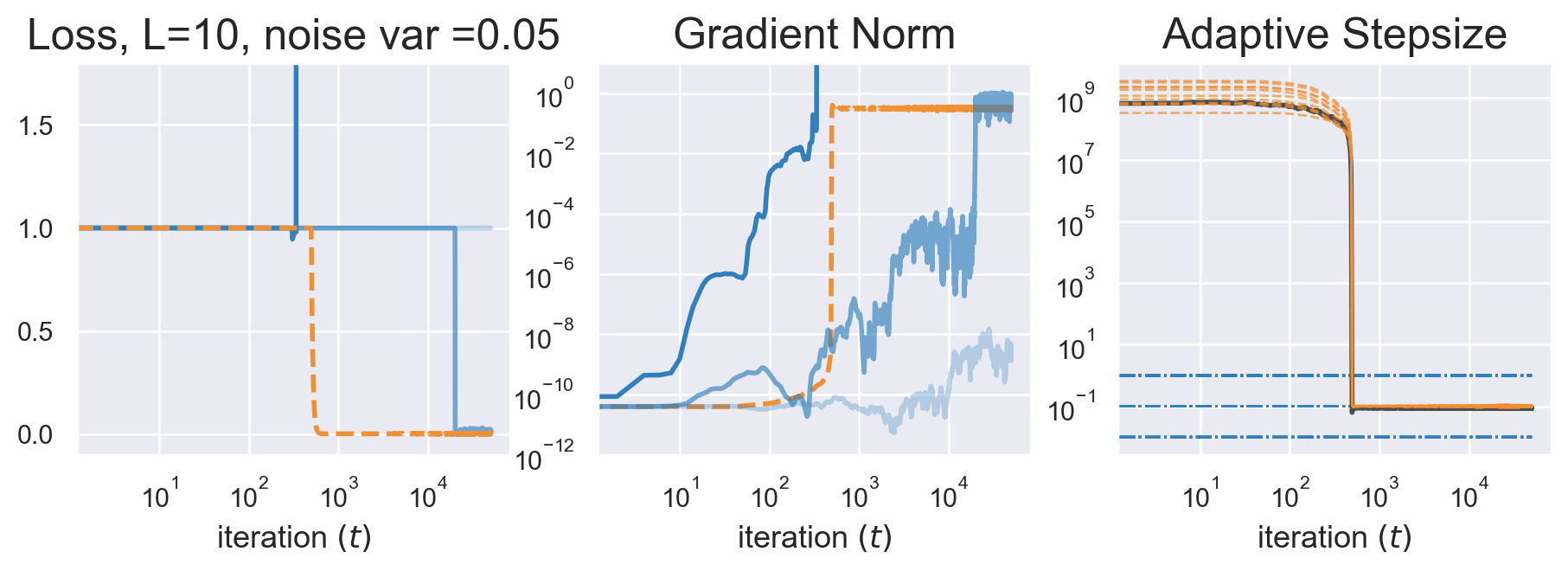} &
    \includegraphics[width=0.48\textwidth]{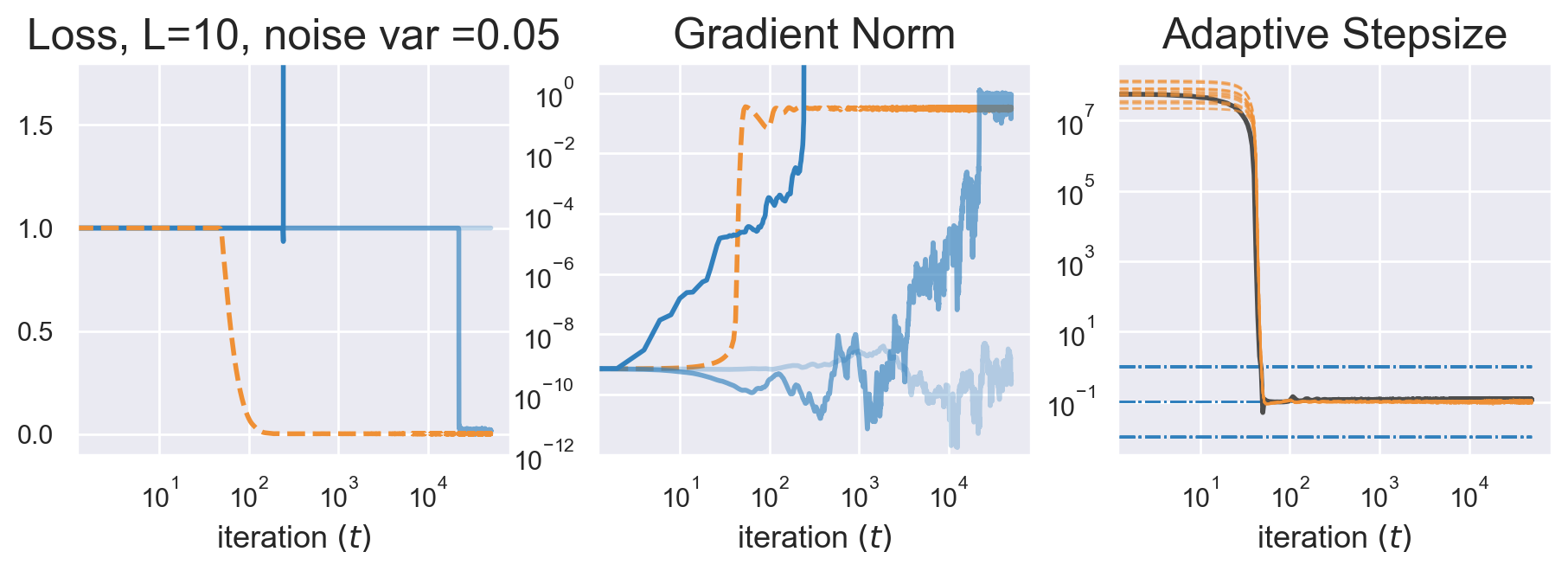}\\
    \hline
    \includegraphics[width=0.48\textwidth]{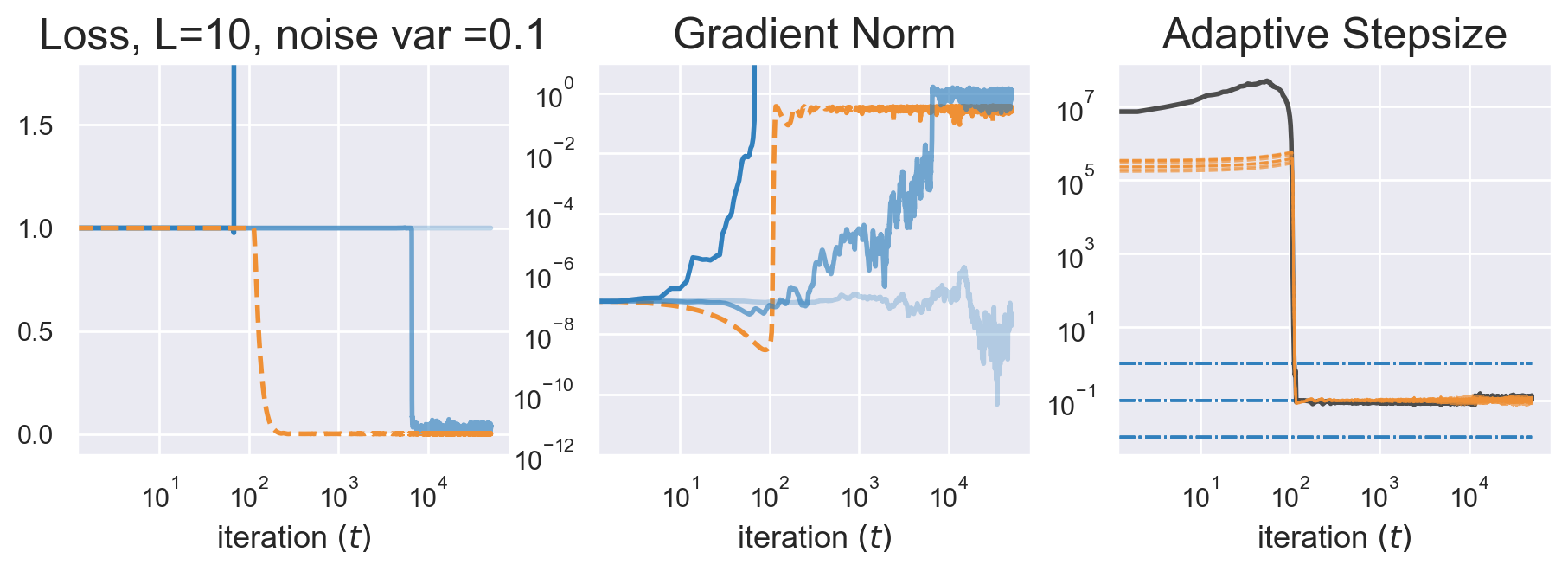}&
    \includegraphics[width=0.48\textwidth]{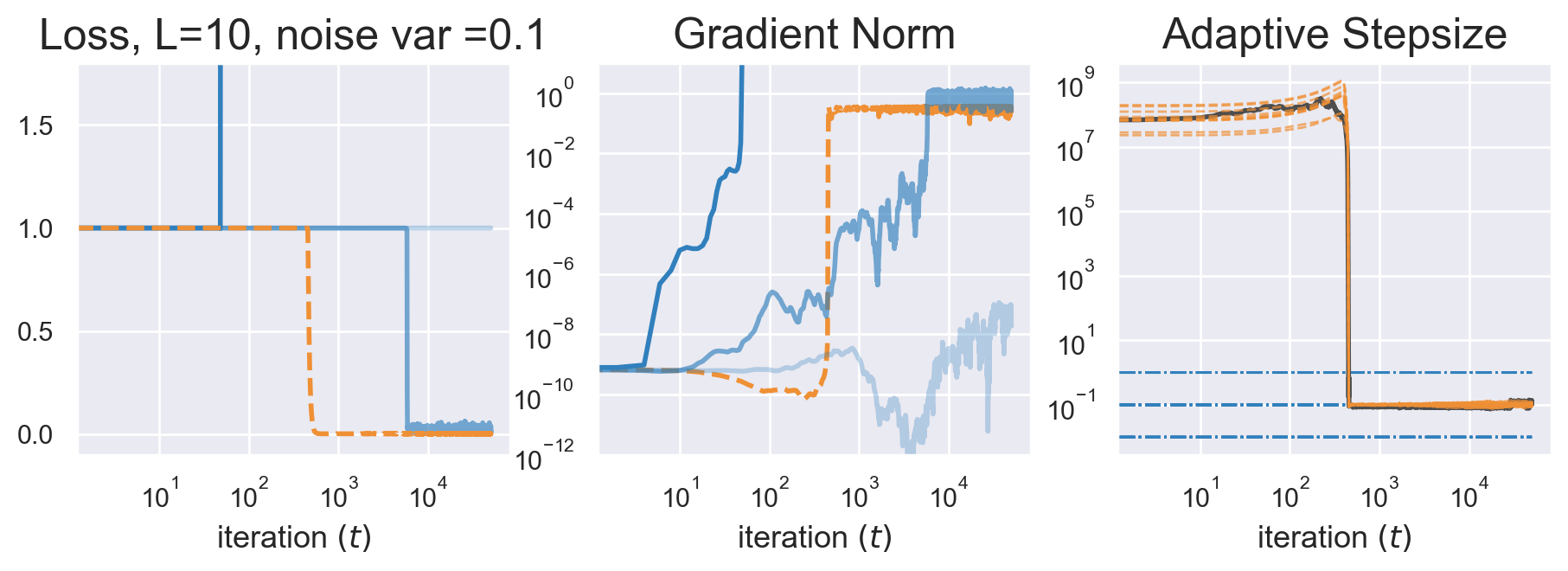}\\
    \hline
    \includegraphics[width=0.48\textwidth]{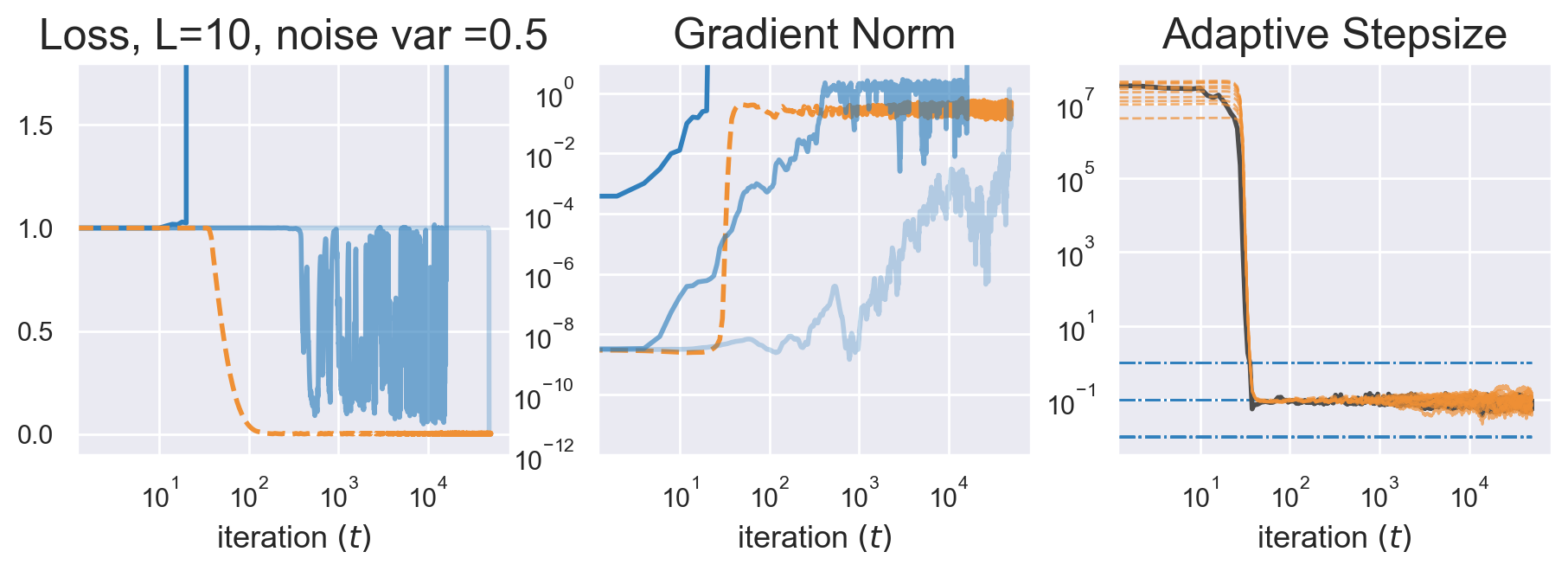}&
    \includegraphics[width=0.48\textwidth]{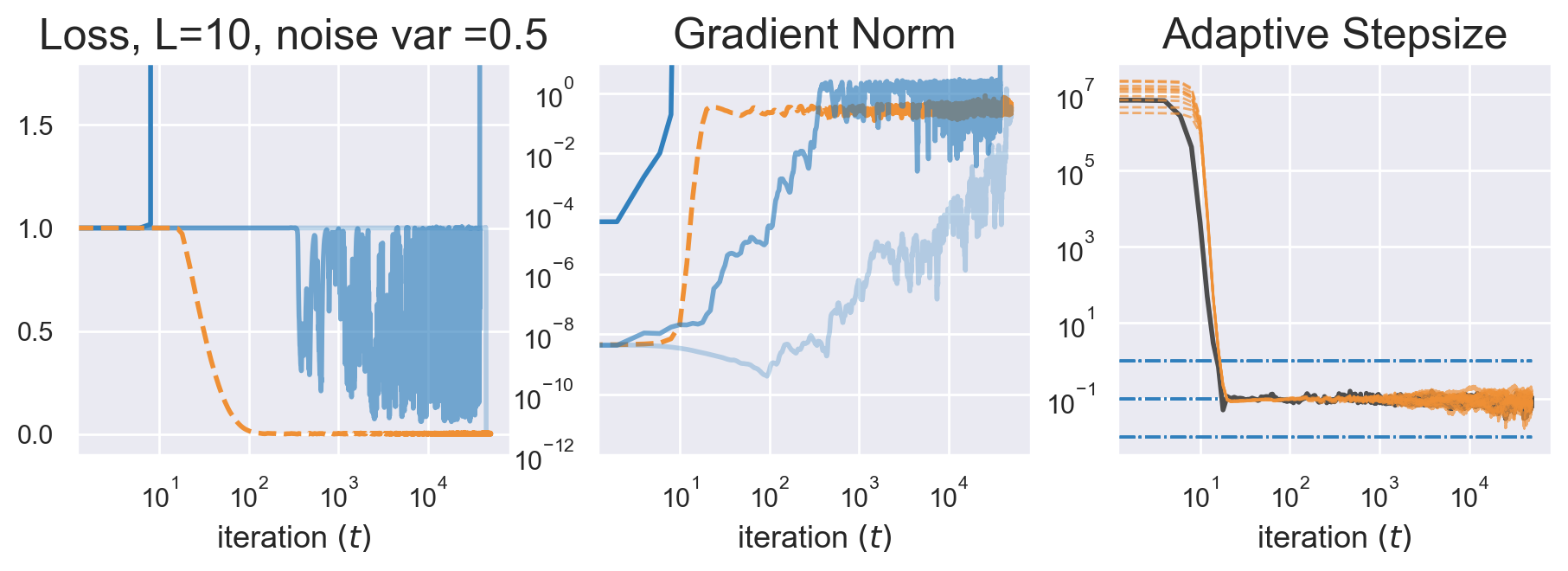}\\
    \end{tabular}
    \caption{Two additional runs, same settings and legend as Figure~\ref{fig:RMSprop_app_d10_true_1}.}
    \label{fig:RMSprop_app_d10_true_2}
\end{figure}

\begin{figure}[ht!]
    \centering
    \begin{tabular}{c|c}
    \includegraphics[width=0.48\textwidth]{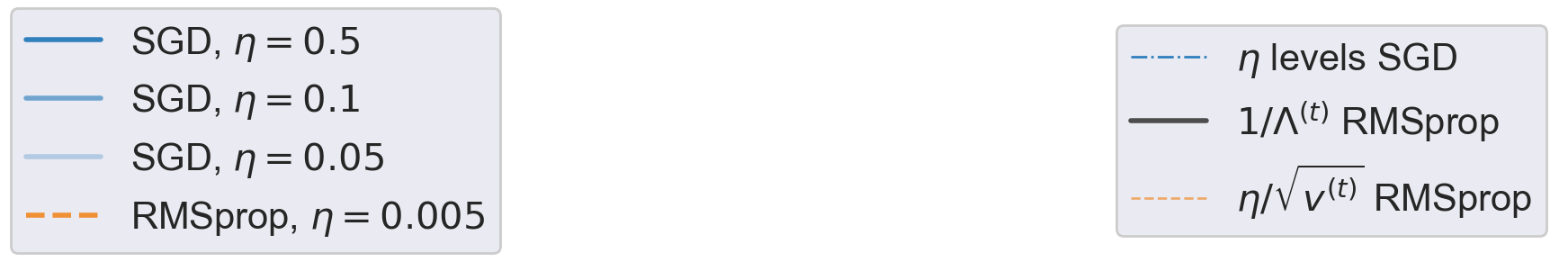} & \\
    Run \#1 $(\tau=1)$ & Run \#2 $(\tau=1)$\\
    \includegraphics[width=0.48\textwidth]{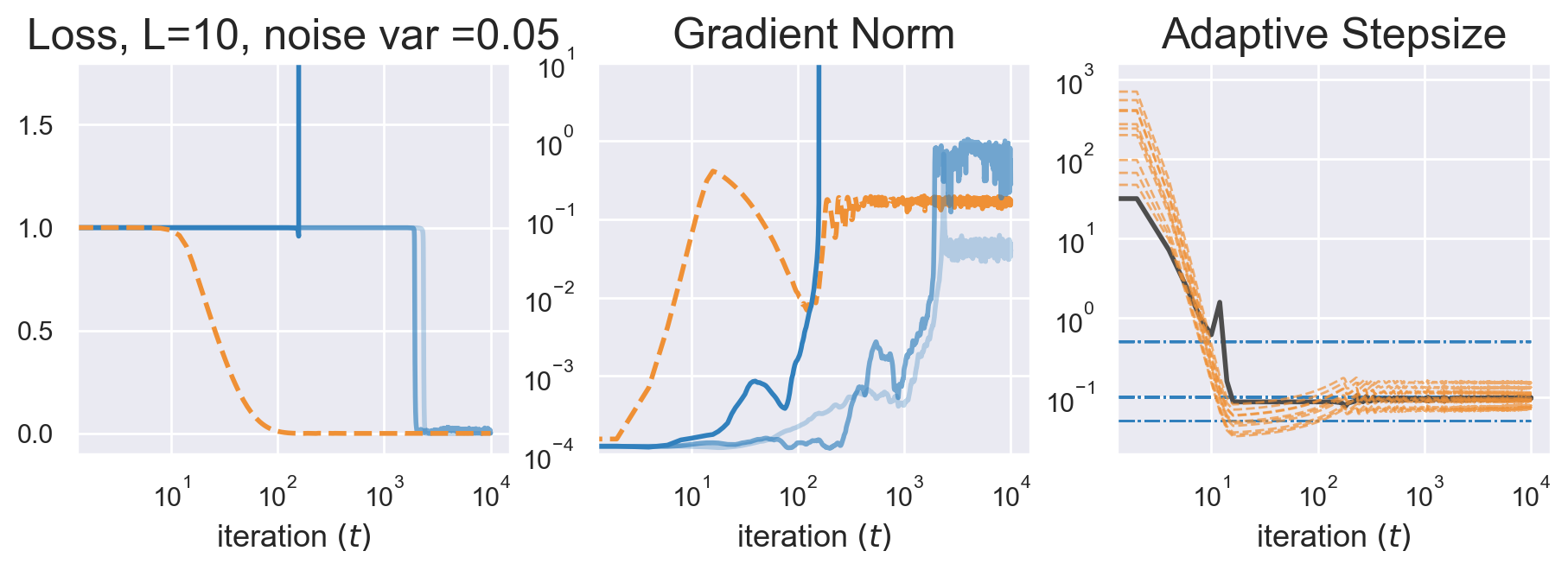} &
    \includegraphics[width=0.48\textwidth]{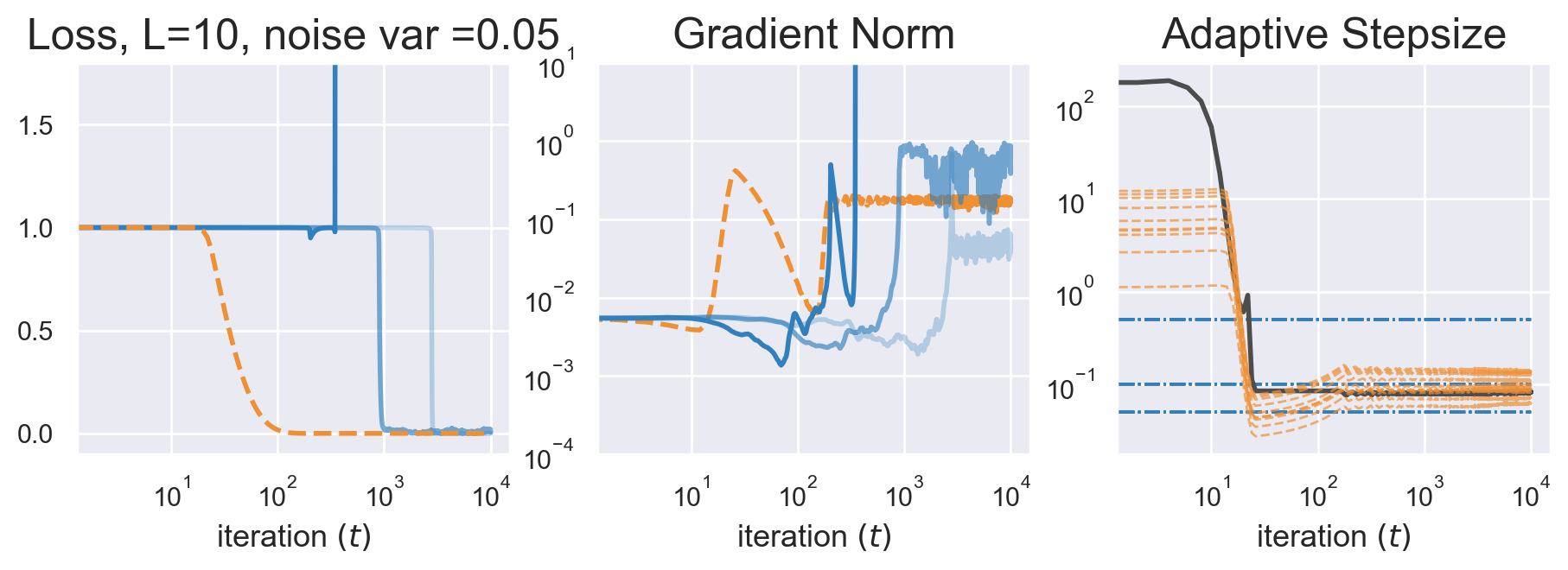}\\
    \hline
    \includegraphics[width=0.48\textwidth]{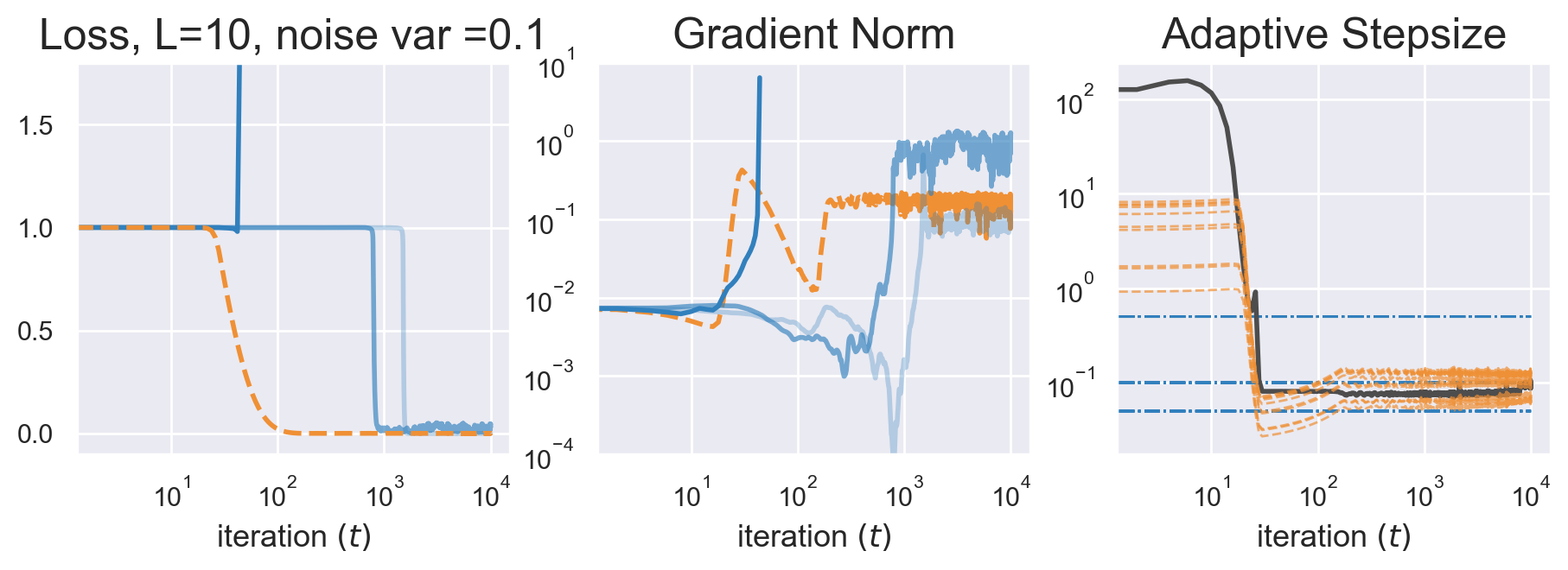}&
    \includegraphics[width=0.48\textwidth]{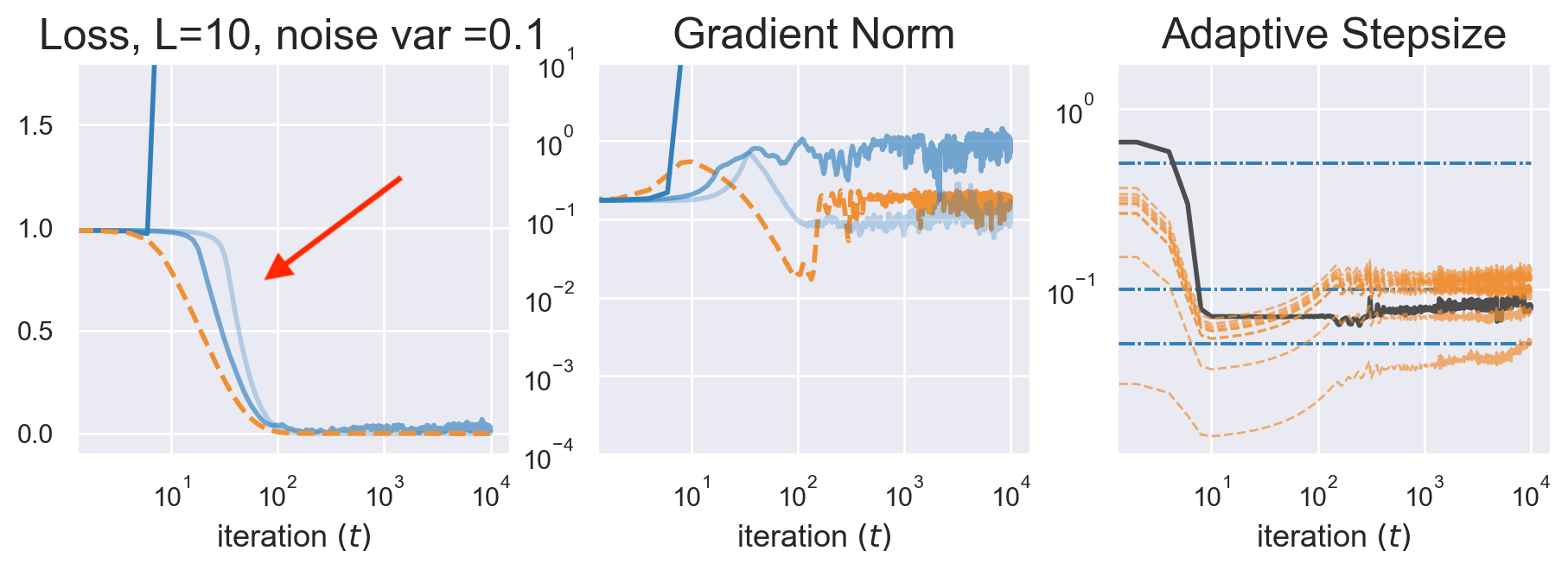}\\
    \hline
    \includegraphics[width=0.48\textwidth]{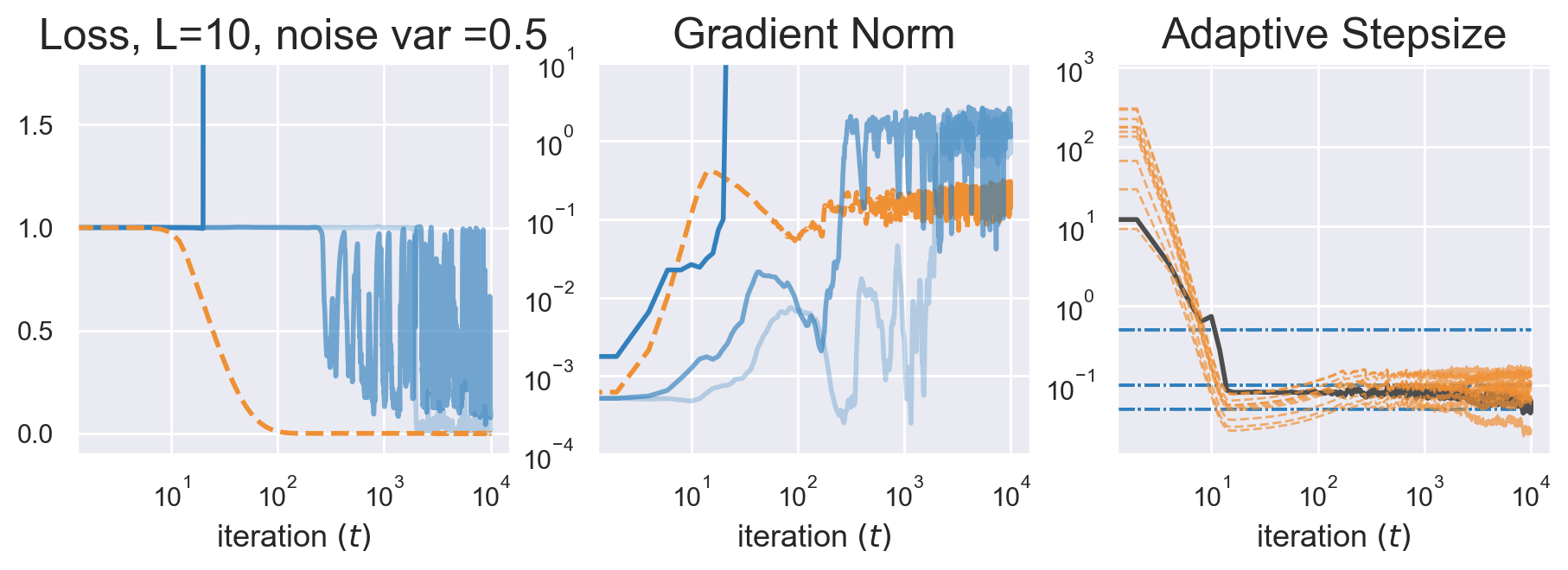}&
    \includegraphics[width=0.48\textwidth]{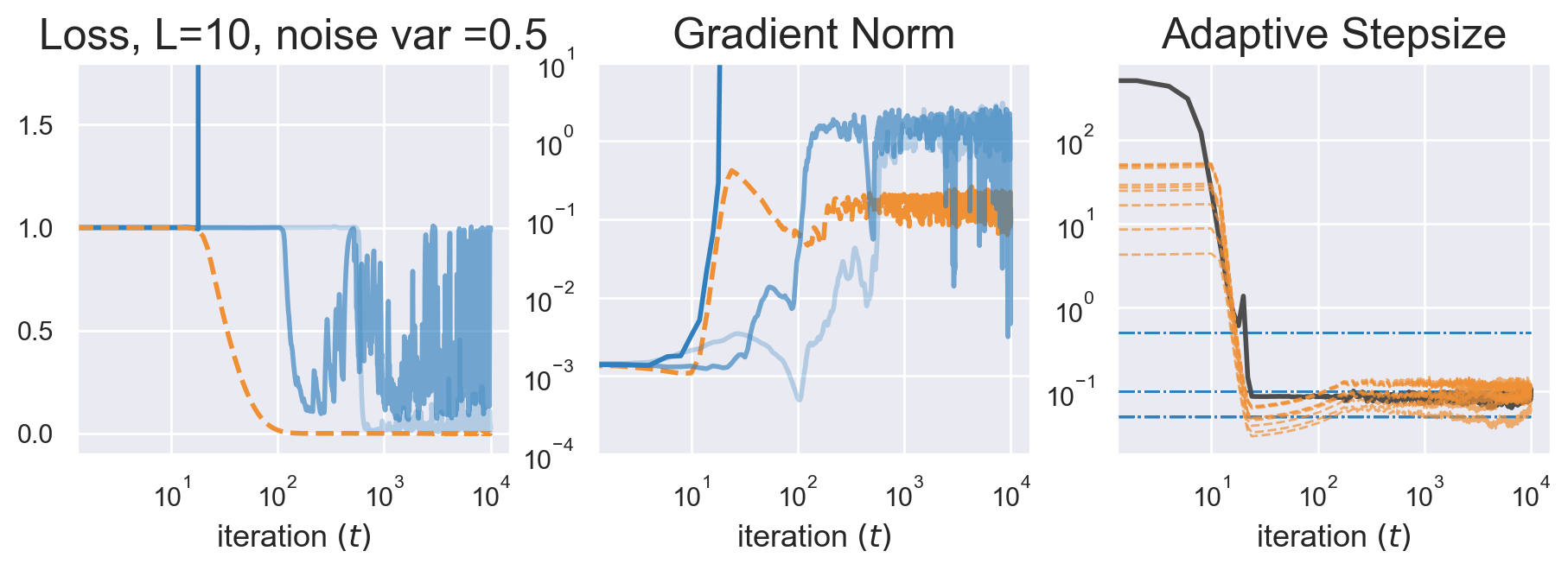}\\
    \end{tabular}
    \caption{Here we instead consider $w_i(0)\sim\mathcal{U}[-1,1]$, which induces less curvature-gradient vanishing compared to Fig.~\ref{fig:RMSprop_app_d10_true_1}\&~\ref{fig:RMSprop_app_d10_true_2}. If the initial gradient norm is high, Perturbed gradient descent gets now closer in performance to RMSprop. However, the gradient/Hessian norm at initialization has a considerable variance, which makes the performance less predictable compared to Fig.~\ref{fig:RMSprop_app_d10_true_1}\&~\ref{fig:RMSprop_app_d10_true_2}. This fact also makes the curvature adaptation feature of RMSprop less apparent.}
    \label{fig:RMSprop_app_d10_true_3}
\end{figure}

We show results for 4 different seeds, 3 different noise injection levels and 3 different GD stepsizes. Results are shown in Figure~\ref{fig:RMSprop_app_d10_true_1}\&~\ref{fig:RMSprop_app_d10_true_2}: while perturbed GD is slow to escape the saddle for any noise injection level and any stepsize, RMSprop is able to adapt to curvature and quickly escapes the saddle. This result validates the claim in Sec.~\ref{sec:width} of the main paper~(Prop.~\ref{prop:GD_slow}). In addition, Figure~\ref{fig:RMSprop_app_d10_true_3} shows that, if the initialization is such that the gradients at initialization are bigger, i.e. $w_i(0)\sim \mathcal{U}[-1,1]$, then the performance of perturbed gradient descent get can get closer to the one of RMSprop.

\paragraph{Notation.} We denote by $v(t)$ the exponential moving average~(with parameter $\beta_2 = 0.9$) of the squared gradients at iteration $t$ and by $\Lambda(t)$ the maximum Hessian eigenvalue at iteration $t$.

\section{Paths in CNNs}\label{sec:width_cnns}
Contrary to the fully connected architectures discussed above, convolutional neural networks are only sparsely connected and yield a large amount of weight sharing. It is thus not surprising to see that, given an MLP and a CNN of the same width to depth ratio (where the CNN width is defined as in Section \ref{subsec:conv}, i.e. $d=k^2c$), the CNN yields much small gradient (and Hessian) magnitudes than the MLP (compare Figure \ref{fig:width_effects_mlp} and \ref{fig:width_effects_cnn} when $d=\sqrt{L}$). This is so, because the paths going into each output neuron share most parameters and thus there is less probability for a path being atypically large.

The following Figure illustrates this for a CNN with one dimensional convolutions\footnote{For simplicity of presentation, the logic directly generalized to higher dimensional convolutions}. To be precise, we consider an input $x\in\mathbb{R}^3$, convolutional kernels $h^l\in\mathbb{R}^3$ and opt for circular padding. In this case one can write the CNN forward pass as $\Km^L\ldots \Km^1\x$ and the MLP forward pass as$ \Wm^L\ldots \Wm^1\x$, where 
$$
\Km^l=\begin{bmatrix}
h^l_2 & h^l_3 & h^l_1\\
h^l_1 & h^l_2 & h^l_3\\
h^l_3 & h^l_1 & h^l_2
\end{bmatrix} \quad \text{and} \quad \Wm^l=\begin{bmatrix} 
w^l_1 & w^l_2 & w^l_3\\
w^l_4 & w^l_5 & w^l_6\\
w^l_7 & w^l_8 & w^l_9
\end{bmatrix}.
$$

\begin{figure}[ht]

\begin{minipage}[b]{0.49\linewidth}
\centering
\includegraphics[width=0.9\textwidth]{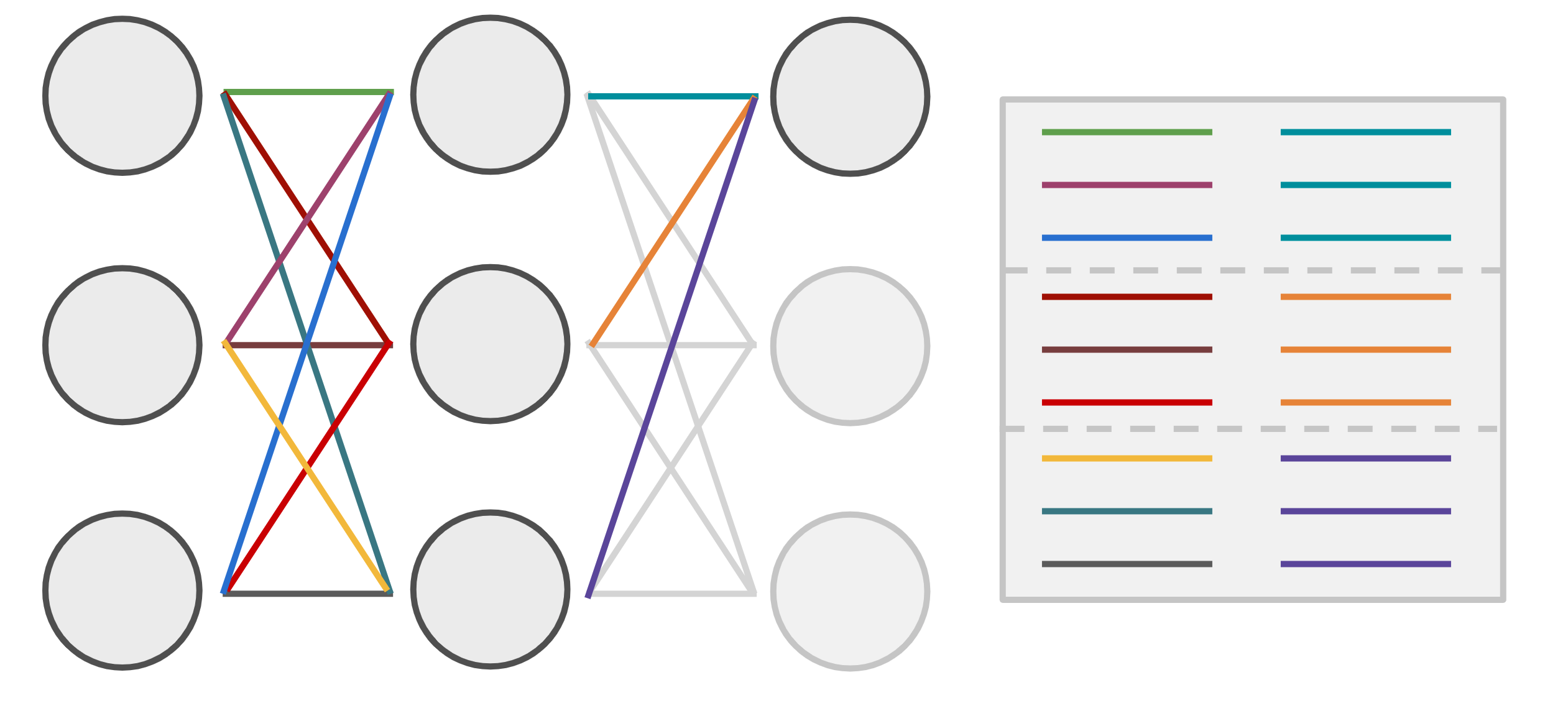}
\caption*{MLP}

\end{minipage}
\hspace{0.005cm}
\begin{minipage}[b]{0.49\linewidth}
\centering
\includegraphics[width=0.9\textwidth]{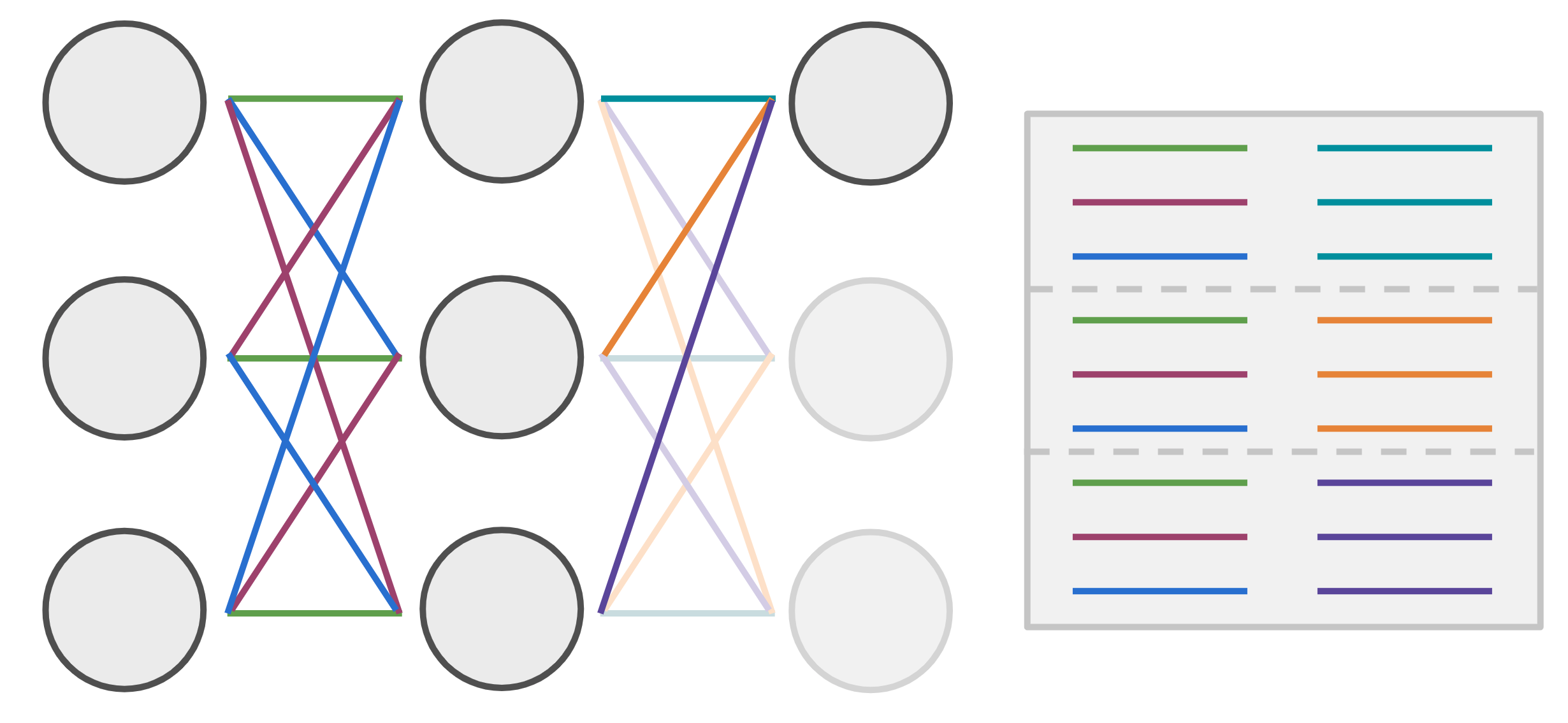}
\caption*{CNN}

\end{minipage}
\caption{Paths in MLP and CNN of same width.}
\label{fig:paths}
\end{figure}

As can be seen from the above figure, the number of i.i.d. weights $w^l_i$ that contribute to an output neuron in the MLP grows \textit{exponentially}, i.e. for each of the $d=3$ weights in layer $L$ there are another $d$ weights contribution in layer $L-1$ and so on. In the CNN case, this growth is (despite the existence of the same number of paths) only \textit{additive} since all neuron in a given layer $l$ share their incoming weights. As a result, it is no surprising to see that when comparing narrow CNNs to narrow MLPs, the vanishing gradients/curvature is much more pronounced in the former (compare Fig. \ref{fig:width_effects_cnn} and \ref{fig:width_effects_mlp}).

\end{document}